%% file: main.tex
\documentclass[10pt,twocolumn,letterpaper]{article}

\usepackage[pagenumbers]{cvpr} 

\input{preamble}

\definecolor{cvprblue}{rgb}{0.21,0.49,0.74}
\usepackage[pagebackref,breaklinks,colorlinks,allcolors=cvprblue]{hyperref}

\def\confName{CVPR}
\def\confYear{2027}

\title{RoboBuilder: Agentic Robot Co-Design with Real-world\\Video Feedback in the Loop}

\author{Ghadi Nehme \quad Dahyun Nam \quad Akshay Govind Srinivasan \quad Faez Ahmed\\
Department of Mechanical Engineering, Massachusetts Institute of Technology\\
Cambridge, MA, USA\\
{\tt\small \{ghadi, diananam, akshaygs, faez\}@mit.edu}
}

\usepackage[capitalize,noabbrev]{cleveref}
\usepackage{placeins}
\usepackage{float}  
\usepackage{titletoc}
\titlecontents{section}[2.0em]{\addvspace{6pt}\bfseries}{\hspace*{-2.0em}\thecontentslabel\hspace{1em}}{}{\hfill\contentspage}
\titlecontents{subsection}[4.6em]{\small}{\hspace*{-2.6em}\thecontentslabel\hspace{1em}}{}{\titlerule*[0.6pc]{.}\contentspage}
\input{numbers}

\renewcommand{\dbltopfraction}{0.8}\renewcommand{\topfraction}{0.85}
\begin{document}
\maketitle
\input{fig/teaser}
\input{sec/0_abstract}
\input{sec/1_intro}
\input{sec/2_related}
\input{sec/3_method}
\input{sec/4_experiments}
\input{sec/5_conclusion}
\input{sec/8_llm_usage}
{
    \small
    \bibliographystyle{ieeenat_fullname}
    \bibliography{main}
}
\clearpage
\appendix
\startcontents[appendices]  
\phantomsection
\onecolumn
\section*{Table of Contents for Appendices}
\printcontents[appendices]{}{1}{\setcounter{tocdepth}{2}}
\newpage
\input{sec/X_suppl}
\end{document}

%% file: preamble.tex
\usepackage{iftex}
\iftutex  
\usepackage{fontspec}
\fi

\usepackage{cuted} 

\usepackage{currfile} 

\usepackage{caption} 

\usepackage{graphicx,booktabs,multirow,xspace,enumitem,amsmath,amssymb,pifont,tabularx,array,colortbl}
\input{logos}
\definecolor{rbBg}{HTML}{FFFFFF}
\definecolor{rbBgTwo}{HTML}{F5F3EF}
\definecolor{rbFg}{HTML}{1C1E22}
\definecolor{rbMute}{HTML}{5E646C}
\definecolor{rbAccent}{HTML}{B87500}
\definecolor{rbAccentTwo}{HTML}{D9580B}
\definecolor{rbCard}{HTML}{F6F4F0}
\definecolor{rbLine}{HTML}{E3DFD8}
\definecolor{rbOk}{HTML}{0E8A55}
\definecolor{rbBad}{HTML}{C2410C}
\definecolor{rbOpus}{HTML}{E69F00}
\definecolor{rbSonnet}{HTML}{56B4E9}
\definecolor{rbAstra}{HTML}{009E73}
\definecolor{rbSol}{HTML}{CC79A7}
\definecolor{rbPh}{HTML}{1F5FBF}   

\newcommand{\bench}{RoboBuilder\xspace}

\newcommand{\taskCross}{\textsc{Cross}\xspace}
\newcommand{\taskPrint}{\textsc{Cross+Print}\xspace}
\newcommand{\taskLight}{\textsc{Light}\xspace}
\newcommand{\taskFast}{\textsc{Fast}\xspace}
\newcommand{\taskAll}{\textsc{All-in-one}\xspace}
\newcommand{\rbBench}{\textsc{RoboBuilder-Bench}\xspace}
\newcommand{\rbSim}{\textsc{RoboBuilder-Sim}\xspace}
\newcommand{\rbLab}{\textsc{RoboBuilder-Lab}\xspace}
\newcommand{\rbLoop}{\textsc{RoboBuilder-Loop}\xspace}
\newcommand{\opus}{Opus~5.5\xspace}
\newcommand{\sonnet}{Sonnet~5.5\xspace}
\newcommand{\astra}{GPT-6 Astra\xspace}
\newcommand{\sol}{GPT-6.1 Sol\xspace}
\newcommand{\cmark}{\textcolor{rbOk}{\ding{51}}}
\newcommand{\xmark}{\textcolor{rbBad}{\ding{55}}}
\newcommand{\pmark}{\textcolor{rbMute}{$\circ$}}

\renewcommand{\paragraph}[1]{\vspace{0.35em}\noindent\textbf{#1}\hspace{0.4em}}
\setlist{leftmargin=1.2em,itemsep=0.15em,topsep=0.25em}
\newcommand{\rowhead}{\rowcolor{rbBgTwo}}

%% file: logos.tex
\newcommand{\logoclaude}{\raisebox{-0.15em}{\includegraphics[height=0.9em]{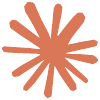}}}
\newcommand{\logoopenai}{\raisebox{-0.15em}{\includegraphics[height=0.9em]{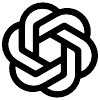}}}
\newcommand{\opusL}{\logoclaude\,Opus~5.5}
\newcommand{\sonnetL}{\logoclaude\,Sonnet~5.5}
\newcommand{\astraL}{\logoopenai\,GPT-6~Astra}
\newcommand{\solL}{\logoopenai\,GPT-6.1~Sol}

%% file: numbers.tex
\newcommand{\nSimLOne}{39}\newcommand{\nSimLTwo}{28}\newcommand{\nSimLThree}{19}
\newcommand{\nRealFail}{18}\newcommand{\nRealRate}{73}
\newcommand{\nFilmedLOne}{39}\newcommand{\nRealPassLOne}{30}\newcommand{\nFilmedLTwo}{28}\newcommand{\nRealPassLTwo}{19}
\newcommand{\nFilmedLThree}{19}\newcommand{\nRealPassLThree}{8}\newcommand{\nRealRateLThree}{42}
\newcommand{\nFilmedAll}{86}\newcommand{\nRealPassAll}{57}\newcommand{\nRealRateAll}{66}

\newcommand{\nAllRealOpus}{6}\newcommand{\nAllRealSol}{6}\newcommand{\nAllRealAstra}{5}\newcommand{\nAllRealSonnet}{2}
\newcommand{\nInitSimPass}{15}\newcommand{\nInitReal}{0}\newcommand{\nInitTries}{44}
\newcommand{\nCalRuns}{38}\newcommand{\nCalRight}{35}\newcommand{\nHeldOut}{10}\newcommand{\nHeldOutRight}{9}

\newcommand{\nVidBench}{304}\newcommand{\nVidTotal}{461}
\newcommand{\nRobotsBuilt}{135}\newcommand{\nPiecesBuilt}{13{,}000}\newcommand{\nWorkHours}{240}

\newcommand{\nAfterLoopRate}{93}\newcommand{\nBestAfterLoop}{100}
\newcommand{\nSimsAX}{3{,}538}\newcommand{\nSimsBC}{11{,}689}\newcommand{\nSimsAll}{1{,}200}\newcommand{\nSimsAgents}{16{,}427}
\newcommand{\nGeoAgents}{12{,}774}\newcommand{\nGeoBaselines}{2{,}493}

\newcommand{\nPool}{7}\newcommand{\nSingle}{6}\newcommand{\nPoolSpecialists}{9}
\newcommand{\pSimLOne}{98}\newcommand{\pSimLTwo}{70}\newcommand{\pSimLThree}{48}
\newcommand{\pRealLOne}{77}\newcommand{\pRealLTwo}{68}

\newcommand{\pLoopCut}{72}  

%% file: fig/teaser.tex
\begin{strip}
\centering
\vspace{-3.6em}
\includegraphics[width=\linewidth]{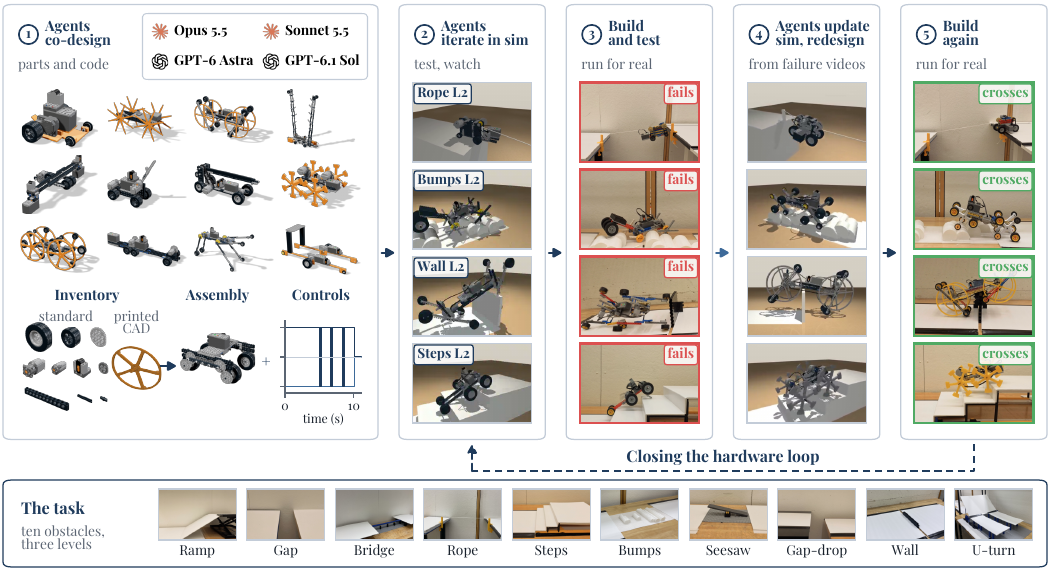}
\captionsetup{hypcap=false}
\captionof{figure}{\textbf{AI agents design robots that pass in simulation, and we physically build and test them; videos of real failures drive a simulator update and a redesign.}
\textbf{(1)}~Agents assemble standard parts, design a custom part that we 3D print, and write the program. \textbf{(2)}~Agents iterate in simulation. \textbf{(3)}~We build and test each passing design on the real course. \textbf{(4)}~From videos of its real failures, the agent updates its simulator and redesigns. \textbf{(5)}~We rebuild and retest. Bottom: the task, ten real obstacles at three difficulty levels each.}
\captionsetup{hypcap=true}
\label{fig:teaser}
\end{strip}

%% file: sec/0_abstract.tex
\begin{figure*}[t]
\centering
\includegraphics[width=\linewidth]{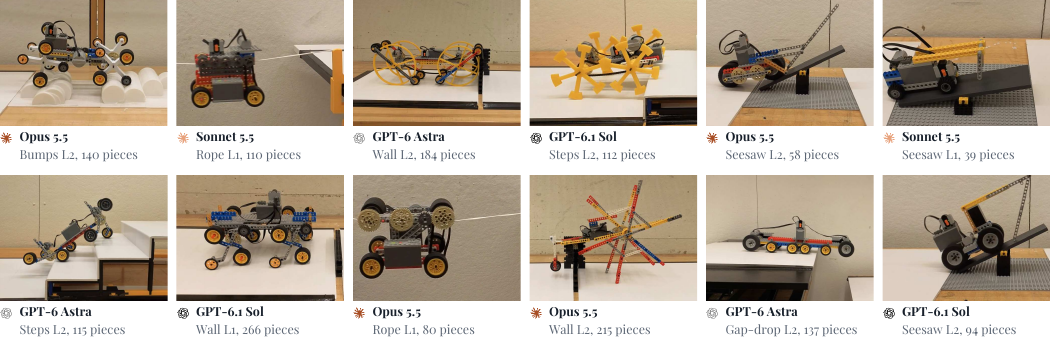}
\caption{\textbf{Agent-designed robots, built by hand on their real courses.} Twelve of the \nRobotsBuilt{} robots we built: whegs, printed hoops and cam wheels, cord trolleys, lever arms, legs and a spoked wheel that vaults the wall. More designs in \cref{sec:supp_gallerysec}.}
\label{fig:real_gallery}
\end{figure*}

\begin{abstract}
Can AI agents design robots that work in the real world? We introduce \bench, a benchmark in which coding agents co-design a robot's hardware and program from standard parts, motors and a custom 3D-printed part, and every design that passes in simulation is built and tested on a physical course. Simulation alone is not enough: under public part data, none of 15 initial passing designs crossed the real course, so an agent learns the simulator from measurements it requests and videos of real runs. Across four frontier coding agents, ten obstacles at three difficulty levels, \nSimsAgents{} simulations of \nGeoAgents{} distinct robots and \nRobotsBuilt{} physical robots built in an estimated \nWorkHours{} hours, agents solve nearly all easy obstacles and invent mechanisms our search baselines do not reach, but falter at harder levels, when one robot must cross every obstacle, and in the real world, where \nRealRateAll\% of simulated passes hold across all levels. Closing the loop helps: shown a video of its real failure, an agent redesigns the robot, sometimes correcting its simulator, and removes \pLoopCut\% of level-1 and level-2 failures. \bench measures progress toward agents that engineer real machines. Project Page: \url{https://ghadinehme.com/robobuilder.github.io/}
\end{abstract}

%% file: sec/1_intro.tex
\vspace*{-22pt}
\section{Introduction}
\label{sec:intro}

Engineering design is a loop: build a prototype, test it, learn why it failed, and iterate. Coding agents have mastered this loop for software, where a test runs in seconds and its verdict is exact. For a machine, the test is a physical build, and the simulator that stands in for it is wrong in ways only real runs reveal. We ask whether agents can run the full loop for functional robots: designing assemblies that achieve a task under constraints and objectives, and closing the sim-to-real gap the way engineers do, by building and testing.

Agents became capable engineers through executable feedback: writing code, running the tests and reading the failure~\cite{jimenez2024swebench,yang2024sweagent,shinn2023reflexion}. The same recipe now produces CAD models~\cite{makatura2023can,khan2024text2cad,nehme2026stepcad,nehme2026cadfit}, robot code~\cite{liang2023codeaspolicies} and machines in physics sandboxes~\cite{zhang2026compositional,xia2026buildarena}. These works judge geometry, or judge function in a simulator whose verdict is taken as truth. A functional assembly is held to more: it must achieve its task within the parts on hand, with joints that hold, gears that mesh and motors that can drive it, often as light or fast as possible, and only the physical world can confirm it.

Simulation alone does not suffice. With a simulator built from public data for our parts, \emph{none} of 15 initial robots that passed in simulation crossed the real course: tyres slipped, joints pulled apart, gears walked out of mesh and loads shifted. Sim-to-real methods calibrate a model of a fixed robot from real data~\cite{tobin2017domain,tan2018sim,chebotar2019closing}; in co-design the robot changes with every iteration, so the simulator must hold for designs nobody has built yet. A simulator never checked against reality rewards agents for exploiting it.

We introduce \bench, a benchmark for agentic robot co-design under constraints and objectives whose verdict is grounded in the real world (\cref{fig:teaser}). Agents design a robot and its open-loop program from an inventory of real parts, motors and a custom 3D-printed part (up to four copies), under building rules, to cross ten obstacles at three levels. They must pass each obstacle, pass with a printed part if needed, then make the design lighter or faster, and build one robot that crosses every obstacle (\cref{fig:benchmark}). The design space is vast. With 132 part types and about 3{,}000 parts in stock, a rough count gives about $10^{68}$ assemblies of only 25 pieces, while passing robots use a median of 114; each motor program adds about $10^{16}$ timings, and the printed part can be any shape an agent can write as CAD code (\cref{sec:supp_space}). Agents also close the sim-to-real gap themselves. An agent (\opus) first learns the simulator, requesting measurements and adjusting the physics until simulation matches training and validation videos of real runs. Every design that passes a task is then built and tested, and each real failure returns to its agent as video, which redesigns the robot, correcting its simulator where the video shows a modelling error, before the redesign is built again.

Four frontier coding agents pass \pSimLOne\% of level-1 tasks where search over the full inventory passes none, and invent whegs, printed hoops, cable cars and walkers that our search baselines, with the same budget but without prior knowledge, do not find (\cref{fig:real_gallery}; more in \cref{sec:supp_gallerysec}). Their success rate drops sharply at harder levels and when one robot must cross every obstacle. Iterating in simulation is essential, since a single call without it passes almost nothing, but more exploration does not predict better designs: the strongest agent passes with its first idea in over half its runs, suggesting that the agent's built-in engineering knowledge is its main edge. In the real world, \nRealRate\% of level-1 and level-2 simulated passes cross with the learned simulator, and \nAfterLoopRate\% after agents redesign from their failures.

\noindent\textbf{Contributions.}
\begin{itemize}[leftmargin=1.2em,itemsep=1pt,topsep=2pt]
\item \textbf{\rbBench.} Functional robot co-design under constraints and objectives: ten obstacles at three levels, four tasks plus integration, and a real-part inventory.
\item \textbf{\rbSim and \rbLoop.} Simulator learning from requested measurements and real videos, and a protocol that turns each real failure into simulator and design adjustments that are built and tested again. At levels 1 and 2, \nRealRate\% of simulated passes cross for real, and \nAfterLoopRate\% after the loop.
\item \textbf{Analysis of Agentic Design.} Performance and optimization, iteration in simulation, the distribution, novelty and diversity of designs (two to three times as diverse as search), how agents explore ideas, and how easily the designed robots can be assembled by hand.
\item \textbf{\rbLab and Dataset.} \nRobotsBuilt{} robots built by hand in an estimated \nWorkHours{} hours, \nSimsAgents{} simulations of \nGeoAgents{} robots, and \nVidTotal{} paired real videos.
\end{itemize}

%% file: sec/2_related.tex
\section{Related Work}
\label{sec:related}

\begin{figure*}[t]
\centering
\begin{minipage}[c]{0.635\linewidth}
\includegraphics[width=\linewidth]{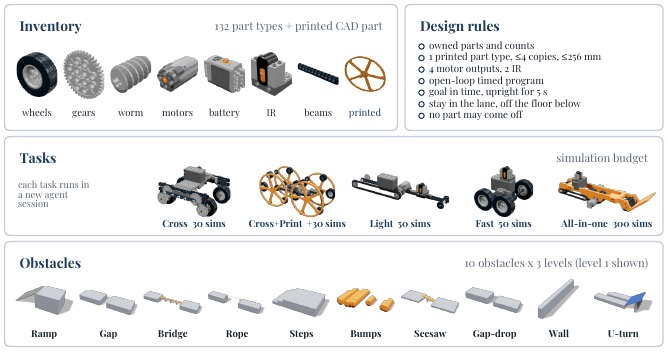}
\end{minipage}\hfill
\begin{minipage}[c]{0.345\linewidth}
\includegraphics[width=\linewidth]{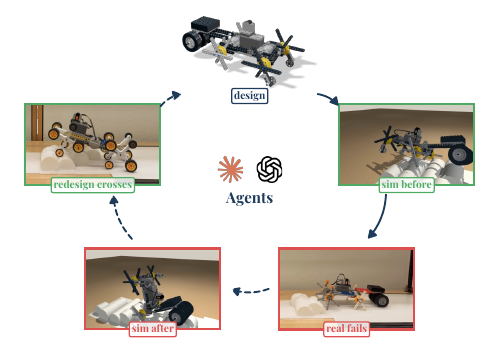}\\[2pt]
\includegraphics[width=\linewidth]{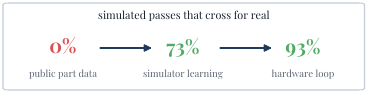}
\end{minipage}
\caption{\textbf{\bench asks for a machine that can be built, and closes the gap to the real world.} Left: agents combine inventory parts, motors and one custom CadQuery part type, printed in up to four copies, under design rules, and write an open-loop program; tasks and their simulation budgets; the ten obstacles, each at three levels. Right: one turn of the hardware loop (level-2 bumps, \opus): spoked wheels cross in simulation but spin against a ridge for real; after the agent lowers the friction of plastic on the bump rods the simulator fails the same way, and the redesign on worm-driven whegs crosses for real. Bottom right: share of simulated passes that cross for real (levels 1 and 2) with the simulator from public part data (initial designs), after simulator learning, and after one round of the loop's redesigns.}
\label{fig:benchmark}
\label{fig:cycle}
\end{figure*}

\noindent\textbf{LLM agents for engineering design.}
LLMs and agents generate CAD programs~\cite{khan2024text2cad,guan2025cadcoder,rukhovich2025cadrecode}, write and repair CAD code from renders and execution feedback~\cite{makatura2023can,nehme2026stepcad,nehme2026cadfit,alrashedy2025cadcodeverify,mallis2025cadassistant}, generate multi-part assemblies~\cite{dong2026muse}, operate CAD software~\cite{man2025videocad} and answer engineering design tasks~\cite{picard2025concept,guo2025engdesign}. For robots, LLMs write control code~\cite{liang2023codeaspolicies} and rewards~\cite{ma2024eureka}, serve as mutation operators that evolve robot designs~\cite{lehman2022elm,song2025laser}, co-design morphology with rewards~\cite{fang2025robomore}, through multi-agent debate~\cite{qiu2025debate2create} or with tools and their actions~\cite{gao2026vlmgineer}, and design mechatronics~\cite{wang2025mechatronics,elrefaie2025agents}. Closest to us, BesiegeField~\cite{zhang2026compositional} and BuildArena~\cite{xia2026buildarena} have LLMs build machines from standard parts in a physics sandbox, and \citet{chen2026aidsr} build three LLM-designed continuum robots and report a sim-to-real gap. Brick assemblies are generated as stable structures and built by hand or by robots~\cite{pun2025brickgpt,goldberg2024bloxnet,khendry2026idfra,liu2025prompt2product}, simulated as snap-fit assemblies~\cite{wen2026bricksim} and posed as agent benchmarks in simulation~\cite{ghugare2025builderbench,liu2026lmbuild}. None feeds real outcomes back into the design or the simulator (\cref{tab:related}).

\begin{table}[!h]
\centering\footnotesize
\setlength{\tabcolsep}{3pt}
\resizebox{\columnwidth}{!}{%
\begin{tabular}{@{}lcccccc@{}}
\toprule
Work & \shortstack{Functional\\task} & \shortstack{Real\\parts} & \shortstack{Built\\for real} & \shortstack{New\\mechanisms} & \shortstack{Agent\\loop} & \shortstack{Real\\feedback} \\
\midrule
RoboGrammar~\cite{zhao2020robogrammar} & \cmark & \xmark & \xmark & \pmark & \xmark & \xmark \\
RoboMorph~\cite{qiu2024robomorph} & \cmark & \xmark & \xmark & \pmark & \cmark & \xmark \\
BesiegeField~\cite{zhang2026compositional} & \cmark & \xmark & \xmark & \cmark & \cmark & \xmark \\
Text2Robot~\cite{ringel2025text2robot} & \cmark & \pmark & \cmark & \pmark & \xmark & \xmark \\
BrickGPT~\cite{pun2025brickgpt} & \xmark & \cmark & \cmark & \xmark & \xmark & \xmark \\
Blox-Net~\cite{goldberg2024bloxnet} & \xmark & \cmark & \cmark & \xmark & \cmark & \xmark \\
IDfRA~\cite{khendry2026idfra} & \xmark & \cmark & \pmark & \xmark & \cmark & \xmark \\
LMBuild~\cite{liu2026lmbuild} & \cmark & \cmark & \xmark & \cmark & \cmark & \xmark \\
\midrule
\bench (ours) & \cmark & \cmark & \cmark & \cmark & \cmark & \cmark \\
\bottomrule
\end{tabular}}
\caption{\textbf{\bench is the first agentic design benchmark with real feedback into design and simulator.} Functional task: judged on doing a task. New mechanisms: designs beyond tuning parameters. Agent loop: an agent iterates on feedback. Real feedback: real outcomes revise the design or simulator. \pmark: partly.}
\label{tab:related}
\end{table}

\noindent\textbf{Computational robot co-design.}
Evolutionary and learned methods search morphologies and controllers together~\cite{sims1994evolving,cheney2013unshackling,bhatia2021evolution,wang2019neural,yuan2022transform2act,gupta2021embodied,schaff2019jointly,luck2019coadaptation,zhao2020robogrammar}. A few fabricate the result: evolved trusses and actuators~\cite{lipson2000automatic}, designs assembled from living cells~\cite{kriegman2020scalable} and printed legged robots~\cite{ringel2025text2robot}, and graphics work designs printable mechanical characters and walking robots~\cite{coros2013mechanical,megaro2015interactive}. All of them search a fixed, hand-defined design representation, such as a graph grammar, a voxel grid or a parametric body, in a fixed simulator. Our agents are given no such fixed design space: they compose any machine the inventory allows, plus a custom CAD part.

\noindent\textbf{Sim-to-real and simulator learning.}
The reality gap was first met in evolutionary robotics~\cite{jakobi1995noise,koos2013transferability}. Domain randomization~\cite{tobin2017domain,peng2018sim}, system identification~\cite{tan2018sim,ramos2019bayessim,du2021auto,hwangbo2019learning} and simulators adapted from real runs~\cite{hanna2017grounded,chebotar2019closing} address it, as do robots that update self-models from experiments~\cite{bongard2006resilient,kwiatkowski2019selfmodeling}. Recent work reconstructs simulatable twins from video~\cite{jiang2025phystwin,chen2026agenticreal2sim}, lets LLMs set randomization ranges~\cite{ma2024dreureka}, fits simulator parameters by comparing simulated and real videos with VLMs~\cite{qiu2026vid2sid,qiu2026swim2real}, and uses a coding agent to correct a simulator from real trials~\cite{hu2026simex}. These calibrate a model of one fixed robot; in co-design the robot changes at every iteration, so our agents adjust the simulator for designs that did not exist when it was calibrated.

\noindent\textbf{Agents, search and multi-agent systems.}
Coding agents act through tools and agent-computer interfaces~\cite{yao2023react,yang2024sweagent,wang2025openhands} and are evaluated with executable verdicts in software, machine-learning engineering and research~\cite{jimenez2024swebench,liu2024agentbench,chan2025mlebench,lu2024aiscientist}. They reflect on their own failures and accumulate reusable skills~\cite{shinn2023reflexion,madaan2023selfrefine,wang2023voyager}, and, paired with an automatic evaluator, search large program spaces~\cite{romeraparedes2024funsearch,novikov2025alphaevolve,yang2024opro}. Recent analyses of such agents find that they settle into a few idea basins and that resetting their context can restore exploration~\cite{cai2026forkflush,tan2026forgetting}; we measure the same behaviour in physical design and find that, there, more exploration does not predict success. Multi-agent frameworks split software work into roles~\cite{wu2024autogen,hong2024metagpt,qian2024chatdev} or let agents debate~\cite{du2024debate}; our pool harness applies such a role split to design, with one specialist per obstacle and a synthesizer. These agents test their work in seconds; in \bench a real test takes hours, so every iteration is costly and must be chosen well.

%% file: sec/3_method.tex
\section{\bench}
\label{sec:method}
\label{sec:benchmark}
\label{sec:robobench}

\bench turns the engineering loop of build, test and iterate into a benchmark (\cref{fig:benchmark}). \rbBench{} poses the design problem: a robot and its program, built from real parts under constraints, to cross an obstacle or meet an objective. \rbSim{} is where agents iterate: a simulator learned from real measurements and videos. \rbLab{} is where every design that passes a task in simulation is built and run on a physical course. \rbLoop{} connects the two: each real failure goes back to its agent, which fixes its simulator and design.

\paragraph{\rbBench{}: co-design as a benchmark task.}
\label{sec:task}
\label{sec:parts}
\label{sec:tasks}

\noindent\textbf{Design and program.}
An agent receives a course and must submit a robot and a motor program that carries it to the goal within a time limit and keeps it upright there for five seconds. The inventory has three layers (\cref{sec:supp_inventory}): standard brick-compatible parts; electronics (motors, a battery box and up to two IR receivers driving at most four outputs); and one custom part type that the agent writes in CadQuery. A design is a list of placed parts and connections; a program is an open-loop, timed sequence of speed levels per motor.

\noindent\textbf{Problem formulation.}
\label{sec:formulation}
A design is an assembly $d=(P,E)$: a set of placed parts $P=\{(p_i,T_i)\}_{i=1}^{N}$, where each part type $p_i$ comes from the inventory $\mathcal{I}$ or is the agent's printed part type $q$, and $T_i\in SE(3)$ is its pose; and a set of connections $E\subseteq P\times P$ (studs, pins, axles in holes, gear meshes, chains). It must satisfy the inventory counts, $|\{i: p_i=p\}|\le n_p$ for every $p\in\mathcal{I}$, at most four copies of $q$, and the building rules, which a validator and a design review check as $R(d)\in\{0,1\}$. A program $u=\{(t_k,m_k,s_k)\}$ sets motor $m_k$ to speed level $s_k$ at time $t_k$, and a candidate is $x=(d,u)$. A simulator $S_\theta$, with code and physical parameters $\theta$, runs $x$ on obstacle $o$; its verdict $V_\theta(x,o)\in\{0,1\}$ is 1 if $R(d)=1$ and the robot reaches the goal in time and stays upright from three starts (nominal, shifted 1\,mm and 50\,ms late). Co-design asks an agent to find, with at most $B$ calls to $V_\theta$,
\begin{equation}
\begin{gathered}
x^\star\in\{x: V_\theta(x,o)=1\}\quad\text{or}\\
x^\star\in\textstyle\arg\min_{x} f(x)\ \ \text{s.t.}\ V_\theta(x,o)=1,
\end{gathered}
\label{eq:codesign}
\end{equation}
with $f$ the mass or the crossing time. With $Y(x,o)\in\{0,1\}$ the real outcome of a build, the sim-to-real gap is $\Delta_\theta=\Pr[\,Y=0\mid V_\theta=1\,]$; \rbSim{} and \rbLoop{} lower it by changing $\theta$ and $x$.

\noindent\textbf{Obstacles and tasks.}
There are ten obstacles (\cref{fig:courses}), each testing a different physical capability: a ramp tests traction and torque; a wall and steps, climbing mechanisms; a gap and a gap with a drop, spanning and landing; a narrow bridge, straight tracking; a rope, gripping and hanging from a cord; round bumps, clearance and compliance; a seesaw, balance and motor timing; and a U-turn on a tilted plate, turning on a slope. Each comes at three levels that raise the obstacle and tighten the time limit, to find how far agents can go. \taskCross asks for a passing design within 30 simulations; if it fails, \taskPrint adds 30 simulations and one printed part type, used in up to four copies. On every task, \taskLight and \taskFast then ask for the lightest and the fastest passing design, with 50 simulations each, starting from the pass if there is one. \taskAll asks for one level-1 robot with a 3D-printed structure that crosses all ten obstacles, with one program per course and 300 simulations. After each task the agent writes a lessons file that it receives at the next level.

\paragraph{\rbSim{}: learning the simulator.}
\label{sec:simbuild}
\label{sec:sim_learning}

Rigid-body engines model what a part is, not how a small robot of snap-fit parts fails. We therefore let an agent, \opus, learn the simulator before the benchmark.

\noindent\textbf{Initialisation from public part data.}
Designs are simulated in MuJoCo~\cite{todorov2012mujoco}: each part has its own collision geometry, mass and joint model, motors follow torque--speed curves, and every course has a simulated twin. All physical parameters first came from public part data and estimates. We built and filmed the \nInitSimPass{} agent designs that passed in this simulator, and \emph{none} crossed the real course in \nInitTries{} tries. The real runs showed what the model missed: tyres grip the boards less than assumed, long frames sag and joints pull apart, gears lift out of mesh, and load shifts on slopes.

\noindent\textbf{Measurements requested by the agent.}
The agent asked for simple bench tests with the parts themselves, such as tilt tests on the course boards and bricks hung from a joint until it gives (\cref{tab:measure}). Measured values replaced estimates: a friction pin holds 4.5\,N\,mm, not the assumed 24.5.

\noindent\textbf{Learning from videos.}
The filmed runs are split into a training set and a validation set. The agent compares each training video with its simulation at matched times, fits parameters and models missing physical effects (\cref{tab:physics}): stud joints that release and peel under overload, bevel meshes that slip, a supply that sags under load, chains that skip, and per-tyre grip. In the notation above, it seeks $\theta_1=\arg\min_\theta \sum_i \ell\big(S_\theta(x_i,o_i),O_i\big)$ over filmed runs $(x_i,o_i,O_i)$, with $O_i$ the video of candidate $x_i$ on obstacle $o_i$, subject to $\theta$ matching the measurements, where $\ell$ compares real and simulated runs at matched times; the agent itself proposes and judges each change. A change is kept only if it reproduces training outcomes and holds on validation runs.

\noindent\textbf{Validation.}
The learned simulator predicts \nCalRight{} of the \nCalRuns{} labelled runs, including \nHeldOutRight{} of the \nHeldOut{} validation ones (\cref{tab:sim}), and on the benchmark designs, never used for fitting, \nRealRateAll\% of its passes cross the real course (\cref{sec:res_real}). On the same 22 initial designs, of which the public-data simulator passed 15 that all failed for real, it predicts 20 real outcomes, in sample, as it was fitted to their videos.

\paragraph{\rbLab{}: building and testing for real.}
\label{sec:testbed}
\label{sec:verdict}
Every design that passes a task in simulation is built by hand from a generated build package and run, with its program unchanged, on a physical replica of its course. Each design is filmed in up to three tries aligned with its simulation and passes for real if one try reaches the goal. We built and tested \nRobotsBuilt{} robots from about \nPiecesBuilt{} pieces in an estimated \nWorkHours{} hours of assembly, repair and testing (\cref{sec:supp_effort}): the only test that friction, flex, backlash and weight distribution cannot fool. Along the way agents ran \nSimsAgents{} simulations of \nGeoAgents{} distinct robots, and we recorded \nVidTotal{} real videos, each paired with its simulation, all released (\cref{sec:supp_datacard}).

\paragraph{\rbLoop{}: closing the hardware loop.}
\label{sec:loop}
\label{sec:hwloop}
\label{sec:watch}
\label{sec:correction}
\label{sec:redesign}

Every design that passes in simulation is built and tested. Each real failure at levels 1 and 2 becomes a loop task for the agent that designed it, in a fresh session (\cref{fig:cycle}, \cref{sec:supp_loop}).
\emph{Watch:} the agent receives its filmed tries, its simulation and tools that show both at matched times, and writes a failure record: where the runs diverge, which part is involved, and whether the cause is the design, the simulator or the build.
\emph{Adjust the simulator:} for a design $x_0$ with $V_{\theta_1}(x_0,o)=1$ but $Y(x_0,o)=0$, the agent may make a minimal change to its own copy of the simulator (at most two parameters, one new parameter and 150 lines of code), giving $\theta'$, and the change is accepted only if the design then fails in simulation as filmed, $V_{\theta'}(x_0,o)=0$. It may also declare uncertainty ranges $\mathcal{B}$ for parameters it cannot identify.
\emph{Adjust the design:} with the budget of \taskCross and \taskPrint, it searches for a redesign $x_1$ that passes the verdict, $V_{\theta'}(x_1,o)=1$, and a robust vote: it must pass in at least three of five simulations with parameters drawn from $\mathcal{B}$ and perturbed motor speed, torque ($\sigma{=}8\%$) and start phase, and is then built and tested.

%% file: sec/4_experiments.tex
\section{Experiments, Results and Discussion}
\label{sec:experiments}
\label{sec:results}
\label{sec:res_sim}
\label{sec:res_opt}
\label{sec:res_all}

\input{sec/tab_main}

\noindent\textbf{Agents and baselines.}
We evaluate \opus and \sonnet in Claude Code, and \astra and \sol in Codex, all at medium reasoning effort. Each agent runs once per task with the same documents, simulation budgets (not token budgets) and tools: validator, design review, simulator and verdict, contact sheets and videos. Random search and a genetic algorithm (GA) build robots from the full inventory with the same simulator, verdict and budget (\cref{tab:baselines}), and a single LLM call gives one design.

\noindent\textbf{Metrics.}
\emph{Passing}: the share of tasks passed in \taskCross, per level, and with \taskPrint; and the median number of simulations to a pass. \emph{Integration}: the share of the ten obstacles that the \taskAll robot crosses, in simulation and for real. \emph{Optimization}: per task, the best mass (or time) reached by any method in this paper divided by the method's own, so scores are relative to these entrants, summed over the 30 tasks into a \taskLight (or \taskFast) score; higher is better, up to 30, and a task without a passing design adds 0, and these tasks run even where \taskCross failed. \emph{Combined score}: the mean of the pass rate, the \taskAll rate and the two optimization scores, each scaled to 1, times 100 (\opus scores highest under every aggregation, but bootstrap intervals over obstacles overlap~\cite{liang2023helm,agarwal2021precipice}). \emph{Novelty and diversity}: the mean distance of a design from its five nearest search designs in a space of function features such as mass, contacts, wheels, gearing, speed and tipping angle, in units of the search designs' own spacing~\cite{lehman2011novelty}, and the effective number of distinct designs among 100 (Vendi score~\cite{friedman2023vendi}; \cref{sec:supp_novelty}). \emph{Real pass rate}: the share of simulated passes that cross the real course in one of up to three tries.

\paragraph{Agents co-design robots beyond our search baselines.}

\noindent\textbf{Agents solve most obstacles, and search almost none.}
Counting \taskPrint, agents pass \pSimLOne\% of level-1 tasks, \pSimLTwo\% at level~2 and \pSimLThree\% at level~3 (\cref{tab:main}); \opus leads with a median of 8.5 simulations to a pass. Search over the full inventory passes no task, and at most 20\% of level-1 tasks with ten times the budget: without the agents' prior knowledge it builds wheeled and tracked vehicles, not the booms, legs and cord grippers the obstacles call for. A single LLM call, with no validator or simulator feedback, passes at most two tasks: working designs come from iterating, not from prior knowledge alone. The printed part rescues nine failed tasks, mostly at level~3 (\cref{sec:supp_inventive}).

\noindent\textbf{Agents optimize, and speed separates them.}
Asked to make a passing design lighter or faster, agents reach a median of 0.52 to 0.70 of its mass by removing structure the obstacle does not need, and 0.22 to 0.58 of its time by adding drive: larger wheels, different gearing or a second motor (\cref{sec:supp_lightfast}). Lightness converges: on the level-1 bridge, all four lightest designs are the same compact car of 166 to 176\,g. Speed does not, and it separates the agents: \opus is the fastest agent on 23 of 30 tasks and the lightest on 16 of 29, with more than twice any other agent's speed score.

\begin{figure}[t]
\centering
\includegraphics[width=\linewidth]{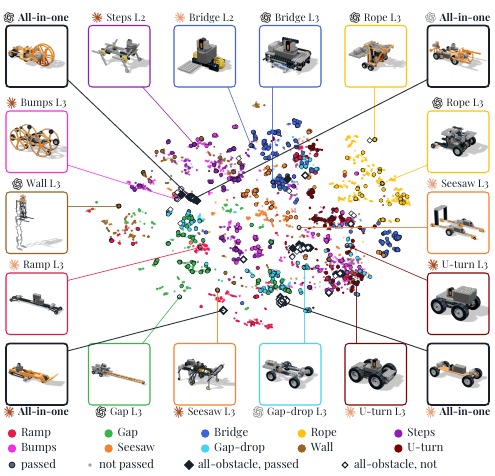}
\caption{\textbf{Designs cluster by obstacle, not by agent.} t-SNE map of the DINOv2~\cite{oquab2024dinov2} features of all simulated designs, coloured by obstacle; outlined: passed; diamonds: \taskAll robots; thumbnails: designs near cluster centres; corners: each agent's \taskAll robot. Per agent in \cref{fig:embedding_agents}.}
\label{fig:embedding}
\end{figure}

\noindent\textbf{Integration is the bottleneck.}
Every agent passes nine or ten level-1 obstacles one at a time, but its single robot for all ten, whose structure must be printed, crosses at most 60\% of them (\cref{tab:main}; robots in \cref{fig:embedding}). The obstacles conflict: the bumps need clearance, the bridge a narrow track and the wall reach, and the rope a light chassis hung from above, which no \taskAll robot combines. On the real courses the four robots cross 20 to 60\% of the obstacles. Splitting the work helps. One specialist agent per obstacle and a synthesizer agent that combines their robots lift \opus, \astra and \sol to 70\% in the best of several single-run harness variants; the first version dropped \opus to 20\% (\cref{sec:supp_pools}), which motivates agentic system design with teams of agents in distinct roles.

\paragraph{How agents design.}
\label{sec:design}
\label{sec:res_behaviour}

\begin{figure}[t]
\centering
\includegraphics[width=\linewidth]{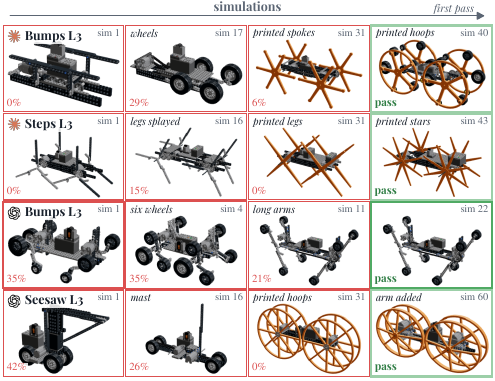}
\caption{\textbf{On hard obstacles, agents keep one machine until the printed part arrives.} Four level-3 runs, first simulation to first plain-run crossing (the official pass, with the nudged starts, can come later). Frame colour: red, fails; green, passes; light green, passes with the printed part, which is available from simulation 31. The labels are our own reading of the renders.}
\label{fig:iter}
\end{figure}

\begin{figure}[t]
\centering
\includegraphics[width=\linewidth]{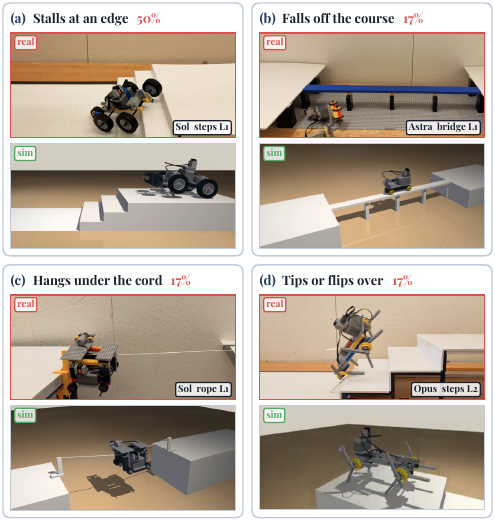}
\caption{\textbf{Real failures stall or fall where the simulated twin crosses.} Failure modes of level-1 and level-2 designs that pass in simulation but fail for real, with each mode's share, a real frame and the simulation at the same time.}
\label{fig:failures}
\end{figure}

\begin{figure*}[t]
\centering
\includegraphics[width=\linewidth]{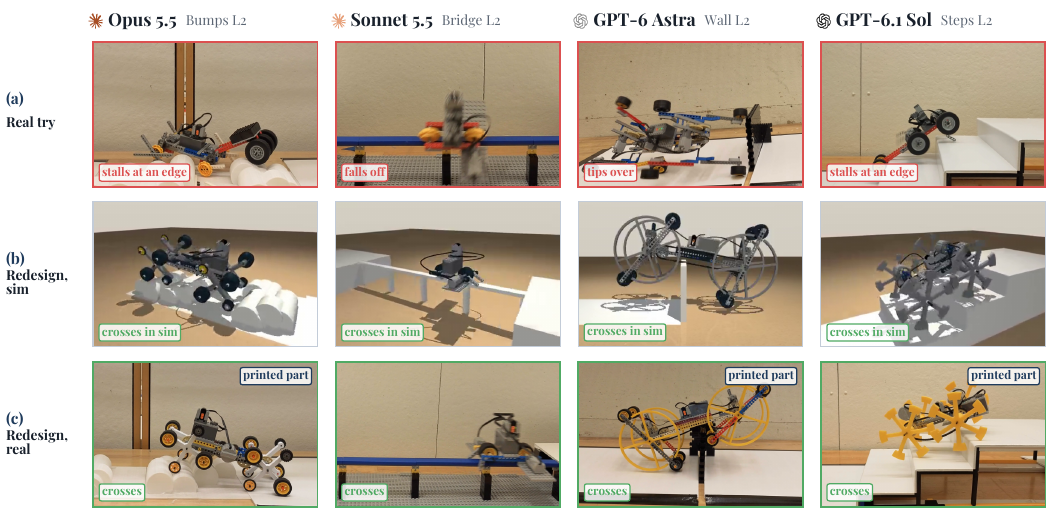}
\caption{\textbf{The loop turns real failures into crossings.} One case per agent: (a)~the original design failing for real, with its failure mode (shares in \cref{fig:failures}); (b)~the redesign crossing in simulation and (c)~on the real course.}
\label{fig:loop}
\end{figure*}

\noindent\textbf{The obstacle, not the agent, decides the machine.}
Designs of different agents converge on a few machine types per obstacle (\cref{fig:embedding}): of each design's 50 nearest neighbours among the other agents' designs, 47\% target the same obstacle, against 10\% by chance. Wheels drive 87 of the 120 final designs and treads 8; legs, rotors and arms (25 designs) appear where the obstacle demands them, and pass 44\% of their tasks against 79\% for wheeled designs. Within those clusters agents still invent (\cref{sec:supp_gallerysec}). By how they work, agents' designs lie 8 to 9 times farther from this search's designs than those lie from each other, and are two to three times as diverse (\cref{tab:main}, \cref{sec:supp_novelty}).

\noindent\textbf{Agents iterate within their first concept.}
Two thirds of the steps between consecutive simulations adjust the same machine and only 4\% change the concept (\cref{fig:iter}, \cref{tab:search}). The first design already holds half or more of the parts of the final passing design. Runs that never pass are not short of budget: they change concept in under 5\% of their steps and spend a median of 33 simulations without improving.

\noindent\textbf{More exploration does not predict success.}
Grouping designs into idea basins, a step into a new basin raises the best progress three to four times as much as a refining step, whether designs are compared by look, parts or function (\cref{sec:supp_basins}). Exploration is front-loaded: agents change machine when the rules change, as when the printed part arrives, not when they stall. Yet trying more ideas does not predict passing, within a task or across agents (\cref{tab:search}; runs that pass early also stop exploring, so this is no evidence that exploration hurts). \opus, the strongest agent, tries the fewest ideas before passing, passes with its first idea in 57\% of its runs, and its new ideas beat its best design 43\% of the time, against about 15\% for the others; \sonnet tries the most ideas and passes the fewest tasks. The gap between agents is consistent with built-in engineering knowledge: the strongest agent starts near a working concept.

\noindent\textbf{Agents build more than they need.}
Agents optimize for passing, not building: the median passing design has 114 pieces, 2.6 times as many as the leanest passing design for the same task; a quarter of all pieces are pins, bushes and connectors whose only role is to join other parts. At our measured building rate of 43\,s per piece, the median design takes about 80 minutes to assemble by hand, and a design-for-assembly analysis~\cite{boothroyd2010dfma} finds 20 insertions blocked by parts placed earlier (\cref{sec:supp_assembly}). No task rewards assemblability, so agents never optimize it.

\paragraph{Real-world tests and closing the sim-to-real gap.}
\label{sec:res_real}
\label{sec:res_loop}

\noindent\textbf{Most simulated passes hold, fewer at harder levels.}
Of the designs that pass in the learned simulator, \nRealRateAll\% cross the real course: \pRealLOne\% at level~1, \pRealLTwo\% at level~2 and \nRealRateLThree\% at level~3 (\cref{tab:real}); per try, 68\% of level-1 and level-2 tries cross. Most failures stall at a step, edge or wall, fall off, tip over or hang under the cord (\cref{fig:failures}), from build asymmetry, the course model, drivetrains built too tight, structural flex, timing and parts working loose. Some causes are outside a deterministic simulation of the nominal design.

\noindent\textbf{The hardware loop removes most real failures.}
Every real failure at levels 1 and 2 went back to its agent. One round of the loop removes \pLoopCut\% of these failures (from \nRealFail{} to 5), and the share of simulated passes that cross for real rises from \nRealRate\% to \nAfterLoopRate\%, and to \nBestAfterLoop\% for \sonnet and \astra (\cref{tab:loop_agents}). Only one failure found no redesign that passed the robust vote, and four redesigns still fail for real. Agents redesign after seeing the failure on the same course, so this measures recovery, not transfer of new designs.

\noindent\textbf{Agents redesign, and correct the simulator where the video shows the cause.}
Redesigns change the mechanism that failed rather than only the program, and six use a printed part; they include \opus's worm-driven whegs, \sonnet's cable car, \astra's hoops and \sol's cam wheels (\cref{fig:loop}). They follow the failure (\cref{tab:redesign}): when a robot fails partway, at a step, ridge or plank, agents usually replace the mechanism (7 of 9 cases), and all seven cross for real; when it barely starts, they usually adjust the same machine (7 of 9), and five of those seven cross (\cref{fig:loop}). Agents also make simulator changes consistent with the video (\cref{fig:cycle}): \opus watched its spoked wheels spin against a bump ridge and lowered the friction of hard plastic on the bump rods from 0.39 to 0.30. Where the cause cannot be seen, agents declare parameter ranges, most often on motor torque, friction and cord tension, and the redesign must pass in simulation across these ranges (\cref{sec:supp_loop}).

\paragraph{Discussion.}
\label{sec:discussion}
Three lessons stand out. First, the reality gap shrinks only with real data: with the learned simulator, 73\% of level-1 and level-2 simulated passes of the new benchmark designs cross for real, where none of the initial designs did (\cref{tab:sim}), and redesigns from per-failure videos raise this to 93\% (\cref{tab:loop_agents}). Second, integration, not single obstacles, limits agents: every agent passes nine or ten level-1 obstacles alone but its single robot crosses at most six (\cref{tab:main}); in an exploratory study, splitting the work among specialist and synthesizer agents lifts three agents to seven in the best variant (\cref{tab:harness,tab:supp_reqfirst}), pointing to teams of agents with distinct roles. Third, agents must still analyse and iterate, since a single call without the simulator passes almost nothing (\cref{tab:main}), but more exploration does not predict better designs (\cref{tab:search,tab:basins_agents}); an agent's built-in engineering knowledge seems its biggest edge, so harnesses that help it choose and screen concepts may beat forced exploration.

%% file: sec/tab_main.tex
\begin{table*}[t]\centering\footnotesize
\caption{\textbf{Agents pass most tasks, optimize and invent; search over the full inventory passes nothing.} \taskCross: tasks passed per level; +\textsc{Print}: with \taskPrint, all levels. Sims: median simulations to a pass. \taskAll: obstacles crossed by one robot. \taskLight, \taskFast: mean mass and time over shared level-1 tasks, and scores (of 30). Score: combined (of 100). Novelty: distance of the designs from the nearest search designs in function space, in units of the search designs' own spacing (search $\approx$1). Diversity: effective number of distinct designs among 100 (Vendi). $\uparrow$/$\downarrow$: higher/lower is better; colour: red (worst), yellow, green (best) per column. Full-inventory search: random search and a GA, 30 simulations per task, not run on \taskAll, \taskLight or \taskFast (\cref{tab:baselines}); single LLM call: best model (\cref{tab:sim_results}).}
\label{tab:main}
\definecolor{heatG}{HTML}{50AA64}\definecolor{heatY}{HTML}{F2C744}\definecolor{heatR}{HTML}{DC5050}
\setlength{\tabcolsep}{3.4pt}
\resizebox{\linewidth}{!}{%
\begin{tabular}{@{}lccccccccccccccc@{}}
\toprule
 & \multicolumn{7}{c}{Passing} & \multicolumn{4}{c}{Optimization} & Overall & \multicolumn{2}{c}{Novelty \& diversity} \\
\cmidrule(lr){2-8}\cmidrule(lr){9-12}\cmidrule(lr){13-13}\cmidrule(l){14-15}
 & \multicolumn{3}{c}{\taskCross (\%)} & +\textsc{Print} & Sims & \multicolumn{2}{c}{\taskAll (\%)} & \multicolumn{2}{c}{\taskLight} & \multicolumn{2}{c}{\taskFast} & Score & Novelty & Diversity \\
\cmidrule(lr){2-4}\cmidrule(lr){7-8}\cmidrule(lr){9-10}\cmidrule(lr){11-12}
Method & L1\,$\uparrow$ & L2\,$\uparrow$ & L3\,$\uparrow$ & (\%)\,$\uparrow$ & (med.)\,$\downarrow$ & sim\,$\uparrow$ & real\,$\uparrow$ & mass (g)\,$\downarrow$ & score\,$\uparrow$ & time (s)\,$\downarrow$ & score\,$\uparrow$ & (/100)\,$\uparrow$ & (function)\,$\uparrow$ & (Vendi)\,$\uparrow$ \\
\midrule
Full-inventory search & \cellcolor{heatY!0!heatR!40!white}0 & \cellcolor{heatY!0!heatR!40!white}0 & \cellcolor{heatY!0!heatR!40!white}0 & \cellcolor{heatY!0!heatR!40!white}0 & -- & -- & -- & -- & -- & -- & -- & \cellcolor{heatY!0!heatR!40!white}0.0 & \cellcolor{heatY!0!heatR!40!white}1.0 & \cellcolor{heatY!0!heatR!40!white}3.4 \\
Single LLM call (\opus) & \cellcolor{heatY!40!heatR!40!white}20 & \cellcolor{heatY!0!heatR!40!white}0 & \cellcolor{heatY!0!heatR!40!white}0 & \cellcolor{heatY!18!heatR!40!white}7 & -- & \cellcolor{heatY!0!heatR!40!white}0 & -- & -- & \cellcolor{heatY!0!heatR!40!white}0.0 & -- & \cellcolor{heatY!0!heatR!40!white}0.0 & \cellcolor{heatY!4!heatR!40!white}1.7 & -- & -- \\
\midrule
\opusL & \cellcolor{heatG!100!heatY!40!white}100 & \cellcolor{heatG!100!heatY!40!white}80 & \cellcolor{heatG!100!heatY!40!white}40 & \cellcolor{heatG!100!heatY!40!white}80 & \cellcolor{heatG!100!heatY!40!white}8.5 & \cellcolor{heatG!66!heatY!40!white}50 & \cellcolor{heatG!100!heatY!40!white}60 & \cellcolor{heatG!100!heatY!40!white}255 & \cellcolor{heatG!100!heatY!40!white}22.6 & \cellcolor{heatG!100!heatY!40!white}1.22 & \cellcolor{heatG!100!heatY!40!white}27.1 & \cellcolor{heatG!100!heatY!40!white}73.9 & \cellcolor{heatG!100!heatY!40!white}9.3 & \cellcolor{heatG!100!heatY!40!white}11.7 \\
\sonnetL & \cellcolor{heatG!80!heatY!40!white}90 & \cellcolor{heatG!0!heatY!40!white}40 & \cellcolor{heatG!0!heatY!40!white}20 & \cellcolor{heatG!32!heatY!40!white}53 & \cellcolor{heatY!0!heatR!40!white}12 & \cellcolor{heatG!0!heatY!40!white}30 & \cellcolor{heatY!0!heatR!40!white}20 & \cellcolor{heatG!18!heatY!40!white}283 & \cellcolor{heatG!52!heatY!40!white}17.2 & \cellcolor{heatG!20!heatY!40!white}2.19 & \cellcolor{heatY!82!heatR!40!white}11.0 & \cellcolor{heatG!20!heatY!40!white}44.3 & \cellcolor{heatG!72!heatY!40!white}8.1 & \cellcolor{heatG!16!heatY!40!white}8.2 \\
\astraL & \cellcolor{heatG!80!heatY!40!white}90 & \cellcolor{heatG!100!heatY!40!white}80 & \cellcolor{heatG!50!heatY!40!white}30 & \cellcolor{heatG!92!heatY!40!white}77 & \cellcolor{heatY!86!heatR!40!white}10.5 & \cellcolor{heatG!100!heatY!40!white}60 & \cellcolor{heatG!50!heatY!40!white}50 & \cellcolor{heatY!0!heatR!40!white}324 & \cellcolor{heatG!60!heatY!40!white}18.1 & \cellcolor{heatY!0!heatR!40!white}3.65 & \cellcolor{heatY!74!heatR!40!white}10.1 & \cellcolor{heatG!56!heatY!40!white}57.6 & \cellcolor{heatG!100!heatY!40!white}9.3 & \cellcolor{heatG!64!heatY!40!white}10.2 \\
\solL & \cellcolor{heatG!100!heatY!40!white}100 & \cellcolor{heatG!50!heatY!40!white}60 & \cellcolor{heatG!100!heatY!40!white}40 & \cellcolor{heatG!92!heatY!40!white}77 & \cellcolor{heatG!14!heatY!40!white}10 & \cellcolor{heatG!66!heatY!40!white}50 & \cellcolor{heatG!100!heatY!40!white}60 & \cellcolor{heatY!12!heatR!40!white}320 & \cellcolor{heatG!64!heatY!40!white}18.5 & \cellcolor{heatY!42!heatR!40!white}3.14 & \cellcolor{heatY!70!heatR!40!white}9.6 & \cellcolor{heatG!50!heatY!40!white}55.1 & \cellcolor{heatG!88!heatY!40!white}8.8 & \cellcolor{heatG!46!heatY!40!white}9.5 \\
\bottomrule
\end{tabular}}
\end{table*}

%% file: sec/5_conclusion.tex
\section{Conclusion}
\label{sec:limits}
\label{sec:conclusion}

\noindent\textbf{Limitations and future work.}
The simulator is fitted from limited observations, a fixed camera and bench tests without force or contact data, so some parameters are not identifiable and other settings might fit as well. Though not shown to be physically exact, it is validated on validation runs and real tests, a physics-grounded environment in which future models are scored on the design tasks over all 30 obstacle levels and on the \taskAll task. Each loop correction is fitted to one failure in the agent's own simulator copy and not checked against other filmed runs; the loop's gains come mainly from redesign after watching the failure, since we do not isolate the corrections' share. Assembly errors and small differences between simulated and real obstacles also cause real failures. \taskLight and \taskFast designs are not built; entries may also report real crossings of level-1 passes on a real-world leaderboard. Future work (\cref{sec:supp_limits}): robots that build and test designs as a tool for agents; sensing and feedback co-designed by agents, as robots now run open loop; one simulator for all failures; and requirement-driven mechatronic systems from specification to prototype.

\noindent\textbf{Conclusion.}
We presented \bench, a benchmark in which coding agents design robots from real parts, and every simulated task pass is built, tested and fed back into the next iteration of simulator and design. With a simulator learned by an agent, 73\% of level-1 and level-2 simulated passes hold for real, and after agents redesign from videos of their failures, 93\% do. They invent mechanisms that our search baselines do not reach, yet falter on hard obstacles and when one machine must meet every requirement. Real-world testing is part of the design loop, not a final check, and benchmark simulators must be checked against real tests to measure whether agents can engineer. The same loop of design, simulation and real tests extends to mechatronic products: from a specification of function, cost, mass and safety, an agent picks parts, electronics and control, checks them in a calibrated simulator and refines them from bench tests, as our agents do here for robots. \bench is a first step toward agents that turn requirements into working prototypes.

%% file: sec/8_llm_usage.tex
\section*{LLM Usage Statement}
LLMs are the object of study: the four coding agents evaluated here (\opus and \sonnet in Claude Code, \astra and \sol in Codex) design every robot and, in the hardware loop, propose simulator changes that are accepted only after an automatic check. Separately, the authors used Claude as a coding assistant to generate code for the harness, simulator tooling, build-package exporters, and analysis scripts; the authors reviewed, edited, or approved this code. For writing and editing text, related work and citations, experimental design and execution and analysis of results, figures and visual material, and data preparation and annotation, LLM use was limited to editing or refining author-produced content. No LLMs were used to develop the research ideas, hypotheses, claims, or conclusions. All physical robots were built and tested by the authors, and every real-world verdict was determined by the authors from the videos. The authors verified all reported numbers against the logged data and take full responsibility for the paper’s content, including all citations.

%% file: sec/X_suppl.tex
\setcounter{topnumber}{3}\setcounter{dbltopnumber}{3}\setcounter{totalnumber}{5}
\renewcommand{\topfraction}{0.95}\renewcommand{\dbltopfraction}{0.95}\renewcommand{\textfraction}{0.05}
\renewcommand{\floatpagefraction}{0.8}\renewcommand{\dblfloatpagefraction}{0.8}

\FloatBarrier
\section{Design gallery}
\label{sec:supp_gallerysec}

\FloatBarrier
\subsection{Design gallery}

\label{sec:supp_gallery}

\begin{figure}[H]
\centering
\includegraphics[width=\linewidth]{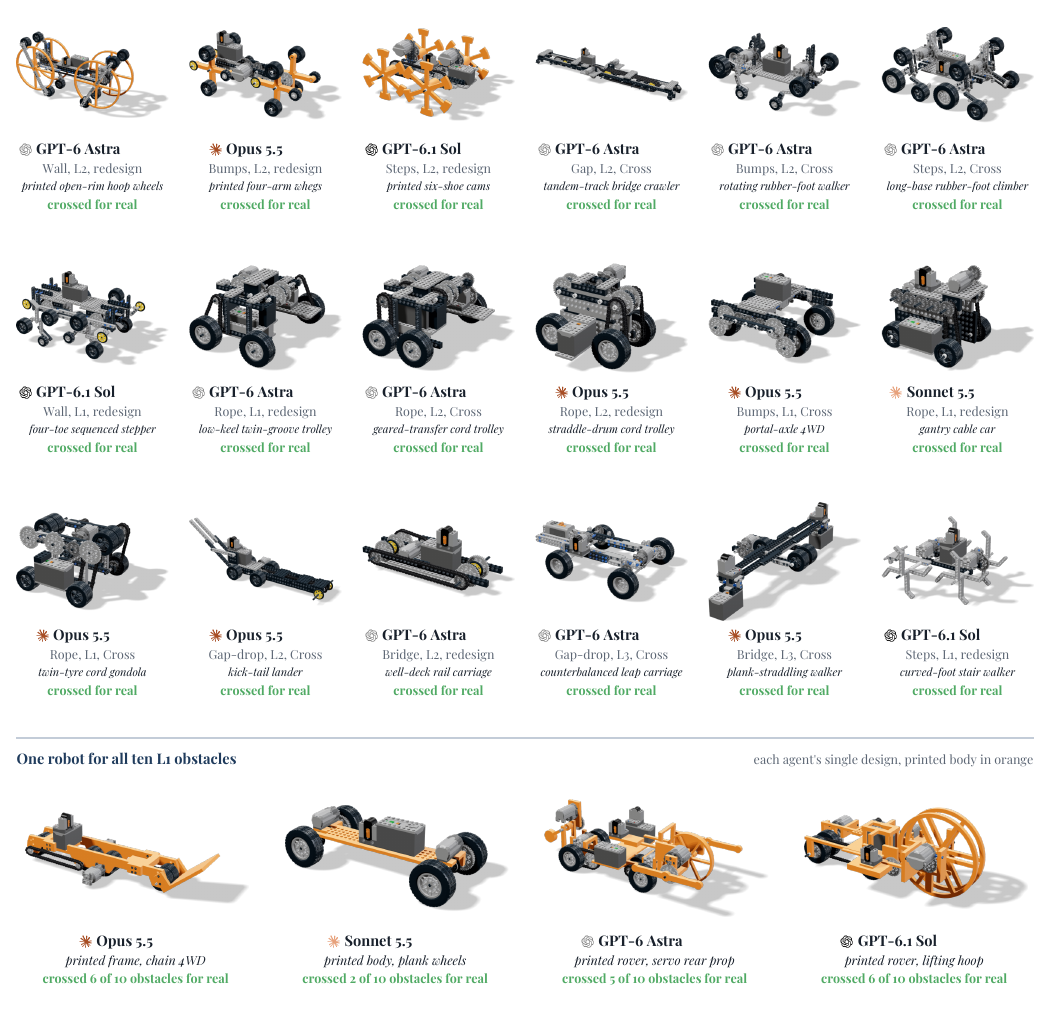}
\caption{\textbf{Robots that cross the real course range from trolleys and girders to whegs, cam climbers and walkers.} Benchmark designs and loop redesigns (printed parts in orange) with agent, obstacle, level and mechanism. Bottom: each agent's \taskAll robot.}
\label{fig:gallery}
\end{figure}

\begin{figure}[H]
\centering
\includegraphics[width=\linewidth]{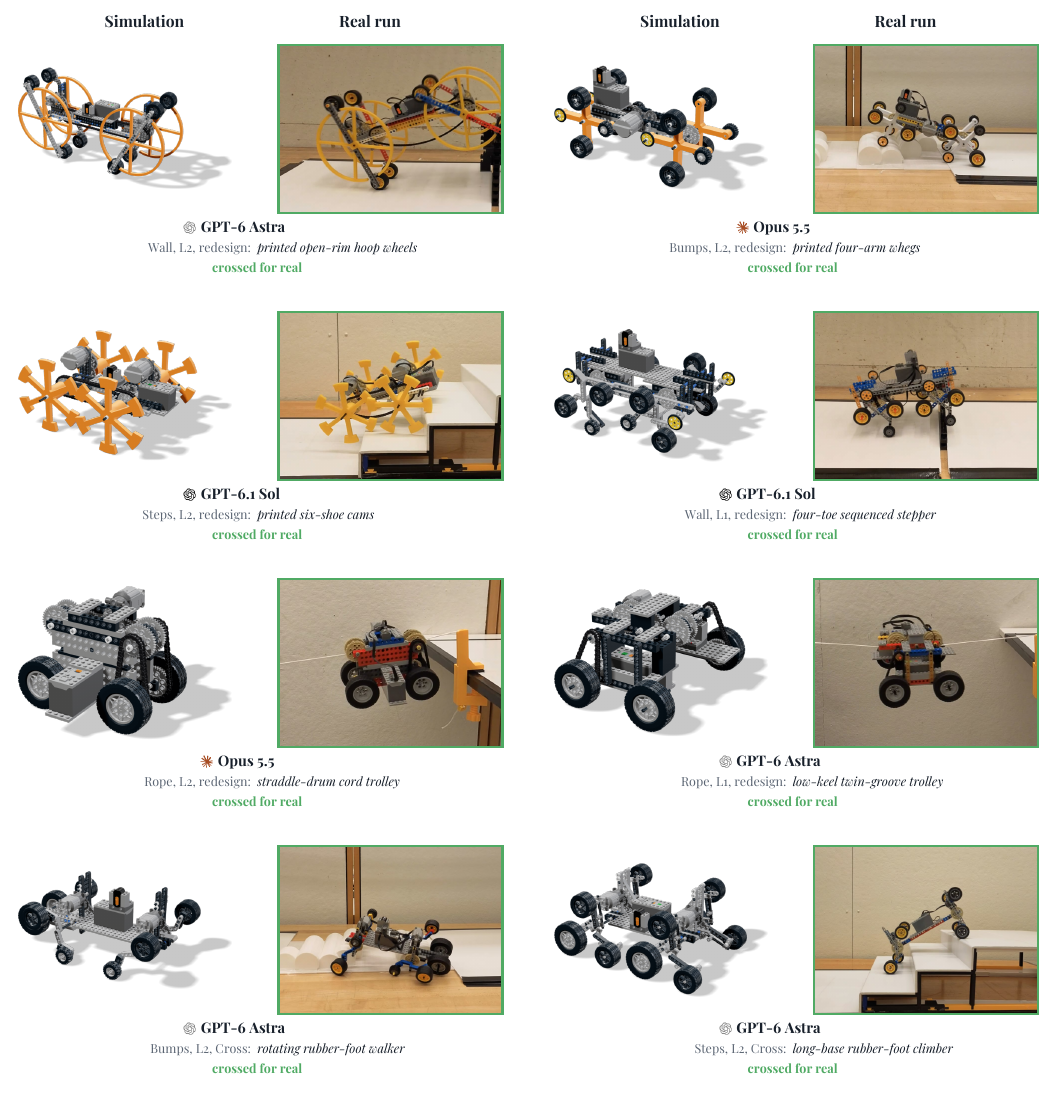}
\caption{\textbf{The designs as simulated and as built.} Render and real crossing frame for eight robots that crossed the real course.}
\label{fig:gallery_real}
\end{figure}

\FloatBarrier
\subsection{Inventive designs}
\label{sec:supp_inventive}
Screened from all benchmark task trajectories for large printed parts (up to the 256\,mm print bed), unusual part mixes (legs, servos, racks, booms) and high levels; 19 of 24 are from level 3; tags come from the agent's own descriptions.

\begin{figure}[H]
\centering
\includegraphics[width=\linewidth]{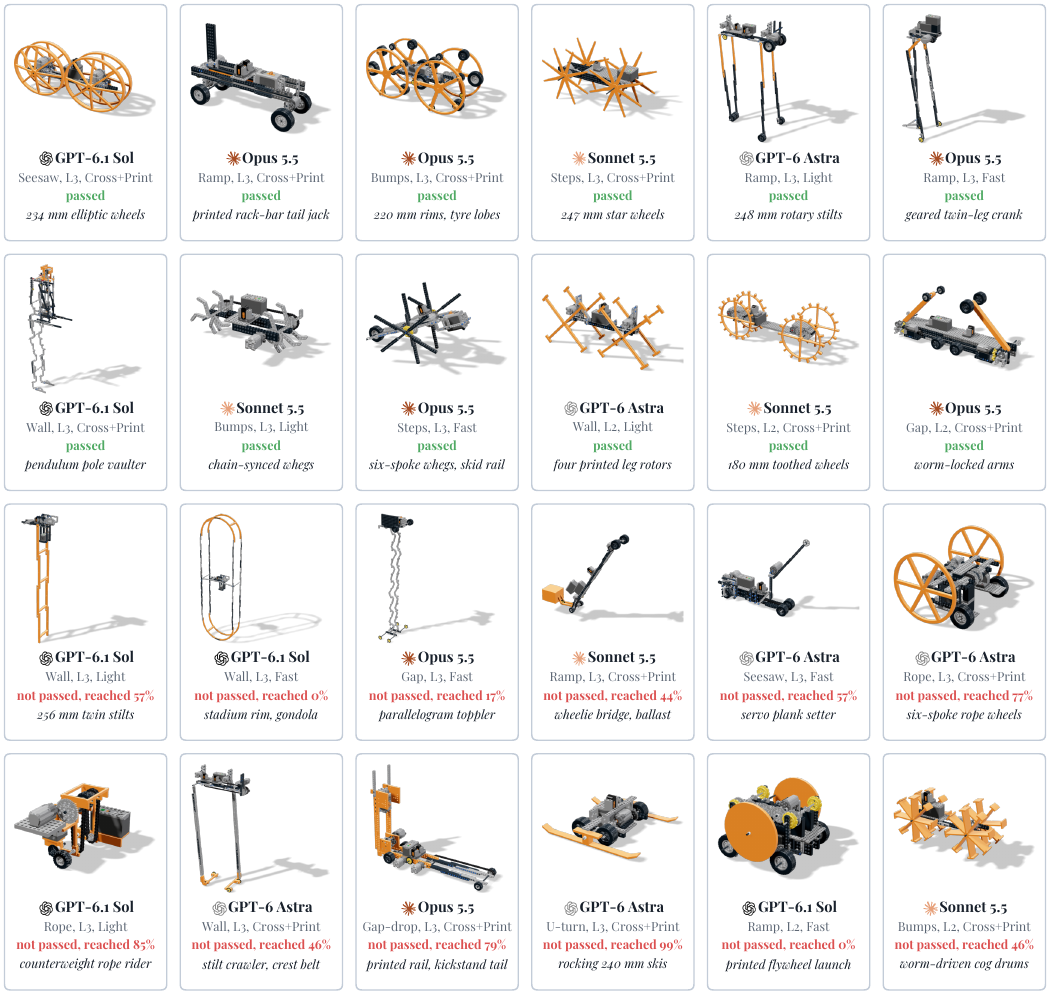}
\caption{\textbf{On the hardest obstacles agents invent printed wheels, stilts, legs, vaulters and booms, though many never pass.} Top two rows passed in simulation, bottom two did not (best progress). Cards: agent, obstacle, level and task; printed parts in orange.}
\label{fig:inventive}
\end{figure}

\FloatBarrier
\section{How agents design}
\label{sec:supp_agentdesign}

\FloatBarrier
\subsection{How agents search}
\label{sec:supp_search}

Step classes, relative to the previous simulated design: \emph{program-only} keeps the same parts at the same poses; a \emph{concept change} replaces at least a third of the parts (multiset Jaccard distance $\geq0.5$) \emph{and} changes the look (DINOv2 cosine distance $\geq0.10$); everything else is a \emph{local adjustment}. Agents differ in style rather than in the pattern of local refinement: \sonnet only reprograms the robot in 44\% of its steps, and \opus changes concept in 2.6\%; \astra and \sol do so in 5.7\% and 5.1\%, \astra mostly before its first pass (8.5\% of those steps) and \sol also after it (7.5\%).

\begin{figure}[H]
\centering
\includegraphics[width=\linewidth]{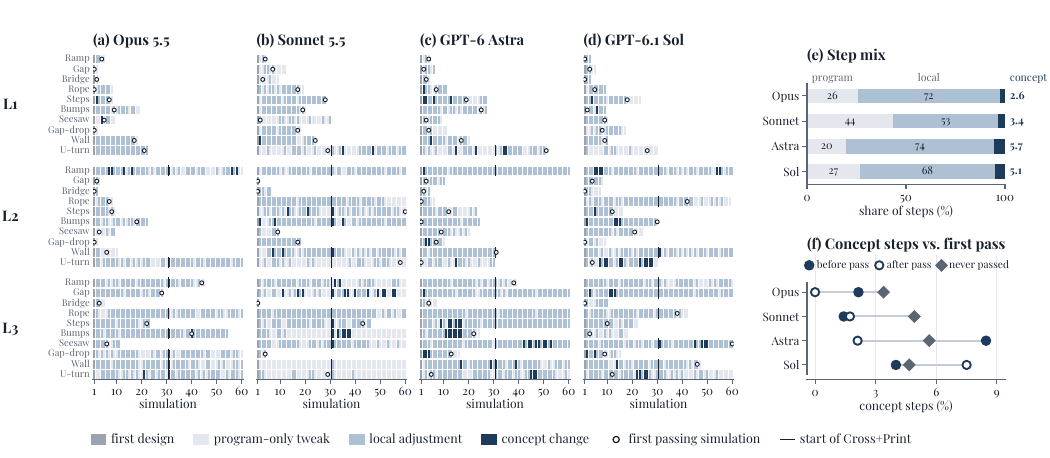}
\caption{\textbf{Agents refine their first concept; under 6\% of steps change concept.} (a--d)~Every simulation of every task (rows: level and obstacle; tick: start of \taskPrint), coloured by the change from the previous design (\cref{sec:supp_search}); circles: first pass. (e)~Share of steps per class. (f)~Concept changes before and after the first pass and in failed runs.}
\label{fig:iterchange}
\end{figure}

\begin{figure}[H]
\centering
\includegraphics[width=0.6\linewidth]{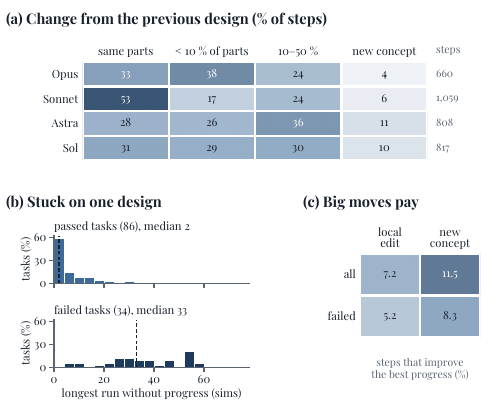}
\caption{\textbf{Agents mostly make small edits, stall on failed tasks, and big moves pay off more often.} (a)~Step mix per agent. (b)~Longest run without improving (dashed: median). (c)~Share of steps that improve, local edits vs.\ new concepts.}
\label{fig:search}
\end{figure}

\begin{figure}[H]
\centering
\includegraphics[width=0.45\linewidth]{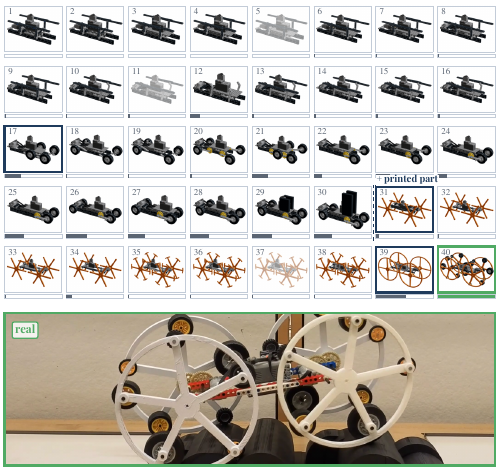}
\caption{\textbf{All 40 simulations of the run in \cref{fig:devprocess}.} Bar: progress; faded: program-only change; navy frame: new machine; green: first pass; photo: the built robot crossing.}
\label{fig:devprocess_all}
\end{figure}

\FloatBarrier
\begin{figure}[H]
\centering
\includegraphics[width=\linewidth]{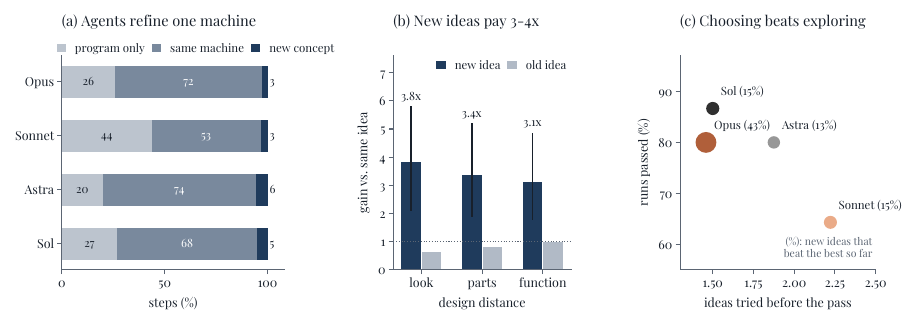}
\caption{\textbf{Agents refine one machine, although new ideas pay more, and exploring more does not predict success.} (a)~Steps between consecutive simulated designs. (b)~Rise of the best progress for a step into a new idea basin, or back to an old one, relative to a step within the same basin, under three design distances (95\% intervals). (c)~Per agent, ideas tried before the first pass against runs passed; marker size and (\%): new ideas that beat the run's best so far.}
\label{fig:searchmain}
\end{figure}

\FloatBarrier
\subsection{Commitment to a first concept}
\label{sec:supp_commit}
\begin{figure}[H]
\centering
\includegraphics[width=0.6\linewidth]{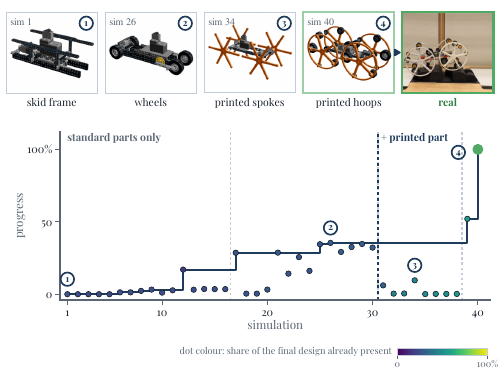}
\caption{\textbf{One run from first simulation to real crossing} (level-3 bumps; simulations shown up to the first plain-run crossing, while the official pass, which also needs the nudged starts, comes later, \cref{tab:expanded}). Most of the final design is present from the first simulation. Each circle is one simulation, placed at the progress the robot made and coloured by the share of the final design already present. The line shows the best progress so far. All 40 runs are in \cref{fig:devprocess_all}.}
\label{fig:devprocess}
\end{figure}

The first design already contains a median of 50 to 69\% of the parts of the final passing design, depending on the level, and the design just before the pass contains nearly all of them (\cref{fig:devprocess,fig:iter}). Runs that never pass change concept in only 4.7\% of their steps, although large moves raise the best progress more often (11.5\% against 7.2\% of steps).

\FloatBarrier
\begin{table}[H]
\centering\footnotesize\setlength{\tabcolsep}{3pt}
\caption{\textbf{The best agent starts closer to a working design rather than exploring more.} Per agent: share of steps between consecutive simulations that keep the same machine or change the concept; ideas (basins) tried before the first pass; runs that pass in their first idea; new ideas that beat the run's best so far; tasks passed (\taskCross or \taskPrint, as in \cref{tab:main}). A new idea pays 3.1 to 3.8 times more than a refining step under every design distance (\cref{sec:supp_basins}).}
\label{tab:search}
\begin{tabular}{@{}lcccccc@{}}
\toprule
 & \multicolumn{2}{c}{Steps (\%)} & Ideas to & First idea & New ideas & Passed \\
\cmidrule(lr){2-3}
Agent & same & concept & pass & passes (\%) & pay (\%) & (\%) \\
\midrule
\opusL & 72 & 3 & \textbf{1.5} & \textbf{57} & \textbf{43} & \textbf{80} \\
\sonnetL & 53 & 3 & 2.2 & 25 & 15 & 53 \\
\astraL & 74 & 6 & 1.9 & 40 & 13 & 77 \\
\solL & 68 & 5 & \textbf{1.5} & 53 & 15 & 77 \\
\bottomrule
\end{tabular}
\end{table}

\FloatBarrier
\subsection{Idea basins}
\label{sec:supp_basins}
Following~\cite{cai2026forkflush,tan2026forgetting}, each task's designs (all agents pooled) are described by DINOv2 feature and log part-type counts, reduced to 16 dimensions (PCA) and clustered with $k$-means ($k\in\{3,\dots,8\}$ by silhouette). A step is \emph{new} if it enters an unvisited basin, a \emph{return} if it re-enters an earlier one, else \emph{same basin}; its gain is the rise in best progress (capped at 1).

The new-basin gain holds across clustering variants (2.5--4.6 times) and after a stall of at least five simulations (4.2 times). None of the 22 failed runs enters a basin in which any agent passed the task (1.07 agents per passing basin). Exploration is front-loaded: about 6\% of steps enter a new basin in simulations 1--10 and under 2\% in 21--30; it rises to about 5\% in 31--45, when the printed part becomes available, and nearly stops afterwards, so agents change machine when the rules change, not when they stall. \sonnet explores most (2.4 basins per run) but gains least per new basin; \opus explores least (half its runs never leave) but its new-basin steps pay most (4.7 times a same-basin step).

\paragraph{Other design distances.}
\emph{Parts}: weighted Jaccard distance between part multisets, $1-\sum_p\min(n_p,n'_p)/\sum_p\max(n_p,n'_p)$, average linkage. \emph{Function}: the free-body and geometry features of \cref{sec:supp_novelty} plus gear and worm meshes and an indicator where a quantity does not apply, scale features log-transformed, standardised per task, reduced to 8 dimensions before $k$-means. Median adjusted Rand index per task: 0.80 look--parts, 0.67 look--function, 0.65 parts--function.

\begin{table}[H]
\centering\footnotesize
\caption{\textbf{New basins pay under every design distance.} Idea basins over the 120 counted runs with three distances. Gain ratio: mean rise of the best progress for a step into a new basin (or back to an old one) divided by that of a same-basin step, with 95\% bootstrap intervals over runs. Concept steps: share of the concept changes of \cref{fig:iterchange} that switch basin.}
\label{tab:basins_alt}
\begin{tabular}{@{}lccccccc@{}}
\toprule
Distance & silhouette & basins/run & never leave & steps switching & new-basin gain & return gain & concept steps \\
\midrule
Look + parts (main) & 0.64 & 2.0 & 34\% & 6.4\% & 3.8$\times$ (2.1--5.8) & 0.6$\times$ & 70\% \\
Parts (Jaccard) & 0.66 & 2.0 & 42\% & 7.1\% & 3.4$\times$ (1.9--5.2) & 0.8$\times$ & 86\% \\
Function & 0.71 & 2.5 & 25\% & 11.6\% & 3.1$\times$ (1.8--4.9) & 1.0$\times$ & 76\% \\
\bottomrule
\end{tabular}
\end{table}

\paragraph{Exploring more or choosing better?}
On the 52 runs of tasks with both passes and failures, basins visited in the first ten simulations do not predict a pass (Spearman $-0.04$, $-0.14$, $-0.09$; none significant). Agents almost never enter a basin where another agent passed (3 to 21\% of first basins and switches).

\begin{table}[H]
\centering\footnotesize
\caption{\textbf{More exploration does not predict success.} Per agent, under the three design distances of \cref{tab:basins_alt} (look + parts / parts / function), over the 118 runs with at least two simulations. Passed: runs with at least one passing simulation in phase A or X (before the final design review and re-scoring, so above the official pass rates of \cref{tab:main}). First basin: share of runs that pass in the basin of their first design. Basins to pass: basins visited up to the first pass, in passed runs. Switches that pay: share of steps into a new basin that raise the run's best progress.}
\label{tab:basins_agents}
\begin{tabular}{@{}lcccc@{}}
\toprule
Agent & Passed & First basin (\%) & Basins to pass & Switches that pay (\%) \\
\midrule
\opus & 80\% & \textbf{57} / \textbf{63} / 47 & \textbf{1.5} / \textbf{1.3} / \textbf{1.6} & \textbf{43} / \textbf{35} / \textbf{33} \\
\sonnet & 64\% & 25 / 32 / 29 & 2.2 / 1.9 / 2.3 & 15 / 12 / 23 \\
\astra & 80\% & 40 / 43 / 43 & 1.9 / 2.0 / 2.1 & 13 / 20 / 14 \\
\sol & \textbf{87\%} & 53 / 50 / \textbf{50} & \textbf{1.5} / 1.7 / 1.9 & 15 / 23 / 14 \\
\midrule
All & 78\% & 44 / 47 / 42 & 1.7 / 1.7 / 1.9 & 20 / 21 / 20 \\
\bottomrule
\end{tabular}
\end{table}

\begin{figure}[H]
\centering
\includegraphics[width=\linewidth]{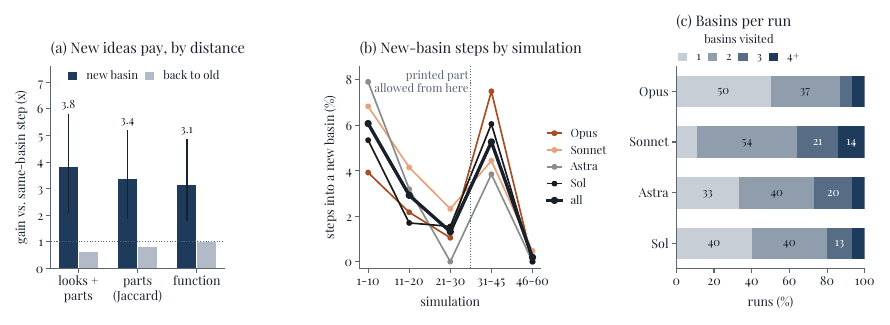}
\caption{\textbf{New ideas pay, but agents stop exploring early.} Idea basins over the 120 counted runs. (a)~Gain of a step into a new basin, or back to an old one, relative to a same-basin step, under the three design distances of \cref{tab:basins_alt}, with 95\% bootstrap intervals; dotted line: no difference. (b)~Share of steps that enter a new basin, by simulation; the printed part is allowed from simulation 31. (c)~Basins visited per run, per agent. (b) and (c) use the look-and-parts distance.}
\label{fig:basins}
\end{figure}

\FloatBarrier
\subsection{Agent behaviour: details}
\label{sec:supp_behaviour}
A \emph{step} is a pair of consecutive simulations of one task (task A, then round X); its size is the multiset Jaccard distance between the part-type counts of the two designs.
\begin{table}[H]\centering\footnotesize\setlength{\tabcolsep}{2.6pt}
\caption{\textbf{Search behaviour per agent.} Step mix over all steps; columns bin steps by the share of parts changed (concept: over half, without the look criterion of \cref{sec:supp_search}, hence above the 4\% of \cref{sec:design}); plateau: median longest run without improving the best progress; switches: new concepts per ten steps (passed / failed tasks).}\label{tab:supp_search}
\begin{tabular}{@{}lccccccc@{}}\toprule
Agent & steps & same & $<$10\% & 10--50\% & concept & plateau P/F & switches P/F\\\midrule
\opus & 660 & 33\% & 38\% & 24\% & 4.4\% & 2 / 39.5 & 0.46 / 0.42\\
\sonnet & 1059 & 53\% & 17\% & 24\% & 5.9\% & 3.5 / 42.5 & 0.39 / 0.65\\
\astra & 808 & 28\% & 26\% & 36\% & 11.0\% & 2 / 28 & 1.14 / 1.06\\
\sol & 817 & 31\% & 29\% & 30\% & 9.7\% & 3 / 25 & 1.04 / 0.89\\
\midrule
All & 3344 & 38\% & 26\% & 28\% & 7.8\% & 2 / 33 & 0.83 / 0.74\\
\bottomrule\end{tabular}\end{table}
\begin{table}[H]\centering\footnotesize\setlength{\tabcolsep}{2.4pt}
\caption{\textbf{Final designs per agent} (accepted, or best attempt; 30 tasks each). Locomotion counts; share of designs with chains, gears and a printed part; medians of mass and parts; mean motors; part types used.}\label{tab:supp_designs}
\begin{tabular}{@{}lccccccccc@{}}\toprule
Agent & wheels & tracks & other & chain & gears & printed & mass (g) & parts & motors\\\midrule
\opus & 20 & 2 & 8 & 30\% & 100\% & 23\% & 486 & 135 & 1.9\\
\sonnet & 22 & 1 & 7 & 33\% & 93\% & 27\% & 449 & 73.5 & 1.7\\
\astra & 21 & 5 & 4 & 40\% & 97\% & 30\% & 523 & 142 & 1.9\\
\sol & 24 & 0 & 6 & 57\% & 100\% & 30\% & 475 & 140 & 1.8\\
\bottomrule\end{tabular}\end{table}
\paragraph{Most used parts.} Chain links, friction pins, half bushes, long friction pins and 1$\times$16 beams are among every agent's six most used parts; 110 part types appear in final designs (131 of the 132 owned types are offered; the train motor is withdrawn).
\begin{table}[H]\centering\footnotesize\setlength{\tabcolsep}{3pt}
\caption{\textbf{Lightest (B) and fastest (C) designs and cost.} Median ratios to the agent's passing design of the same task; pass-round tokens exclude cache reads (Codex input includes cached tokens, output includes reasoning).}\label{tab:cost}
\begin{tabular}{@{}lcccccc@{}}\toprule
Agent & B mass & C time & tokens in/out (M) & A hours & B+C sims & B+C hours\\\midrule
\opus & 0.52 & 0.22 & 16.7/12.4 & 82 & 3000 & 463\\
\sonnet & 0.68 & 0.32 & 11.6/8.5 & 34 & 2689 & 267\\
\astra & 0.70 & 0.47 & 7.2/3.0 & 58 & 3000 & 326\\
\sol & 0.60 & 0.58 & 12.7/6.6 & 79 & 3000 & 310\\
\bottomrule\end{tabular}\end{table}
\paragraph{Memory ablation (partial).} Tasks passed in task A with / without lessons of lower levels: \sonnet L2 4 / 3 of 10; \astra L2 7 / 6 of 8; \sonnet L3 2 / 3 of 10. No conclusion beyond the absence of a large effect.

\FloatBarrier
\section{Design space, novelty and assembly}
\label{sec:supp_designs}

\FloatBarrier
\subsection{The space of all simulated designs}
\label{sec:supp_embedding}

Data for \cref{fig:embedding}: 12{,}871 unique geometries from 23{,}395 design files (including 300 simulations of the specialist-pool harness; 12{,}774 from benchmark runs), each rendered from two opposite three-quarter views and described by the mean DINOv2 (ViT-B/14 with registers) feature. t-SNE (cosine) keeps neighbourhoods better than UMAP (10-NN overlap 0.73 vs.\ 0.56). Of a design's 50 nearest neighbours targeting another obstacle, 36\% come from the same agent (25\% by chance). Thumbnails: nearest design to each of 16 $k$-means centres; small points: not passed or unknown (349 restarted lightest/fastest runs).

\begin{figure}[H]
\centering
\includegraphics[width=0.6\linewidth]{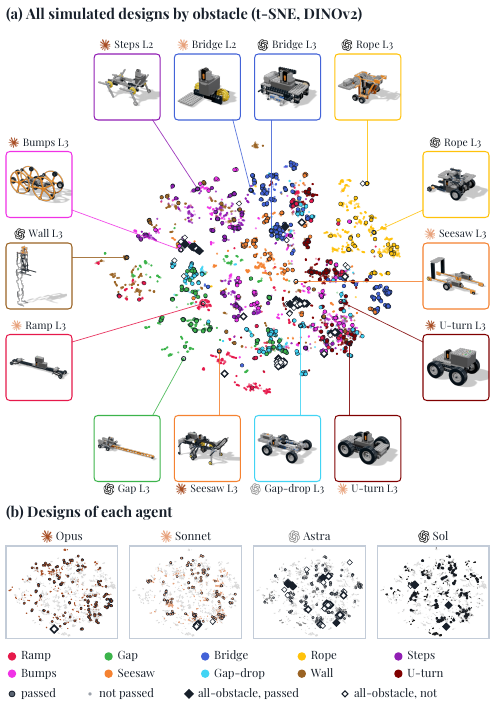}
\caption{\textbf{The design map of \cref{fig:embedding}, per agent.} (a)~All simulated designs, coloured by obstacle; (b)~the same map split by agent.}
\label{fig:embedding_agents}
\end{figure}

\begin{figure}[H]
\centering
\includegraphics[width=\linewidth]{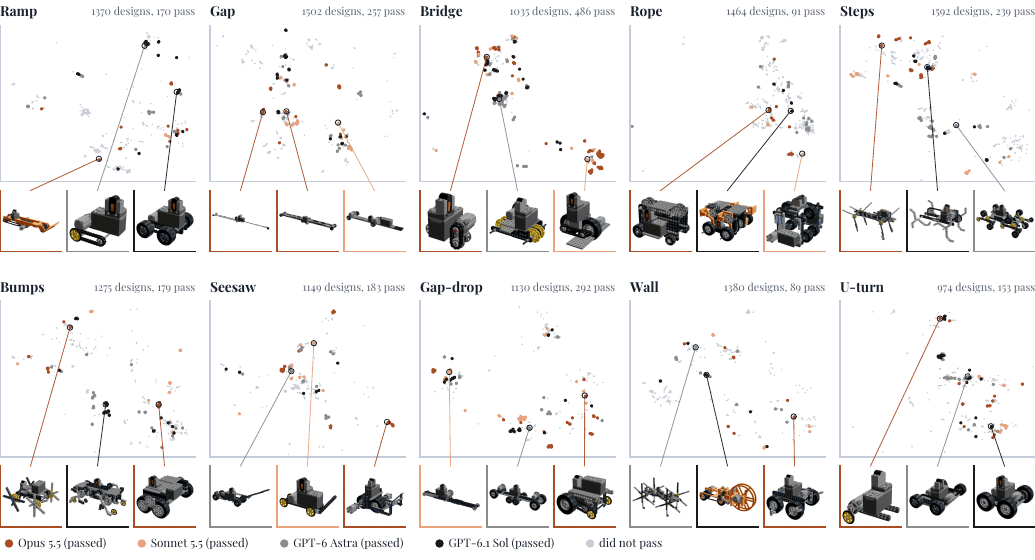}
\caption{\textbf{The same map, one obstacle at a time: agents reach different machines for the same task.} Each panel shows only the designs simulated on one obstacle, all levels, at their positions in \cref{fig:embedding} (grey: did not pass; coloured by agent: passed), zoomed to their extent. Below each panel, three passing designs far apart in the map (the design nearest each of three $k$-means centres of the passing designs, different agents preferred).}
\label{fig:embedding_per_obstacle}
\end{figure}

\begin{figure}[H]
\centering
\includegraphics[width=\linewidth]{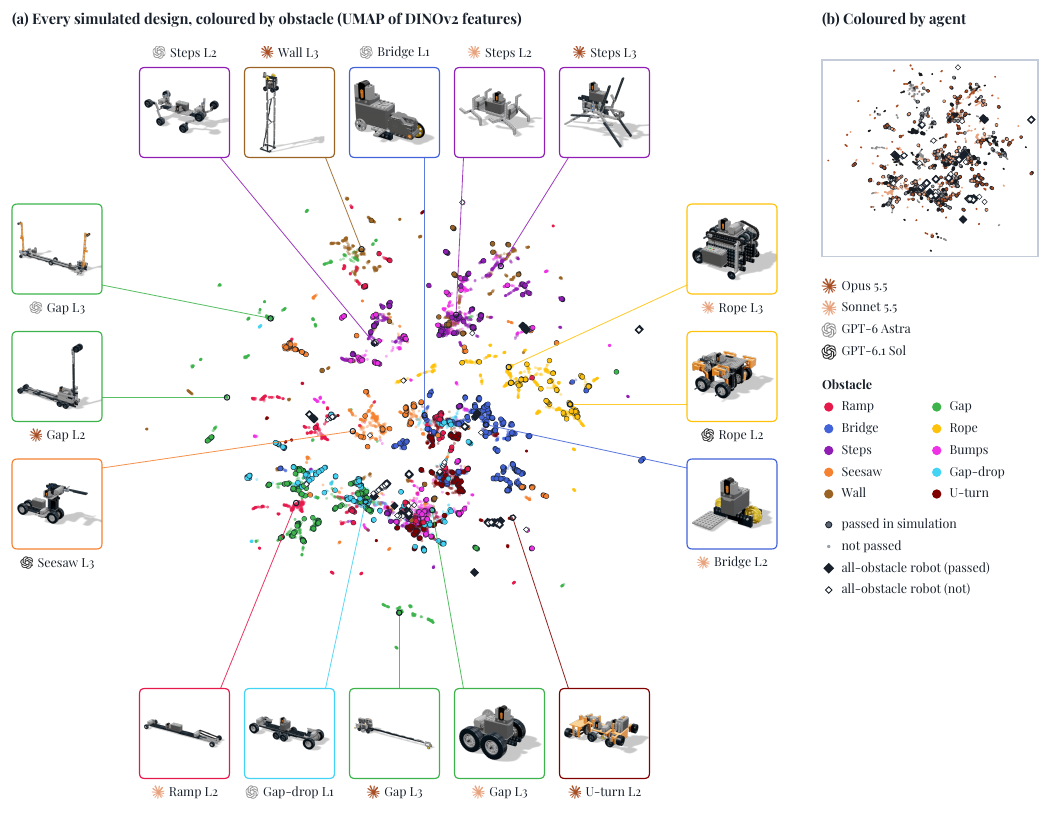}
\caption{\textbf{The same designs projected with UMAP.} Same features and colours as \cref{fig:embedding}; the same organisation by obstacle.}
\label{fig:embedding_umap}
\end{figure}

\FloatBarrier
\subsection{Novelty}
\label{sec:supp_novelty}
\emph{Basic car}: the parametric baselines' default point (the documents' example car, 27 parts). Agent designs: unique geometries with at least one passing simulation (2{,}142, of which 2{,}139 have renders for DINOv2). \emph{NN}: DINOv2 distance (features of \cref{fig:embedding}) to the nearest of the baselines' 2{,}493 unique simulated geometries (pooled across methods; per method they sum to 2{,}670); \emph{out}: share beyond the 95th percentile of the baseline pool's leave-one-out NN distance (0.021). Over all their simulated designs, the baselines stay close to the basic car (mean distance 0.10 random search, 0.07 CMA-ES, 0.12 GA, 0.02 template search), while the agents' passing designs lie 0.20 (L1), 0.26 (L2) and 0.26 (L3) from it.

\paragraph{How the robots work, not only how they look.}
Appearance-only novelty~\cite{regenwetter2023beyond} saturates (98 to 100\% of every agent's designs lie outside the baselines' spread), so \cref{tab:novelty_func} adds function-based measures (template search excluded). Function features: the harness's static free-body and geometry analysis (mass, centre of mass, contact lines, wheels and radii, contact span, track, size, clearance, tractive force, speeds, tipping angle) and part kinds, standardised; \emph{Function NN} follows novelty search~\cite{lehman2011novelty}. \emph{Shah}~\cite{shah2003ideation,sarkar2011assessing} attributes: locomotion, motors, transmission, contact lines, wheel types, printed part, length, mass; a value's novelty is one minus its share among all designs (methods weighted equally). \emph{Vendi}~\cite{friedman2023vendi}: Gaussian kernel on function features, cosine on DINOv2.

\begin{table}[H]
\centering\footnotesize
\caption{\textbf{By how they work, agents' designs are far from the baselines and more varied.} Pool: the parametric baselines; against the full-inventory search (6{,}489 designs, 25\% on treads), agents' function NN is 9.3 (\opus), 8.1 (\sonnet), 9.3 (\astra) and 8.8 (\sol), and the search's Vendi (function) is 3.4 (\cref{tab:main}). Novelty and diversity of each method's designs (agents: unique passing geometries; baselines: unique simulated geometries). Function NN: mean distance to the 5 nearest baseline designs in the function space, in units of the baselines' own nearest-neighbour distance. Shah: a-posteriori novelty over 8 function attributes (0 to 1). Variety: distinct attribute combinations among 100 random designs (mean of 50 draws); concepts: distinct combinations overall. Vendi: effective number of distinct designs among 100 random designs, by function and by DINOv2 appearance. Non-wheeled: share of designs that do not roll only on standard wheels.}
\label{tab:novelty_func}
\begin{tabular}{@{}lrccccccc@{}}
\toprule
Method & Designs & Function NN & Shah & Variety & Concepts & Vendi (function) & Vendi (look) & Non-wheeled \\
\midrule
Random search & 729 & 1.11 & 0.35 & 10.0 & 12 & 2.2 & 2.0 & 0\% \\
CMA-ES & 805 & 0.98 & 0.34 & 7.1 & 9 & 1.9 & 1.8 & 0\% \\
Genetic algorithm & 1{,}136 & 0.94 & 0.38 & 16.6 & 21 & 2.1 & 2.3 & 0\% \\
\midrule
\opus & 789 & \textbf{19.6} & \textbf{0.53} & 50.7 & \textbf{135} & \textbf{13.0} & 4.0 & 11\% \\
\sonnet & 454 & 15.4 & 0.49 & 51.1 & 89 & 8.9 & 3.7 & 12\% \\
\astra & 511 & 16.9 & 0.52 & \textbf{58.6} & 129 & 11.5 & \textbf{4.4} & \textbf{29\%} \\
\sol & 388 & 16.8 & 0.48 & 47.0 & 94 & 10.1 & 3.8 & 19\% \\
\bottomrule
\end{tabular}
\end{table}

\FloatBarrier
\subsection{Assemblability score}
\label{sec:supp_assembly}

Adapts Boothroyd--Dewhurst DFA~\cite{boothroyd2010dfma,samy2010complexity} from the design file: estimated assembly time $T_M$ and efficiency $E = 3\,\mathrm{s}\cdot N_{\min}/T_M$. Handling (s): 1.5 bricks, plates, liftarms, gears, wheels; 1.95 pins, bushes, short axles; 2.0 motors, receivers; 2.5 battery box, printed part; 1.0 per chain or tread link. Insertion (s): 1.5 stud press, 2.0 hinge, 2.5 friction pin or cross axle, +1.0 per further joint. Penalties (s): 3.0 blocked approach, 2.0 spur and 3.5 bevel or worm mesh, 10 chain, 12 tread, 4.0 printed-part fit, 3.0 mounted motor, receiver or battery box, 6.0 cable, 4.5 extra insertion direction, 2.0 per precedence level beyond six, 4.0 sub-assembly, 5.0 axle that can slide out; repeated parts $\times0.8$. Weights are ours, set before looking at real outcomes, not fitted or validated; $T_M$ underestimates real builds, so only $E$ is reported. $N_{\min}$: motors, battery box, needed receivers, rotating elements and their axles, chains, treads, printed parts, shock absorbers, one frame and each other moving rigid body. AUC of the score for real success over the 86 designs: 0.66 ($p{=}0.014$, permutation test; 0.67 at level~1, 0.57 at level~2, 0.56 at level~3); piece count 0.33 (fewer pieces, more passes); the score's residual after regressing on log pieces 0.56 ($p{=}0.37$).

\begin{table}[H]\centering\footnotesize\setlength{\tabcolsep}{3.5pt}
\caption{\textbf{Passing designs use far more pieces than they need.} Medians over the 86 designs that pass \taskCross or \taskPrint. Conn.: connections. Reorient.: re-orientations during assembly. $N_{\min}/N$: share of parts that are necessary. DFA eff.: design-for-assembly efficiency, $3\,\mathrm{s}\cdot N_{\min}/T$, where $T$ is the estimated assembly time (\cref{sec:supp_assembly}). Higher is leaner.}
\label{tab:assembly}
\begin{tabular}{lrrrrrr}
\toprule
Agent & $n$ & Pieces & Conn. & Reorient. & $N_{\min}/N$ & DFA eff. \\
\midrule
\opusL & 24 & 117 & 140 & 2 & 0.22 & 15\% \\
\sonnetL & 16 & \textbf{53} & \textbf{50} & 2 & \textbf{0.24} & \textbf{18\%} \\
\astraL & 23 & 128 & 126 & 2 & 0.21 & 17\% \\
\solL & 23 & 139 & 89 & 2 & 0.18 & 15\% \\
\midrule
All & 86 & 114 & 107 & 2 & 0.22 & 16\% \\
\bottomrule
\end{tabular}
\end{table}

\FloatBarrier
\section{Real tests and the hardware loop}
\label{sec:supp_realloop}

\FloatBarrier
\subsection{Building and testing effort}
\label{sec:supp_effort}
\begin{table}[H]
\centering\footnotesize\setlength{\tabcolsep}{4pt}
\caption{\textbf{Every real result is a robot built by hand.} Robots built and tested, all levels. Hours: 43\,s of assembly per piece plus 30\,min per robot and course for build fixes (repairing parts broken in an earlier failed try, back to the design as specified), course resets and filming, both measured on a 125-piece robot. \taskAll robots are tested on ten courses each. Hours exclude 3D printing, which is machine time (last column).}
\label{tab:effort}
\begin{tabular}{@{}lrrrr@{}}
\toprule
 & Robots & Pieces & Hours & Print (h) \\
\midrule
Simulator learning & 26 & ${\sim}$1{,}130 & ${\sim}$27 & -- \\
Benchmark designs (sim passes) & 87 & 9{,}390 & 156 & ${\sim}$46 \\
\taskAll robots & 4 & 231$^{*}$ & 23 & ${\sim}$23 \\
Hardware loop redesigns & 18 & 2{,}170 & 35 & ${\sim}$37 \\
\midrule
Total & \nRobotsBuilt{} & ${\sim}$\nPiecesBuilt{} & ${\sim}$\nWorkHours{} & ${\sim}$106 \\
\bottomrule
\end{tabular}\\[2pt]
{\scriptsize $^{*}$standard parts only: the \taskAll task asks for a 3D-printed structure.\\ Print: machine time for the printed parts (9, 4 and 6 robots; 0.9, 0.5 and 0.7\,kg of PLA), estimated at 20\,g/h.}
\end{table}

\FloatBarrier
\subsection{Simulated and real trajectories}
\label{sec:supp_traj}
\label{sec:traj}
Twelve level-1 tries (eleven designs): robot centre tracked by frame differencing (854$\times$480, 10\,fps), goal line marked by eye ($\pm$15\,px), progress aligned with the simulation at motion onset. Final-progress error and dynamic-time-warping distance separate passes (0 and $\leq$0.01) from failures (0.14--0.86 and 0.09--0.54); Pearson correlation does not. Over the 48 designs that cross in simulation and for real, simulated and real crossing times correlate (Spearman 0.78, Pearson 0.57).

\begin{figure}[H]
\centering
\includegraphics[width=0.245\linewidth]{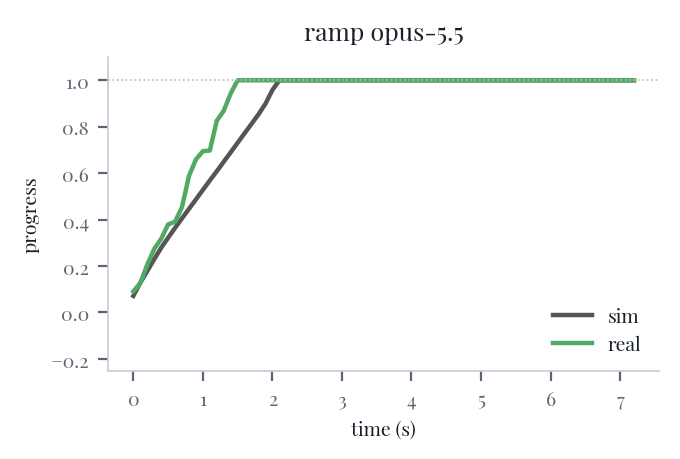}\hfill
\includegraphics[width=0.245\linewidth]{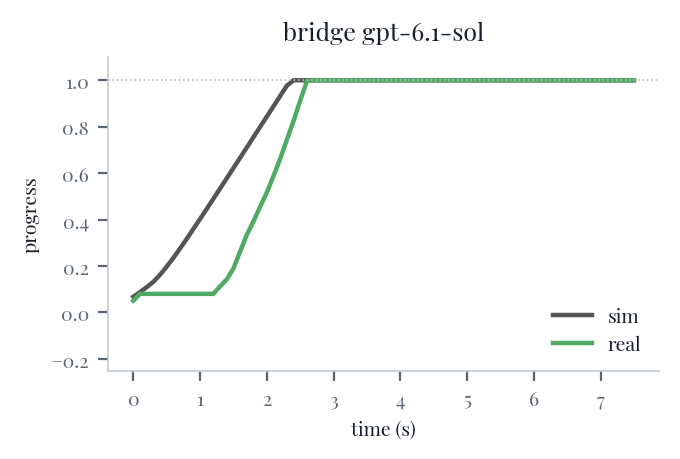}\hfill
\includegraphics[width=0.245\linewidth]{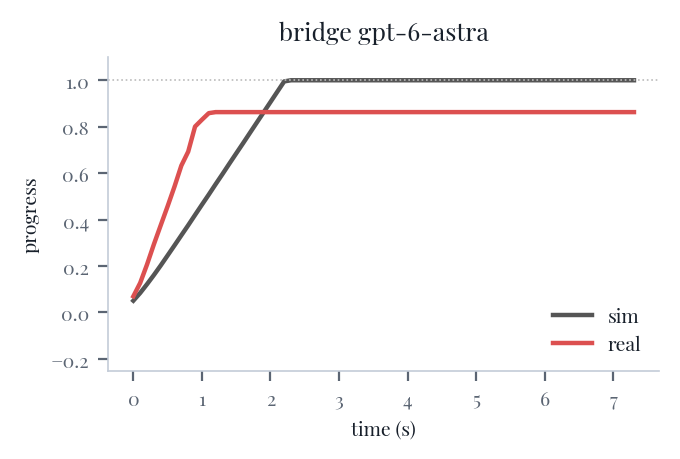}\hfill
\includegraphics[width=0.245\linewidth]{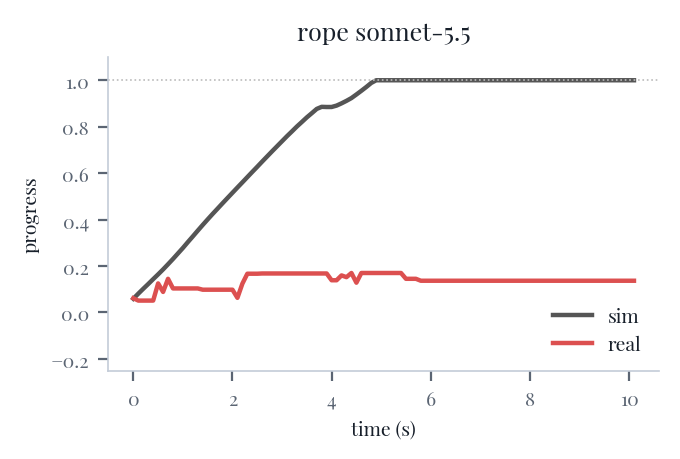}
\caption{\textbf{Simulated and real progress of four level-1 designs.} Top: real passes follow their simulation. Bottom: real failures end short of the goal.}
\label{fig:traj}
\end{figure}

\FloatBarrier
\subsection{Hardware loop protocol and redesigns}
\label{sec:supp_loop}

\begin{table}[H]
\centering\footnotesize\setlength{\tabcolsep}{3.5pt}
\caption{\textbf{The hardware loop fixes most real failures for every agent.} Levels 1 and 2: share of simulated passes that cross the real course before and after one round of the loop, and real failures.}
\label{tab:loop_agents}
\begin{tabular}{@{}lcccc@{}}
\toprule
 & \multicolumn{2}{c}{Real pass (\%)} & \multicolumn{2}{c}{Real failures} \\
\cmidrule(lr){2-3}\cmidrule(l){4-5}
Agent & before & after & before & after \\
\midrule
\opusL & 67 & 83 & 6 & 3 \\
\sonnetL & 77 & \textbf{100} & 3 & \textbf{0} \\
\astraL & 79 & \textbf{100} & 4 & \textbf{0} \\
\solL & 71 & 88 & 5 & 2 \\
\midrule
All & 73 & 93 & 18 & 5 \\
\bottomrule
\end{tabular}
\end{table}

\begin{table}[H]
\centering\small
\caption{\textbf{Closing the loop on every real failure.} Per agent: level-1--2 real failures sent to the loop, redesigns that pass the verdict and the hand-in vote, and redesigns that cross for real. $^*$Includes \sonnet's U-turn level-1 design, filmed as its \taskLight design and left out of the real pass rates; its redesign crosses for real.}
\label{tab:loop}
\setlength{\tabcolsep}{3pt}\footnotesize
\begin{tabular}{@{}lcccc@{}}
\toprule
\rowhead Agent & failures & redesign & vote & real \\
\midrule
\opus & 6 & 5 & 5 & 3/5 \\
\sonnet$^*$ & 4 & 4 & 4 & 4/4 \\
\astra & 4 & 4 & 4 & 4/4 \\
\sol & 5 & 5 & 5 & 3/5 \\
\midrule
All$^*$ & 19 & 18 & 18 & 14/18 \\
\bottomrule
\end{tabular}
\end{table}

The per-failure loop of \cref{sec:watch}, for every level-1--2 design that passes in simulation but fails for real, run by the agent that designed it in a fresh session.

\paragraph{Watching a real failure.}

Inputs: design, filmed tries, simulation, and frame-sheet and real-versus-simulation tools. Before simulating, the agent registers a failure record: per try a timeline of $\geq$5 timestamped events, every motor's action, $\geq$3 real--simulated differences, why tries differ, and per failure the part and class (design flaw, simulator gap, build and operator).

\paragraph{A minimal correction from one failure.}

Optional: at most two changed and one new parameter and 150 changed lines, nothing course-, design- or part-name-specific, each change cited with video evidence. Accepted if the earlier design fails under it as it did for real, runtime stays within twice the original and results are unchanged with names removed; used only for this redesign.

\paragraph{Robust redesign and retest.}

Redesign with the \taskCross budget and verdict under the corrected simulator. Vote: five seeded runs with declared uncertain parameters drawn from their ranges, per-motor speed and torque factors (s.d.\ 8\%) and driven wheel or leg groups started up to $\pm$30$^\circ$ off; at least three must pass. A fix map names, per recorded failure, the change or why it is not addressed. Passing redesigns are built and filmed (\cref{sec:testbed}).

\paragraph{Corrections and rounds.} Accepted corrections include rod friction for hard plastic on the printed bump rods, 0.387 to 0.30 (\opus, bumps L2), tyre friction on the glossy wall face (\opus, wall L2), wheel contact on the cord at the groove radius instead of the outer radius (\sol, rope L1) and tyre contact width on the cord (\sol, rope L2). Thirteen redesigns passed in the \taskCross round and five needed \taskPrint; six use a printed part, including \sonnet's printed tub for the U-turn, which crosses for real, and \sol's level-2 rope trolley, which keeps its original printed part. Three sessions that started before uncertainty ranges were required (\opus wall L2, \sonnet bridge L2 and seesaw L1) voted without ranges; the two \sonnet redesigns cross for real and \opus's does not.

\FloatBarrier
\paragraph{New concept or adjustment.}
\emph{New concept}: at least a third of the parts replaced relative to the failed original (Jaccard distance $\geq0.5$) (as for concept steps), else \emph{adjustment}. \emph{Early}: the robot barely started the obstacle in its real tries; else \emph{partway}. Counts are small; the table is descriptive.

\begin{table}[H]
\centering\footnotesize
\caption{\textbf{Agents replace the mechanism when the robot fails partway and adjust it when it barely starts.} The 18 loop redesigns (with \sonnet's U-turn) by response and by how far the original got in its real tries; cells give redesigns that cross for real / redesigns.}
\label{tab:redesign}
\begin{tabular}{@{}lccc@{}}
\toprule
 & Failed partway & Failed early & All \\
\midrule
New concept & 7/7 & 1/2 & 8/9 \\
Adjustment & 1/2 & 5/7 & 6/9 \\
\bottomrule
\end{tabular}
\end{table}

\FloatBarrier
\subsection{One design through the loop}
\label{sec:supp_method}
The \opus bumps L2 case of \cref{fig:teaser,fig:benchmark}, step by step.

\begin{figure}[H]
\centering
\includegraphics[width=0.6\linewidth]{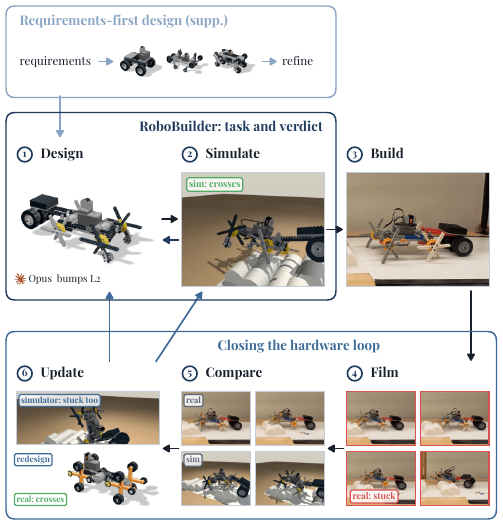}
\caption{\textbf{A simulator pass is only the middle of the loop.} One design (\opus, bumps L2): simulated pass (2), build and film (3--4), real failure compared with the simulation (5), updated simulator and a redesign that crosses for real (6).}
\label{fig:method}
\end{figure}

\FloatBarrier
\section{Harnesses for system design}
\label{sec:supp_harness}
These results are exploratory: each harness variant ran once.

\FloatBarrier
\subsection{Requirements-first design}
\label{sec:supp_reqfirst}
A requirements-first harness around the same agents, simulator and budgets.

\paragraph{Method.}
Before simulating, the agent lists the obstacle's challenges, each with a numeric requirement and $\geq$3 mechanisms on different physical principles, registers measurable requirements (metric, unit, threshold) and names $\geq$3 concepts; each concept is simulated once before any twice, and a plateau forces a restart with an untried mechanism. Requirements may be revised only with the justifying simulation. In rev.~5 the harness measures requirements on the design (clearance, track, reach, mass, torque margins) and refuses to simulate one that misses them. Budgets: 30 + 30 (printed part) per obstacle; 300 for \taskAll.

\paragraph{Single level-3 obstacles: a small gain for three agents.}
Harness (baseline), one run per task, \taskCross and \taskPrint counted: \opus 7 (6), \sonnet 5 (3), \astra 5 (4), \sol 6 (6), 23 against 19 of 40; one or two tasks per agent is within single-run noise.

\paragraph{\taskAll task: no reliable gain.}
Rescored as in \cref{sec:supp_variance}. Two earlier exploration variants lose almost everything under the released simulator. The \astra requirements-first robot fails the official verdict on all obstacles (a seesaw failure flagged by the design review); the repair session fixes this and keeps seven.

\begin{table}[H]
\centering\small
\caption{\textbf{The requirements-first harness does not reliably beat single agents on the \taskAll task.} Obstacles crossed (of 10): official verdict, robust ($\geq$6 of 10 perturbed repeats) and all ten repeats. +\,fix, +\,repair: follow-up repair sessions; pools, v2, optimizer, fork: \cref{sec:supp_pools}.}
\label{tab:supp_reqfirst}
\setlength{\tabcolsep}{3pt}\footnotesize
\begin{tabular}{@{}llccc@{}}
\toprule
\rowhead Run & Agent & official & robust & 10/10 \\
\midrule
Baseline & \opus & 5 & 6 & 4 \\
Baseline & \sonnet & 3 & 3 & 2 \\
Baseline & \astra & 6 & 7 & 4 \\
Baseline & \sol & 5 & 5 & 2 \\
\midrule
Req.-first & \sonnet & 2 & 2 & -- \\
Req.-first & \astra & 0 & 7 & 7 \\
Req.-first + fix & \opus & 4 & 4 & -- \\
Req.-first + repair & \astra & \textbf{7} & \textbf{7} & \textbf{7} \\
\midrule
Verified req. & \opus & 4 & 4 & 3 \\
Verified req. & \sonnet & 4 & 4 & -- \\
Pools & \opus & 2 & 3 & 1 \\
Pools & \astra & \textbf{7} & \textbf{7} & 6 \\
Pools, v2 & \opus & \textbf{7} & 6 & 4 \\
Pools, v2 & \sonnet & 3 & 2 & 0 \\
Pools, v2 & \astra & \textbf{7} & \textbf{7} & \textbf{7} \\
Pools, v2 & \sol & \textbf{7} & \textbf{7} & 5 \\
Pools, v2 + optimizer & \opus & \textbf{7} & \textbf{7} & 5 \\
Pools, v2 + optimizer & \sonnet & 3 & 3 & 2 \\
Pools, v2 + optimizer & \astra & \textbf{7} & \textbf{7} & 6 \\
Pools, v2 + optimizer & \sol & \textbf{7} & \textbf{7} & 6 \\
Pools, v2 + fork & \opus & 6 & 6 & 3 \\
Pools, v2 + fork & \sonnet & 3 & 4 & 2 \\
Pools, v2 + fork & \astra & \textbf{7} & 6 & 6 \\
Pools, v2 + fork & \sol & \textbf{7} & \textbf{7} & 5 \\
\bottomrule
\end{tabular}
\end{table}

\paragraph{Verified requirements (rev.~5).}
\opus at 90 simulations: 31 revisions; 67 of the 88 designs (76\%) that met all requirements failed in simulation, mostly stuck. \sonnet by simulation 265: 288 revisions, 170 simulations refused; 109 of 259 compliant designs (42\%) failed.

\begin{figure}[H]
\centering
\includegraphics[width=\linewidth]{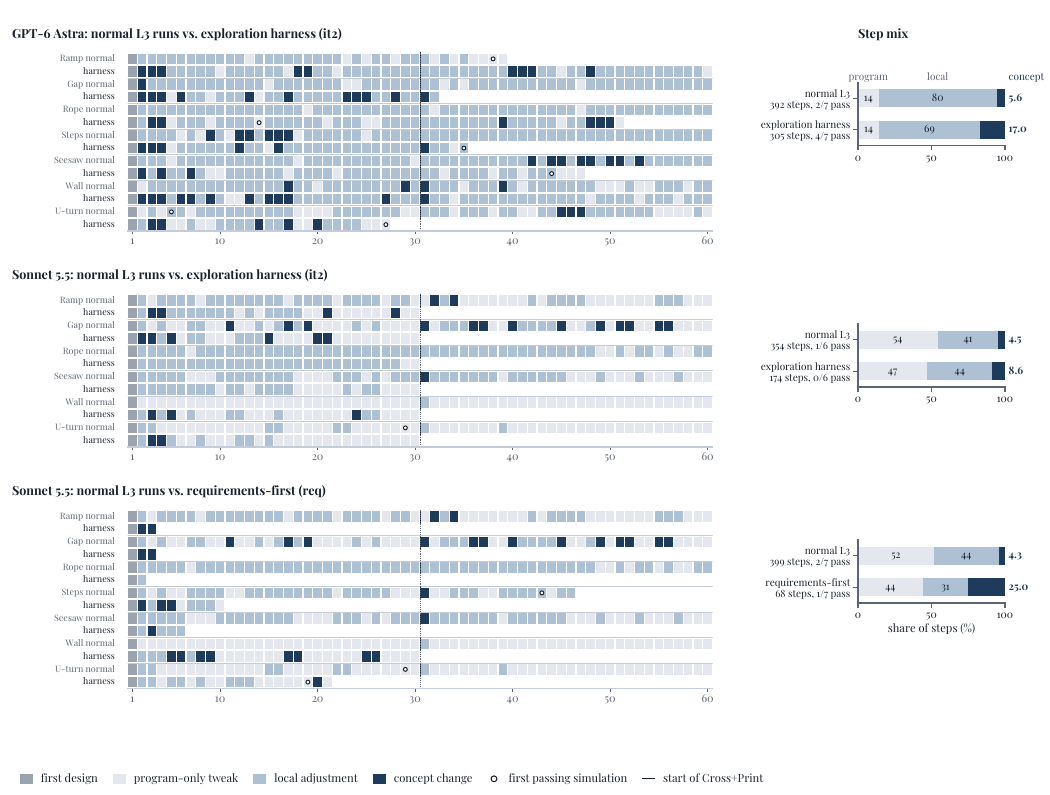}
\caption{\textbf{Design changes per simulation with the requirements-first harness overlaid.} As \cref{fig:iterchange}, with the runs of the requirements-first harness added for comparison.}
\label{fig:iterchange_harness}
\end{figure}

\FloatBarrier
\subsection{Specialist pools and a synthesizer}
\label{sec:supp_pools}
\begin{figure}[H]
\centering
\includegraphics[width=0.6\linewidth]{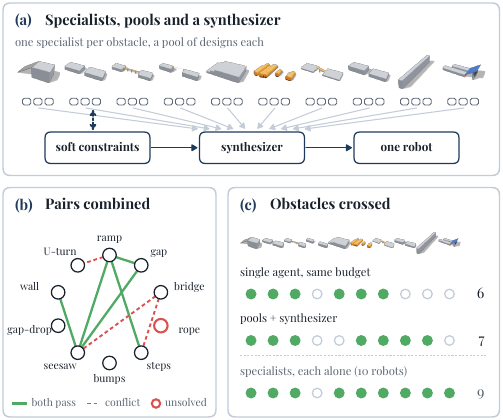}
\caption{\textbf{Specialists solve nine obstacles alone, but one robot keeps seven.} \astra, level-1 \taskAll task. (a)~Specialist pools and a synthesizer. (b)~Synthesized pairs: both pass (solid) or not (dashed). (c)~Obstacles crossed by a single agent, the synthesized robot and the specialists alone.}
\label{fig:harness}
\end{figure}
Some obstacle requirements conflict (clearance for bumps, narrow track for the bridge, cord grip for the rope, reach for the wall), so the harness solves them separately, keeps alternatives and integrates.

\paragraph{Specialists and pools.}
Ten specialists, one per obstacle (15 simulations each), explore several mechanisms per challenge and keep pools of diverse passing or near-passing designs; after each round they post \emph{soft constraints} (e.g.\ maximum width, clearance, drive layout) to a shared board.

\paragraph{Synthesizer.}
Combines pool designs pairwise, then into one robot, screening on three obstacles before the full set (150 simulations; 300 in total); combined designs record their sources (\cref{fig:harness}b). \emph{v2}: reuses a finished pool run, cross-tests each specialist's best robot on every obstacle, never hands in a robot worse than its best. \emph{Optimizer}: numerical optimization of agent-declared parameters. \emph{Fork}~\cite{cai2026forkflush}: three fresh-context sessions, each from a new base (part distance $\geq$0.5).

\begin{table}[H]
\centering\small
\caption{\textbf{Integration, not single obstacles, limits the \taskAll robot.} \taskAll task (level 1, \astra, 300 simulations): obstacles crossed by the submitted robot (official verdict).}
\label{tab:harness}
\setlength{\tabcolsep}{3pt}\footnotesize
\begin{tabular}{@{}lcc@{}}
\toprule
\rowhead Method & crossed /10 & real /10 \\
\midrule
Single agent & \nSingle & \nAllRealAstra \\
Specialist pools + synthesizer & \nPool & -- \\
\quad specialists alone (best per obstacle) & \nPoolSpecialists & -- \\
\bottomrule
\end{tabular}
\end{table}

Integration loses two obstacles: the rope (gripping a cord demands a narrow, light, top-hung chassis the other obstacles do not tolerate) and the U-turn, solved by its specialist but lost when combined. Each method ran once, so a one-obstacle difference in \cref{tab:supp_reqfirst} is within run-to-run noise; the larger gains of v2 for \sol and \astra are the clearest effect.

\FloatBarrier
\section{Detailed results}
\label{sec:supp_results}

\FloatBarrier
\subsection{Per-design results}

\Cref{tab:expanded}: every task and agent; the released runs add designs, tokens, wall time and videos.

\FloatBarrier
\subsection{Every task, every agent}
\label{sec:supp_detailed}
\Cref{tab:sim_results,tab:opt,tab:allobst,tab:real} break the columns of \cref{tab:main} down by level and agent; \cref{tab:robust} checks the combined score.

\begin{table}[H]\centering\setlength{\tabcolsep}{3pt}
\caption{\textbf{No search baseline passes more than one task.} All non-agentic baselines, in the columns of \cref{tab:main}. Full inventory: random search and a GA over build actions on the whole inventory (30 simulations per task; novelty and diversity against this pool, as in \cref{tab:main}). Parametric: each searches a hand-built family of cars (19 parameters: motor, wheel, rails, wheelbase, gearing, drive, deck and speed program) with the same simulator, verdict and budget as the agents.}
\label{tab:baselines}
\begin{tabular}{@{}lcccccccccc@{}}
\toprule
 & \multicolumn{3}{c}{\taskCross (\%)} & +\textsc{Print} & Sims & \taskLight & \taskFast & Score & Novelty & Diversity \\
\cmidrule(lr){2-4}
Method & L1 & L2 & L3 & (\%) & (med.) & score & score & (/100) & (function) & (Vendi) \\
\midrule
Full inventory, random & 0 & 0 & 0 & 0 & -- & -- & -- & 0.0 & 1.1 & 3.4 \\
Full inventory, GA & 0 & 0 & 0 & 0 & -- & -- & -- & 0.0 & 0.9 & 3.2 \\
\midrule
Parametric, random & 10 & 0 & 0 & 3 & 9 & 0.0 & 0.0 & 0.8 & 1.1 & 2.2 \\
Parametric, CMA-ES & 0 & 0 & 0 & 0 & -- & 0.0 & 0.0 & 0.0 & 1.0 & 1.9 \\
Parametric, template & 0 & 0 & 0 & 0 & -- & 0.0 & 0.0 & 0.0 & -- & -- \\
Parametric, GA & 10 & 0 & 0 & 3 & 9 & 0.7 & 0.4 & 1.7 & 0.9 & 2.1 \\
\bottomrule
\end{tabular}
\end{table}

\paragraph{Baselines and scale.} The full-inventory search decodes a vector of 98 numbers into a robot: rail length, chassis width, battery position and the motor program, then up to 12 build actions (axles with any owned wheel, narrow, wide or chain treads, one printed rim type in up to four copies, motor drives with legal gear trains, chains, ballast, liftarm arms); invalid robots are repaired or rejected without using a simulation. Random search samples the vector; the GA crosses over action sequences (population 6). Both pass no task in 30 simulations; with 300 simulations at level~1 they pass three of 20 runs, two of them with a printed rim, which the agents' rules would count as \taskPrint. Their designs roll on wheels (75\%) or treads (25\%). The parametric baselines start from the documents' example car. GA: real-coded, population 6 (example car in the first generation), tournament $k{=}2$, BLX-0.5 or uniform crossover (p\,=\,0.9), Gaussian mutation ($\sigma{=}0.1$ per gene, p\,=\,0.25), one elite; it also runs \taskPrint (a printed rim beside each wheel), \taskLight and \taskFast. Template search: only the example car survives the design review; its program is searched. Limits: \taskCross 24\,h, \taskAll 72\,h (one named course per simulation, scored on all ten). Agents ran \nSimsAX{} (\taskCross, \taskPrint), \nSimsBC{} (\taskLight, \taskFast) and \nSimsAll{} (\taskAll) simulations, \nGeoAgents{} distinct geometries (baselines \nGeoBaselines{}). Real tests: \nFilmedAll{} sim-passing designs and the four \taskAll robots, \nVidBench{} trimmed tries.

\paragraph{Notes on the tables.} \Cref{tab:main}: \taskLight mass and \taskFast time average the 10 and 9 level-1 tasks where all four agents have a design; the GA ran them only on ramp L1 (agents: 178--184\,g, 0.90--3.14\,s); single LLM call: best model (\opus), \taskAll column best of four; baseline novelty and diversity use all distinct simulated geometries. \Cref{tab:robust}: geometric mean floors components at 0.01; bootstrap: 2{,}000 resamples of the ten obstacles (with their three level tasks and \taskAll verdict). \Cref{tab:expanded}: a \taskPrint pass's simulation index counts the 30 of \taskCross; \taskLight and \taskFast ran on every task (from scratch where nothing passed) and show the passing design when nothing lighter or faster was accepted; Real: crosses (\cmark) or fails (\xmark); $^\ddagger$\sonnet's U-turn L1 robot was filmed as its \taskLight design although \taskCross and \taskPrint did not pass, and is not counted in the real results. Tokens exclude cache reads (Codex input includes cached tokens, output includes reasoning).

\input{sec/tab_expanded}
\input{sec/tab_robust}

\paragraph{Combined score: robustness.} Geometric mean and mean rank order the agents as the arithmetic mean, and every baseline stays below every agent (\cref{tab:robust}); bootstrap intervals are wide (ten obstacles, single runs).

\begin{figure}[H]
\centering
\includegraphics[width=\linewidth]{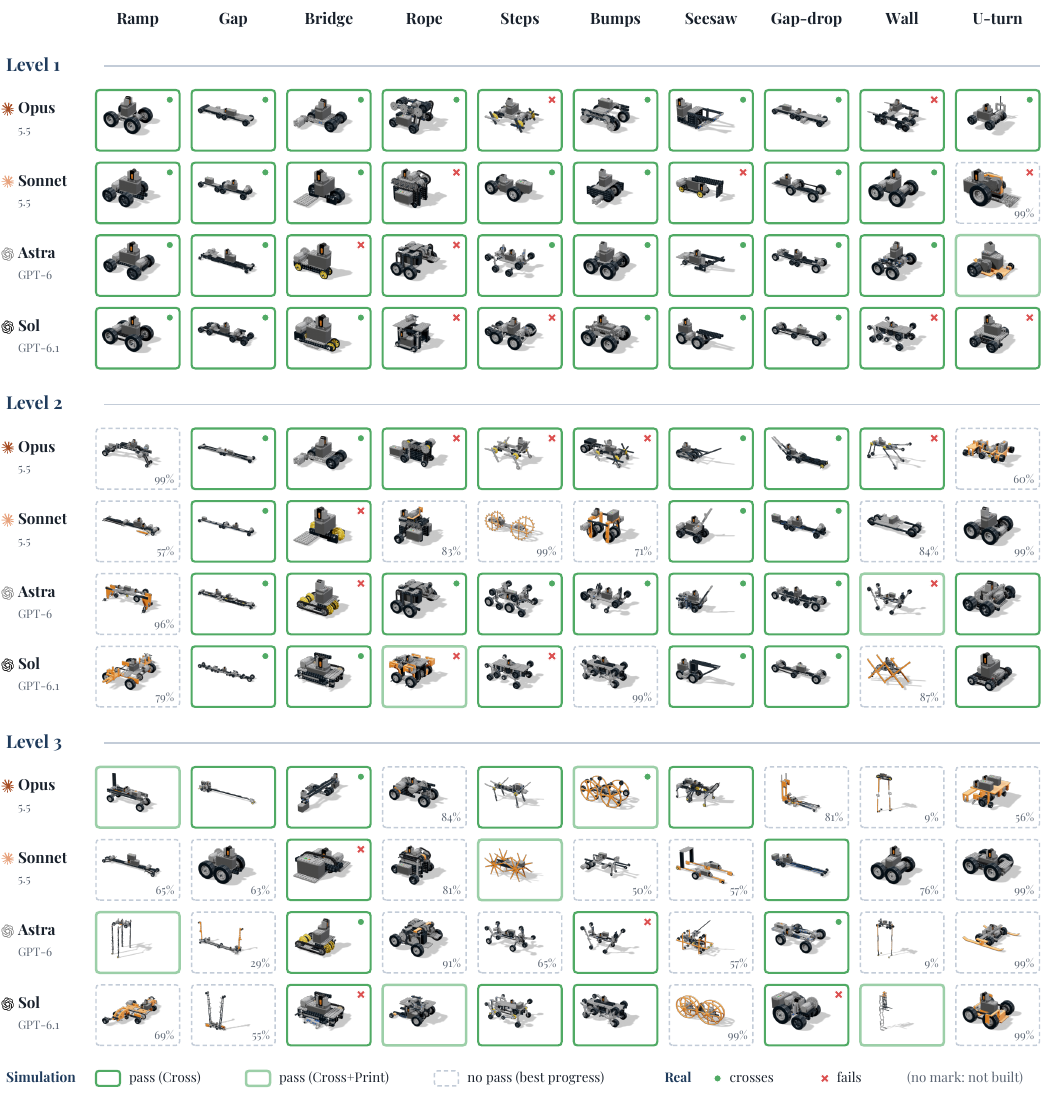}
\caption{\textbf{Agents pass nearly every level-1 obstacle; harder levels bring legs, whegs, printed parts and fewer passes.} One design per agent, level and obstacle. Frame: pass (dark green), pass with a printed part (light green), no pass (grey, best progress). Badge: crosses (dot) or fails (cross) for real.}
\label{fig:results}
\end{figure}

\begin{table}[H]
\centering\small
\caption{\textbf{Agents pass nearly every level-1 task but under half at level~3; baselines pass almost none.} Tasks passed (of 10 per level; in parentheses, with the printed part). All-obst.: obstacles crossed by one robot (of 10).}
\label{tab:sim_results}
\setlength{\tabcolsep}{3pt}\footnotesize
\begin{tabular}{@{}lcccc@{}}
\toprule
\rowhead Method & L1 & L2 & L3 & All-obst. \\
\midrule
Random search (parametric) & 1 & 0 & 0 & -- \\
CMA-ES (parametric) & 0 & 0 & 0 & -- \\
Template search & 0 & 0 & 0 & -- \\
Genetic algorithm & 1 & 0 & 0 & -- \\
Single LLM call, \opus & 2 & 0 & 0 & 0 \\
Single LLM call, \sonnet & 1 & 0 & 0 & 0 \\
Single LLM call, \astra & 0 & 0 & 1 & 0 \\
Single LLM call, \sol & 1 & 0 & 0 & 0 \\
\midrule
\opus & 10 & 8 & 6 (2) & 5 \\
\sonnet & 9 & 4 & 3 (1) & 3 \\
\astra & 10 (1) & 9 (1) & 4 (1) & \nSingle \\
\sol & 10 & 7 (1) & 6 (2) & 5 \\
\midrule
All agents (of 40) & \nSimLOne & \nSimLTwo & \nSimLThree & 19 \\
\bottomrule
\end{tabular}
\end{table}

\begin{table}[H]
\centering\small
\caption{\textbf{\opus leads both optimization objectives; speed separates the agents more than mass.} Per task: best mass (time) among agents divided by the agent's, 0 if none; summed per level (max 10).}
\label{tab:opt}
\setlength{\tabcolsep}{3pt}\footnotesize
\begin{tabular}{@{}lcccccccc@{}}
\toprule
\rowhead & \multicolumn{4}{c}{Lightest (\taskLight)} & \multicolumn{4}{c}{Fastest (\taskFast)} \\
\rowhead Agent & L1 & L2 & L3 & total & L1 & L2 & L3 & total \\
\midrule
\opus & \textbf{9.2} & \textbf{8.3} & 5.1 & \textbf{22.6} & \textbf{9.5} & \textbf{9.6} & \textbf{8.0} & \textbf{27.1} \\
\sonnet & 8.4 & 5.3 & 3.5 & 17.2 & 6.4 & 3.1 & 1.5 & 11.0 \\
\astra & 7.4 & 6.5 & 4.2 & 18.1 & 4.1 & 4.0 & 2.0 & 10.1 \\
\sol & 7.8 & 4.7 & \textbf{6.0} & 18.5 & 4.2 & 2.0 & 3.4 & 9.6 \\
\bottomrule
\end{tabular}
\end{table}

\begin{table}[H]
\centering\small
\caption{\textbf{One robot for all ten obstacles crosses at most six.} Obstacles passed by the handed-in level-1 robot in simulation (official verdict: plain run and both nudged starts) and for real. \opus and \sol cross one obstacle more for real than in simulation: \opus's robot passes bumps in a rescore of the same design and in 7 of 10 perturbed repeats, and \sol's robot passes gap-drop in the plain run but not with a 50\,ms late start.}
\label{tab:allobst}
\setlength{\tabcolsep}{4pt}\footnotesize
\begin{tabular}{@{}lcc@{}}
\toprule
\rowhead Agent & sim /10 & real /10 \\
\midrule
\opus & 5 & \nAllRealOpus \\
\sonnet & 3 & \nAllRealSonnet \\
\astra & \nSingle & \nAllRealAstra \\
\sol & 5 & \nAllRealSol \\
\bottomrule
\end{tabular}
\end{table}

\begin{table}[H]
\centering\small
\caption{\textbf{About two in three simulated successes cross for real, fewer at harder levels.} Per try, 127 of 188 level-1 and level-2 tries cross (68\%).}
\label{tab:real}
\setlength{\tabcolsep}{3pt}\footnotesize
\begin{tabular}{@{}lcccccc@{}}
\toprule
\rowhead & \multicolumn{2}{c}{L1} & \multicolumn{2}{c}{L2} & \multicolumn{2}{c}{L3} \\
\rowhead Agent & cross & tested & cross & tested & cross & tested \\
\midrule
\opus & 8 & 10 & 4 & 8 & 3 & 6 \\
\sonnet & 7 & 9 & 3 & 4 & 1 & 3 \\
\astra & 8 & 10 & 7 & 9 & 2 & 4 \\
\sol & 7 & 10 & 5 & 7 & 2 & 6 \\
\midrule
All & \nRealPassLOne & \nFilmedLOne & \nRealPassLTwo & \nFilmedLTwo & \nRealPassLThree & \nFilmedLThree \\
\bottomrule
\end{tabular}
\end{table}

\begin{figure}[H]
\centering
\includegraphics[width=\linewidth]{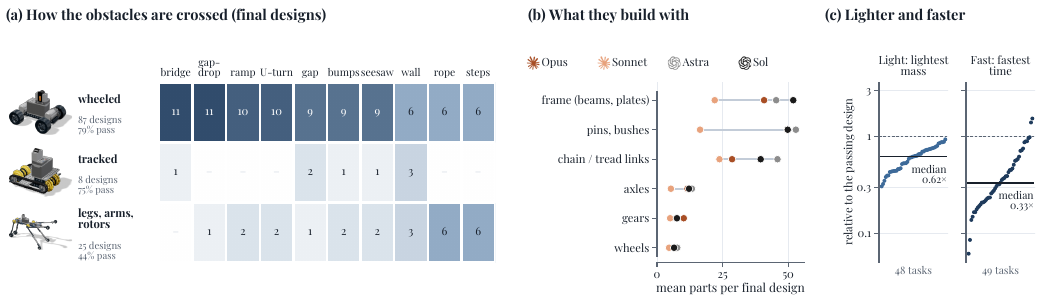}
\caption{\textbf{Agents build wheeled machines and use legs, arms and rollers only where needed.} (a)~Final designs per locomotion type and obstacle, with each type's pass rate. (b)~Mean parts per design by category, one dot per agent. (c)~Lightest (\taskLight) and fastest (\taskFast) designs relative to the passing design; line: median.}
\label{fig:designs}
\end{figure}

\FloatBarrier
\subsection{Results at a glance}
\label{sec:supp_glance}

\begin{figure}[H]
\centering
\includegraphics[width=\linewidth]{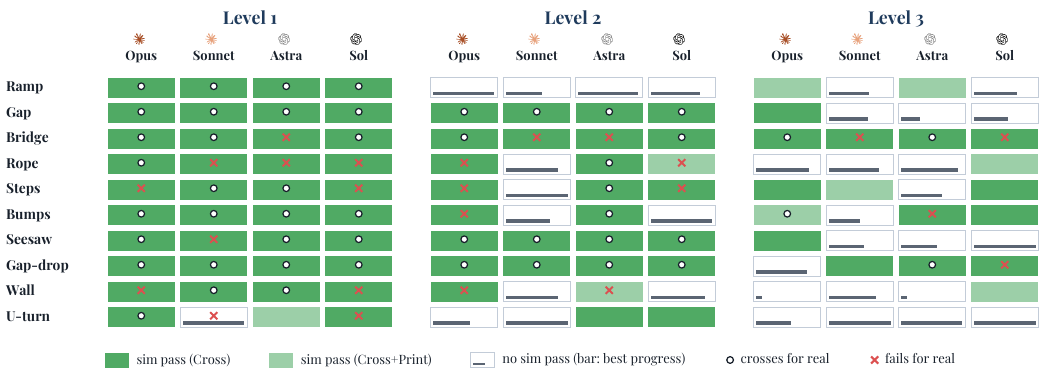}
\caption{\textbf{Simulation and real results per task.} One cell per agent, obstacle and level: pass in \taskCross, pass in \taskPrint or no pass (shade: best progress); symbols: built and filmed, crosses or fails for real.}
\label{fig:glance}
\end{figure}

\FloatBarrier
\subsection{Lightest and fastest designs}
\label{sec:supp_lightfast}
\begin{figure}[H]
\centering
\includegraphics[width=0.6\linewidth]{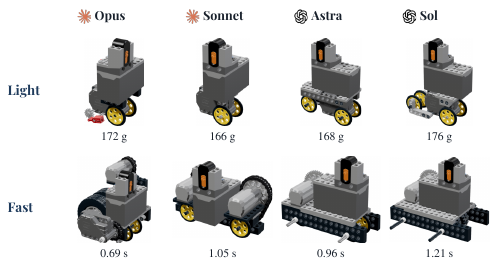}
\caption{\textbf{Lightest and fastest designs on bridge level~1.} One column per agent; mass of the \taskLight design, time to the goal of the \taskFast design.}
\label{fig:lightfast}
\end{figure}

\Cref{fig:lightfast}: bridge level~1, the one task where all four agents have both designs; the lightest converge on one compact car, the fastest differ in gearing and drive.

\begin{table}[H]
\centering\footnotesize\setlength{\tabcolsep}{4pt}
\caption{\textbf{\taskLight and \taskFast leaderboard.} Tasks with a design and tasks won (lightest mass, fastest time) per agent; ties go to the first listed.}
\label{tab:lightfast}
\begin{tabular}{@{}lcccc@{}}
\toprule
 & \multicolumn{2}{c}{\taskLight (29 tasks)} & \multicolumn{2}{c}{\taskFast (30 tasks)} \\
\cmidrule(lr){2-3}\cmidrule(lr){4-5}
Agent & designs & best & designs & best \\
\midrule
\opusL & 25 & \textbf{16} & 28 & \textbf{23} \\
\sonnetL & 20 & 7 & 18 & 3 \\
\astraL & 24 & 2 & 24 & 1 \\
\solL & 24 & 4 & 23 & 3 \\
\bottomrule
\end{tabular}
\end{table}

\FloatBarrier
\subsection{Run-to-run variance}
\label{sec:supp_variance}
Full agent runs were not repeated (one pass over all tasks takes nearly 4{,}000 simulations per agent). The four \taskAll robots were rescored with ten seeded perturbed repeats per obstacle (start offset up to 2\,mm, commands up to 50\,ms late, battery 0.95 to 1.05): against the official counts (\opus 5, \sonnet 3, \astra 6, \sol 5), 6, 3, 7 and 5 obstacles pass in at least six of ten repeats, and two to four pass all ten. Under every aggregation and bootstrap resample (\cref{tab:robust}) \opus leads; the intervals of the other three agents overlap, so their order is not resolved.

\FloatBarrier
\section{Simulator learning}
\label{sec:supp_simlearn}
How the agent (\opus) turned the simulator initialised from public part data into the learned simulator of the benchmark (\cref{sec:simbuild}).

\FloatBarrier
\subsection{Before and after simulator learning}
\label{sec:supp_sim}

\paragraph{Before.}
MuJoCo models built from per-part geometry and masses, with motor models and course twins; friction, joint strength and motor curves from public part data and estimates, none fitted to a real run.

\paragraph{After.}
Parts stay rigid; stud joints are breakable welds (pull out, peel, shear); axles have compliance and can slide; overloaded gears skip and re-engage; motors follow a torque--speed model with IR latency and battery sag. Parameters come from the bench measurements (\cref{tab:measure}) and are fitted by the agent to filmed runs of five calibration rounds (train / validation split); changes were added one at a time (\cref{tab:physics}) and the remaining inventory mechanisms audited and modelled. A run counts as crossing if any try crosses; the three labelled runs still predicted wrong are two gap cars (simulated crosses, real stops) and one steps climber.

\begin{table}[H]
\centering\small
\caption{\textbf{Before the first round no simulated pass held for real; after it, two in three do.} Rows: simulated passes that cross for real; labelled runs predicted (all, validation); first-round agent designs (before: the 15 that passed; after: all 22 filmed). The before and after rows use different designs.}
\label{tab:sim}
\setlength{\tabcolsep}{3pt}\footnotesize
\begin{tabular}{@{}lcc@{}}
\toprule
\rowhead & before & after \\
\midrule
Sim passes crossing for real & \nInitReal/\nInitSimPass{} & \nRealPassAll/\nFilmedAll{} (\nRealRateAll\%) \\
First-round agent designs predicted & 0/15 & 20/22 \\
Labelled runs predicted & -- & \nCalRight/\nCalRuns \\
\quad of which validation & -- & \nHeldOutRight/\nHeldOut \\
\bottomrule
\end{tabular}
\end{table}

\begin{table}[H]
\centering\small
\caption{\textbf{Bench measurements the agent asked for after the first builds failed}, made with the parts themselves: tilting the course boards, and hanging bricks of known mass from a joint until it gives.}
\label{tab:measure}
\setlength{\tabcolsep}{3pt}\footnotesize
\begin{tabular}{@{}p{0.45\linewidth}p{0.49\linewidth}@{}}
\toprule
\rowhead Measurement & Result \\
\midrule
Stud pull-off: motor on 4 / 2 studs, hanging 2$\times$4 bricks & 515\,g (1.26\,N per stud) / about 214\,g (1.05\,N per stud) \\
Stud peel: long brick on 2 studs, hanging 1$\times$16 beams & 35--47\,g at 105\,mm (about 36--48\,N\,mm) \\
Reference car tilt test on the course board & slides at 23$^\circ$ and 15$^\circ$ (two orientations) \\
Plain brick tilt test on the course board & slides at 23$^\circ$; $\mu=0.42$ \\
Friction-pin torque: beams hung on a pinned joint & holds about 4.5\,N\,mm (24.5 assumed before) \\
Lowest speed level that starts a car & level 1 on the flat; level 3 (not 2) on 32$^\circ$ \\
Speed of the reference car (filmed) & about 1.3\,m/s at level 7, 1.0\,m/s at level 4 \\
\bottomrule
\end{tabular}
\end{table}

\begin{table}[H]
\centering\small
\caption{\textbf{What the simulator after the first round adds, and the real observation behind each change.}}
\label{tab:physics}
\setlength{\tabcolsep}{3pt}\footnotesize
\begin{tabular}{@{}p{0.42\linewidth}p{0.52\linewidth}@{}}
\toprule
\rowhead Physics added & Real observation it explains \\
\midrule
Stud joints release under overload and on impact & the battery box comes off on a steep ramp \\
Joints peel progressively & the battery stack peels off its stud joints on the steps \\
Loosely built bevel-on-spur meshes slip & the motor whines and the wheels never turn \\
Battery and receiver supply sag under load & fitted to the measured loaded speeds \\
Chains skip teeth above a load per tooth & set from a public filmed lifting test; slack rule \\
Free axles rub under radial load & cars teeter onto free middle wheels over the gap \\
Axles stop hard at their end stops; ramp board hinged at its top & calibration runs on the ramp \\
\bottomrule
\end{tabular}
\end{table}

\FloatBarrier
\section{Benchmark details}
\label{sec:supp_bench}

\FloatBarrier
\subsection{The courses}
\label{sec:supp_courses}
Each course is built for real from the same specification as its simulated twin.

\begin{figure}[H]
\centering
\includegraphics[width=\linewidth]{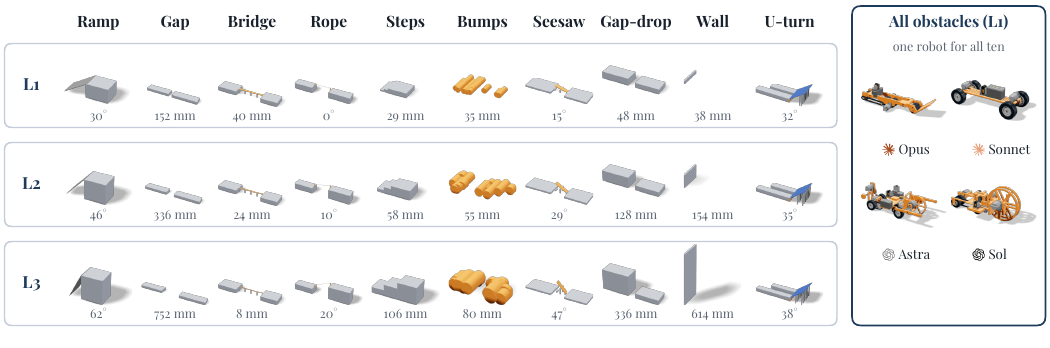}
\caption{\textbf{Ten obstacles at three levels, and one robot for all ten.} Each obstacle is drawn from the same viewpoint and at the same scale at all three levels; the number under each course is the dimension that changes with the level: ramp angle, gap width, plank width, cord incline, step height, rod height, seesaw tilt, drop to the landing deck, wall height and plate tilt. Levels also tighten the time limit. Right: each agent's single robot for all ten level-1 obstacles.}
\label{fig:courses}
\end{figure}

\FloatBarrier
\subsection{Inventory and building rules}
\label{sec:supp_inventory}

\begin{figure}[H]
\centering
\includegraphics[width=0.92\linewidth]{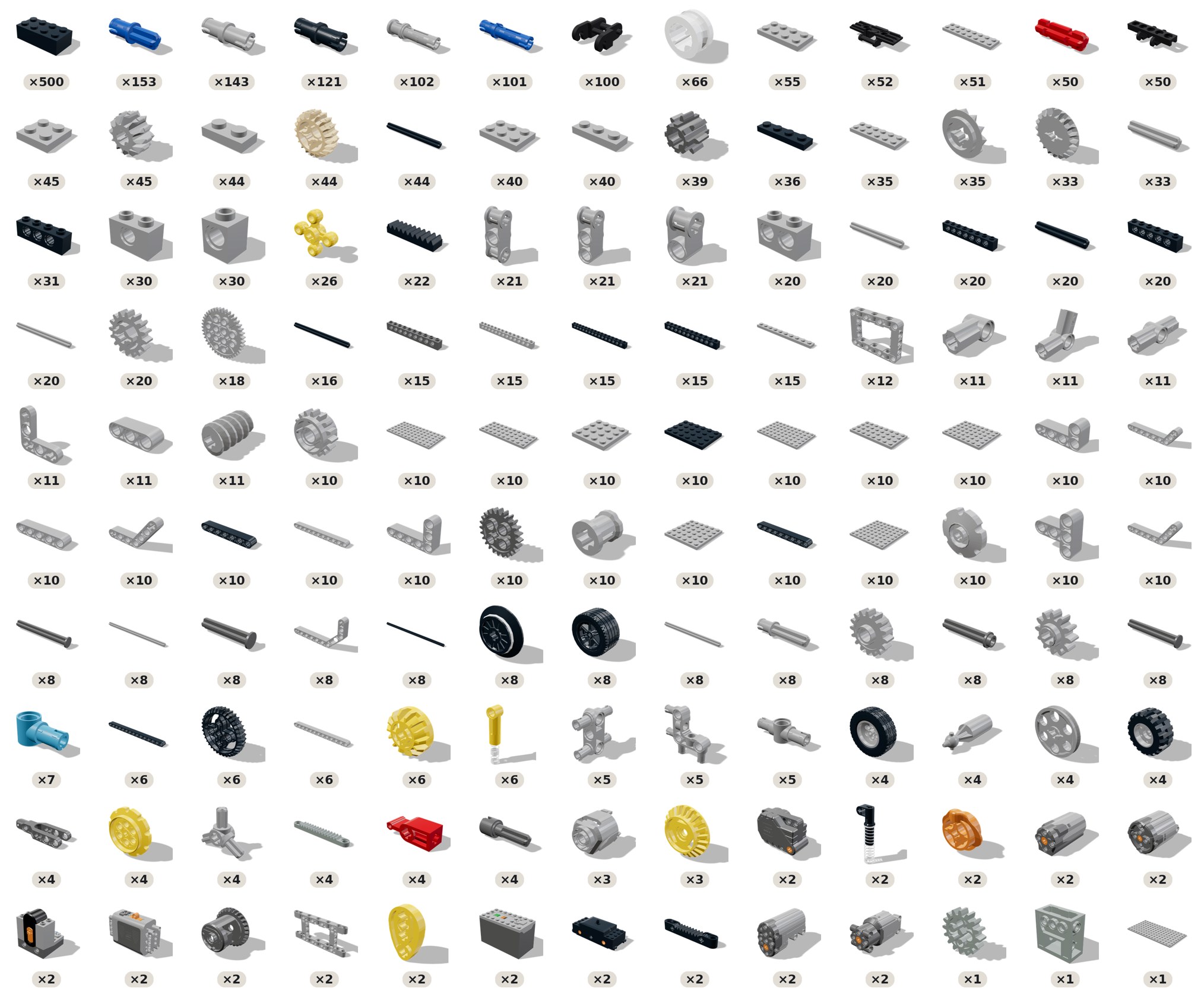}
\caption{\textbf{One box of parts.} The owner's stock of standard parts, one tile per part type with the number of pieces; agents may use only what is in the box, so every design can be built. Motors, receivers and the battery box are in the last rows.}
\label{fig:inventory_box}
\end{figure}
Each owned part type has full geometry (studs, holes, axle lengths, teeth, tyre friction) and mass. We will release the list of part numbers for the whole inventory, so that the parts can be bought and every robot built, together with instructions to build the obstacles.

\begin{table}[H]
\centering\footnotesize\setlength{\tabcolsep}{4pt}
\caption{\textbf{The inventory by family.}}
\label{tab:inventory}
\begin{tabular}{@{}lrr@{}}
\toprule
Family & types & pieces \\
\midrule
Beam bricks, liftarms and frames & 29 & 372 \\
Bricks and plates & 20 & 952 \\
Connectors & 13 & 121 \\
Axles & 15 & 263 \\
Pins & 8 & 640 \\
Bushes and CV ball & 3 & 80 \\
Gears, racks and differential & 21 & 353 \\
Shock absorbers & 2 & 8 \\
Wheels and sprockets & 8 & 46 \\
Chain and tread links, tyres & 4 & 206 \\
Motors & 6 & 10 \\
Battery boxes & 2 & 4 \\
IR receivers & 1 & 2 \\
\midrule
Total & 132 & 3{,}057 \\
\bottomrule
\end{tabular}
\end{table}

\paragraph{Electro-mechanics.} Two each of L (390\,rpm, 0.18\,N\,m stall), M (400\,rpm, 0.15\,N\,m), XL (220\,rpm, 0.40\,N\,m) and servo motors (\textpm90° in 7 steps per side), two pull-back spring motors (not remotely controlled); two AA and two AAA battery boxes ($\leq$3 motors each); two IR receivers with two outputs each ($\leq$4 sequenced motors). Motors share an output only if they always turn together; a motor at full power for the whole run may plug into a battery box. Open-loop programs: $\leq$200 timed commands, each one of 16 values: a speed level from $-7$ to $+7$ (0 floats) or brake (servo: level maps to angle).

\paragraph{Printed part.} In \taskPrint: one custom part type, up to four copies, a CadQuery script printed in PLA (35\% infill, walls $\geq$1.2\,mm) within a $256\times256\times256$\,mm build volume. It connects to catalogue parts only through declared interfaces (top-face studs, bottom-face anti-stud sockets, pin holes, cross holes, all on the 8\,mm grid), checked like catalogue connections. The \taskAll robot prints its whole structure: any number of part types, up to eight copies each, within 250\,mm. Real-build interface sizes, tuned with a fit coupon: studs 5.0\,mm, pin holes 5.0\,mm, cross holes $5.1\times2.1$\,mm, sockets 5.1\,mm.

\paragraph{Validator rules.} A design is rejected if it breaks any of the following (tolerances in brackets):
\begin{itemize}[nosep,leftmargin=1.2em]
\item \emph{inventory}: only owned parts, at most the owned count of each;
\item \emph{real joints}: every part connects to the rest through a stud joint (faces touching, \textpm0.3\,mm, studs on the 8\,mm grid), a pin or axle in a hole (on the hole's centre line, \textpm0.35\,mm, at least 4\,mm in) or a cross fit (coaxial, \textpm0.35\,mm); parts on one cross axle sit at multiples of 90° to each other;
\item \emph{no interpenetration}: no two parts overlap by more than 0.3\,mm (meshing gears and connected pairs excepted); unconnected parts clear the studs;
\item \emph{gears}: spur centre distance $(z_1+z_2)/2$ ($-0.15/+0.6$\,mm), a double bevel meshing a spur gear \textpm0.15\,mm, gears in one plane (\textpm1\,mm), no locked loops; worm and bevel geometry as specified;
\item \emph{chains and treads}: at most two chain loops per design; a chain fits its loop with at most half a link (3.2\,mm) of slack, with the sprockets at least $(z_1+z_2)/2+2$\,mm apart; tread loops with at most one link of slack;
\item \emph{electrics}: an IR receiver on board, at least one motor, every lead reaches its plug, every battery-box connector stays uncovered; at most four sequenced motor outputs;
\item \emph{open loop}: only timed commands, no sensor or position triggers.
\end{itemize}
A design review also blocks acceptance on critical findings: an axle or pin that can slide out (or a gear out of mesh), a sub-assembly held only by studs that pull off the same way, or a loaded group on a single hinge line. A pass requires the plain run and two nudged runs (1\,mm off the centreline; commands 50\,ms late) to reach the goal and stay there, upright and intact, for 5\,s.

\providecommand{\promptboxdefined}{}
\newenvironment{promptbox}[1]{\par\smallskip\noindent{\small\textbf{#1}}\par\nobreak\vspace{1pt}\hrule\vspace{3pt}\begingroup\footnotesize\setlength{\parskip}{3pt}\setlength{\parindent}{0pt}\raggedright}{\par\endgroup\vspace{2pt}\hrule\smallskip}

\FloatBarrier
\subsection{Prompts and documents}
\label{sec:supp_prompts}
Agents work in a sandbox without web access, with the task prompt, a JSON file of the level (geometry, start, goal, time limit) and reference documents: rules, obstacles, inventory, design-file format and Python library, connection, mechanism and design-review rules, motor program and wiring, simulator, and an example design; the redesign round adds the failed design with its tries, the simulator source and parameter documentation, and a loop description. Prompts are verbatim except: braces mark per-task values (budgets in \cref{sec:benchmark}), brand and printer names are neutral, and [\dots] marks omitted passages.

\begin{promptbox}{Task prompt (\taskCross)}
You are a robot design engineer: you design both the hardware and its control. Using only the brick parts in the owner's inventory, design an assembly (parts, poses and connections, with any mechanisms or tools it needs) and its motor program that crosses \textbf{\{bench\} level \{level\}} of the brick obstacle benchmark: \{description\}

\textbf{Your goal: pass this level in the simulator.} Think about the obstacle and come up with a smart, innovative robot that solves it: you decide what kind of machine it is and how it moves, and clever mechanisms are welcome. Do not settle for the first idea that could work. The design will then be built from the real parts and run on the real course with the same commands, open loop: nobody steers it, and the battery level, tyre slip and exact start pose vary from run to run. Before settling on a design, explore both hardware options (the vehicle's mechanics and geometry) and software options (the motor program), and weigh their pros and cons for how reliably each would repeat in the real world, not only whether it passes once in the simulator.

[\dots]

\textbf{How to work}

\begin{enumerate}[nosep,leftmargin=1.2em]
\item Read the rules and the task.
\item Think about what crossing the obstacle physically takes and what would stop a simple vehicle. Then go through the whole inventory in \texttt{parts.md} and the mechanisms in \texttt{manufacturing\_constraints.md}, part by part, and ask what each one could do for this obstacle: the owner has many more kinds of parts than a car needs, and the best solution may use one you would not reach for first. Sketch several quite different concepts (different ways of moving, mechanisms, structures) and compare them before choosing one deliberately; check it with simple physics.
\item Build it with a generator script; validate it.
\item Simulate it. Diagnose any failure from the numbers and both camera views before changing anything; make each run answer a clear question.
\item Iterate until it passes (use more pieces if needed), then submit.
\end{enumerate}

Within the constraints you are free: any parts in the inventory, any number of motors, any way of moving and any mechanism (wheels are not required). Aim for a design an expert would call smart: one that uses the obstacle, the parts and the physics cleverly, not just the most obvious vehicle made bigger or stronger.

Work autonomously; nobody will answer questions. Do not stop until a design is accepted or your simulations are used up: stopping early is not allowed, and unused simulations and time earn nothing. \textbf{Always end with a submission:} if nothing has passed once your simulations are used up, submit your furthest-progress design with \texttt{--best-attempt}. Further requests may follow in this session.
\end{promptbox}

\begin{promptbox}{\taskPrint: extension after a failed \taskCross}
No design was accepted yet, so you get an extension: \textbf{\{sims\} more simulations} and \{hours\} h (\texttt{brick-bench budget}). The goal is the same: pass this level.

\textbf{New: you may design one custom 3D-printed part} to help. It is one part type, used in up to \{copies\} copies in your design, printed in PLA, fitting the printer's build volume (\{max\_mm\} × \{max\_mm\} × \{max\_mm\} mm), with walls of at least \{wall\} mm. You write it as a CadQuery script and declare how it connects to the bricks (studs, sockets, pin holes, cross holes). \texttt{/work/docs/custom\_part.md} explains how; \texttt{brick-bench part <script.py>} builds it, checks it, exports the STL that will be printed, and gives you its part id to use in your design like any other part.

Use what you learned so far: think about why your attempts failed, and whether a printed part (or something else) removes that cause. Submit as before, and as before always end with a submission (\texttt{brick-bench submit --best-attempt <file>} if nothing passes in this extension).
\end{promptbox}

\begin{promptbox}{\taskLight: follow-up after an accepted design}
Your design \texttt{/work/\{file\}} was accepted: it passes this level (\{mass\_g\} g, \{n\_parts\} pieces, it reaches the goal at \{time\_s\} s).

\textbf{New request: the lightest design that still passes.} Minimise the total mass of the assembly (\texttt{brick-bench validate} gives \texttt{mass\_g}). Every gram counts. The design must still pass exactly as before: a simulator pass, the nudged runs and no critical design-review finding.

Genuinely try to make it lighter; do not just hand back what you have. Your passing design is a starting point, not a constraint: you may reconsider the whole concept and explore very different designs (another way of moving, another drive, a different structure), as long as the new design passes and is lighter. Think about where the mass actually goes and what each part is really needed for before you simulate.

\begin{itemize}[nosep,leftmargin=1.2em]
\item You have \textbf{\{sims\} new simulations} and \{hours\} h for this request (\texttt{brick-bench budget}).
\item Work on new files (for example \texttt{/work/light.json}); \texttt{/work/accepted/} keeps your accepted designs.
\item Submit with \texttt{brick-bench submit <file>}. The design you get accepted is the one that counts, and after an acceptance the harness moves on, so submit when you have the lightest passing design you can reach. If nothing lighter passes, you may submit your passing design unchanged.
\end{itemize}
\end{promptbox}

\begin{promptbox}{\taskFast: follow-up}
\textbf{New request: the fastest crossing that still passes.} Your passing design \texttt{/work/\{file\}} reaches the goal at \{time\_s\} s. Minimise the time at which the vehicle reaches the goal (\texttt{time\_s} of a simulation). Mass and pieces do not matter for this request. The design must still pass exactly as before: a simulator pass, the nudged runs and no critical design-review finding.

Genuinely try to make it faster; do not just hand back what you have. Your passing design is a starting point, not a constraint: you may reconsider the whole concept and explore very different designs (gearing, wheels, the way it takes the obstacle, the start gap, the motor program), as long as the new design passes and crosses faster. Think about what limits its speed now before you simulate.

\begin{itemize}[nosep,leftmargin=1.2em]
\item You have \textbf{\{sims\} new simulations} and \{hours\} h for this request (\texttt{brick-bench budget}).
\item Work on new files (for example \texttt{/work/fast.json}); \texttt{/work/accepted/} keeps your accepted designs.
\item Submit with \texttt{brick-bench submit <file>}. The design you get accepted is the one that counts, and after an acceptance the harness moves on, so submit when you have the fastest passing design you can reach. If nothing faster passes, you may submit your passing design unchanged.
\end{itemize}
\end{promptbox}

\begin{promptbox}{\taskAll: task prompt (excerpt)}
You are a robot design engineer: you design both the hardware and its control. Your task: \textbf{one robot that crosses as many as possible of the 10 level-1 obstacles} of the brick obstacle benchmark (ramp, gap, bridge, rope, steps, bumps, seesaw, gap-drop, wall, U-turn), the same physical robot on every one. Your score is the number of benches the robot you hand in passes.

\textbf{What is different in this task (read \texttt{/work/docs/allbench.md} for the details)}

\begin{itemize}[nosep,leftmargin=1.2em]
\item \textbf{One assembly, one motor program per bench.} The design file has \texttt{motor\_programs}: \texttt{\{"ramp": [...], "gap": [...], ...\}} instead of one \texttt{motor\_program}. The hardware is shared; each bench can have its own commands (timing, speeds, which motors run).
\item \textbf{The structure is 3D printed.} No bricks, beam bricks, beams or plates from the inventory. You may use the inventory's mechanical and electrical parts: gears, axles, pins, connectors, bushes, wheels and tyres, chains and treads, racks, shock absorbers, motors, battery boxes and IR receivers. Everything else (chassis, frames, brackets, arms, wedges, mounts) you design as printed parts: \textbf{any number of part types}, up to \{copies\} copies of each, each fitting the printer's build plate: at most \textbf{\{max\_mm\} mm} in every direction, walls of at least \{wall\} mm (\texttt{custom\_part.md} explains how; \texttt{brick-bench part <script.py>} builds and checks a part and gives you its id).
\item \textbf{Electrics:} up to two IR receivers and as many motors as they can control: each receiver output drives one motor, or several motors whose plugs stack because they always turn together; a motor that runs at full power the whole run can plug straight into a battery box (\texttt{motor\_program.md}).
\item \textbf{Budget:} \{sims\} simulations and \{hours\} h for the whole task. Each simulation runs the robot on \textbf{one} bench: \texttt{brick-bench simulate <file> --bench <name>}.
\end{itemize}

[\dots]
\end{promptbox}

\begin{promptbox}{Redesign round (excerpt)}
\textbf{Your goal: a robot that crosses this level on the real course.} It must pass this level in the simulator first; it is then built and filmed. Your earlier design for this level passed in the simulator but failed on the real course: study it in \texttt{/work/evidence/} before you start. Think about the obstacle and come up with a smart, innovative robot that solves it: you decide what kind of machine it is and how it moves, and clever mechanisms are welcome. Do not settle for the first idea that could work. The design will then be built from the real parts and run on the real course with the same commands, open loop: nobody steers it, and the battery level, tyre slip and exact start pose vary from run to run. Before settling on a design, explore both hardware options (the vehicle's mechanics and geometry) and software options (the motor program), and weigh their pros and cons for how reliably each would repeat in the real world, not only whether it passes once in the simulator.

[\dots]

\begin{enumerate}[nosep,leftmargin=1.2em]
\item \textbf{Diagnose.} Read the rules, the task and \texttt{docs/loop\_process.md}. Study your earlier design and its filmed tries (the result, the contact sheets, the videos) and work out what happened on the real course: what was seen, when, which component, and whether it is a design flaw, something the simulator gets wrong, or a build or operator problem. Go through every try closely (\texttt{brick-sim frames}, \texttt{brick-sim compare}; \texttt{docs/loop\_process.md}, "Watching the tries"), then write your failure record and register it with \texttt{brick-sim record}.
\item \textbf{Rules.} Write \texttt{/work/loop/design\_rules.md} (general design lessons from this failure) and \texttt{/work/loop/physics\_claims.md} (what the simulator gets wrong, if anything, and the evidence). Keep them up to date as you learn more.
\item \textbf{Simulator (optional).} If the simulator misses the detail that made your design fail for real, add it: edit your simulator files, check them, and show with \texttt{brick-sim run} that your earlier design now fails the way it did for real. Then \texttt{brick-sim submit} it to the gate. If the gate accepts it, it becomes your simulator for the rest of the task. Register with \texttt{brick-sim uncertainty} how sure you are of the values you fitted from your few tries.
\end{enumerate}

[\dots]
\end{promptbox}

\begin{promptbox}{Closing turn (lessons for the next task)}
The task is over: simulations and submissions are closed. One last thing.

Write \textbf{\texttt{/work/lessons.md}} for your future self: your next tasks (the other levels of these benches) will start with it in \texttt{/work/memory/}, next to your accepted designs and their results. Keep it short (at most about 400 words) and concrete:
\begin{itemize}[nosep,leftmargin=1.2em]
\item \textbf{what failed}, and why (the cause, not just the symptom);
\item \textbf{what worked};
\item \textbf{design rules} you learned about these benches, these parts and this simulator (numbers where you have them).
\end{itemize}

Write the file, then stop.
\end{promptbox}

\FloatBarrier
\subsection{Size of the design space}
\label{sec:supp_space}
Orders of magnitude. \emph{Connection model} (headline): each new part picks one of $T{=}132$ types, $S{=}10$ free sites and $O{=}4$ orientations, orderings divided out: $(TSO)^N/N!$ is $10^{56.1}$, $10^{67.9}$, $10^{121.6}$, $10^{214.3}$ for $N = 20$, 25, 50, 100 (collisions and invalid joints ignored; only ten sites and no chain lengths). \emph{Grid model}: 22{,}200 cells (8\,mm by 3.2\,mm in $240\times160\times120$\,mm) and 24 orientations give $10^{172}$ at 25 pieces. \emph{Programs}: 0.1\,s grid, 30\,s run, 16 levels; $k$ commands per motor give $\binom{300}{k}16^k$, $10^{16.3}$ for $k{=}5$. Passing robots use 20 to 398 pieces.

\FloatBarrier
\section{Limitations and future work}
\label{sec:supp_limits}

\paragraph{Simulator.} The simulator is learned in the sense of a training and a validation set: parameters are fitted and missing physical effects modelled on training runs, then checked on validation runs, which also guided which changes were kept. The observations are limited (one fixed camera per course, finite frame rate and resolution, perspective, a few bench measurements, no direct force or contact data), so some parameters are not identifiable and other settings might fit as well. It predicts \nHeldOutRight{} of \nHeldOut{} validation runs and \nRealRateAll\% of its passes on new designs cross for real, which makes it a physics-grounded environment for testing agents. Each loop correction is fitted to one failure in that agent's simulator copy and not checked against other filmed runs; the loop's goal is a working redesign, not a shared simulator.

\paragraph{Real tests.} Only simulated passes are built, because a design that fails in simulation most likely fails for real; designs that fail in simulation but would cross are not measured. Each design is built once, so assembly errors and build-to-build variation are not separated from try-to-try variation, and small differences between simulated and real obstacles remain. \taskLight and \taskFast designs are not built, each agent runs each task once, and the real tests took an estimated \nWorkHours{} hours. The benchmark score is computed in the simulator, on all 30 obstacle levels and the \taskAll task; parts and obstacle instructions are released for real tests.

The search baselines are a reference over the full inventory, not the strongest possible search.

\paragraph{Future work.} A robot in the loop that builds and tests designs as a tool agents call; sensors and feedback control co-designed with the mechanism; one simulator fitted to all real failures, with instrumented measurements and multi-view or high-speed video; repeated runs per task; and requirement-driven mechatronic systems specified by function, cost, mass and safety.

\FloatBarrier
\section{Release and responsibility}
\label{sec:supp_release}

\FloatBarrier
\subsection{Dataset card}
\label{sec:supp_datacard}
\paragraph{Contents.} Courses (simulated twins, build plans), task definitions and budgets, inventory, validator; simulator source, parameters before and after the first real round, bench checks; for every task and agent, all simulated designs and programs, verdicts, contact sheets, tokens, wall times and accepted designs (\nGeoAgents{} robots from \nSimsAgents{} simulations); \nVidTotal{} trimmed real videos paired with their simulations, trimmed and paired with their simulations, per-try verdicts and the loop's records, corrections and redesigns; printed parts (CadQuery, STL) and build packages.
\paragraph{Formats.} Designs and programs are JSON (placed parts, connections, gear meshes, chains, timed motor commands); simulations are JSON results with MP4 renders; real videos are 4K, 30\,fps H.264 MP4 without audio; printed parts are Python (CadQuery) and STL; verdicts and manifests are JSON and CSV.
\paragraph{Alignment of real and simulated runs.} Each real clip is trimmed from just before the robot moves to just after it stops or crosses, and is aligned with its simulation at motion onset.
\paragraph{Licence.} Code will be released under the MIT licence and data under CC BY 4.0.
\paragraph{Intended use and limitations.} Evaluating and training agents that design, simulate and test machines, and studying the sim-to-real gap. Robots are small and slow; real tests cover every simulation pass and the loop levels 1 and 2; few tries per design make real verdicts coarse. No person or hand is in frame during a run. Parts have brand-neutral names.

\paragraph{Potential negative societal impacts.} Agents that design machines with little human input could, on larger hardware, produce devices that fail unsafely or serve harmful uses without a human understanding the design. Our robots are small, slow and battery powered and pass simulation and a human build review before building; full-size machines would need much stronger checks. The benchmark measures capability, not safety.

\FloatBarrier

%% file: sec/tab_expanded.tex
\begin{table}[H]\centering\scriptsize
\caption{\textbf{Every task, every agent.} Pass: simulation at which the task passed ($^{\mathrm{P}}$: in \taskPrint), or \textcolor{rbBad}{\xmark} if it failed in simulation; \textsc{Light}: lightest mass (g); \textsc{Fast}: fastest time (s); Real: crosses (\cmark) or fails (\xmark) for real, $\rightarrow$ loop redesign (\xmark\ also when no redesign passed the vote); --: none (\taskLight, \taskFast: no design; Real: not built, as the task failed in simulation). $^\ddagger$See \cref{sec:supp_detailed}.}
\label{tab:expanded}
\setlength{\tabcolsep}{2pt}
\resizebox{\linewidth}{!}{%
\begin{tabular}{@{}llcccccccccccccccc@{}}
\toprule
\rowhead & & \multicolumn{4}{c}{\opusL} & \multicolumn{4}{c}{\sonnetL} & \multicolumn{4}{c}{\astraL} & \multicolumn{4}{c}{\solL} \\
\cmidrule(lr){3-6}\cmidrule(lr){7-10}\cmidrule(lr){11-14}\cmidrule(lr){15-18}
\rowhead Obstacle & L & pass & \textsc{Light} & \textsc{Fast} & real & pass & \textsc{Light} & \textsc{Fast} & real & pass & \textsc{Light} & \textsc{Fast} & real & pass & \textsc{Light} & \textsc{Fast} & real \\
\midrule
Ramp & L1 & 5 & 178 & 0.90 & \cmark & 4 & 184 & 1.02 & \cmark & 4 & 181 & 3.14 & \cmark & 3 & 184 & 2.61 & \cmark \\
 & L2 & \textcolor{rbBad}{\xmark} & -- & 1.42 & -- & \textcolor{rbBad}{\xmark} & -- & -- & -- & \textcolor{rbBad}{\xmark} & -- & -- & -- & \textcolor{rbBad}{\xmark} & -- & -- & -- \\
 & L3 & 45$^{\mathrm{P}}$ & 757 & 2.58 & \xmark & \textcolor{rbBad}{\xmark} & -- & -- & -- & 39$^{\mathrm{P}}$ & 591 & 4.54 & \xmark & \textcolor{rbBad}{\xmark} & -- & -- & -- \\
\addlinespace[1pt]
Gap & L1 & 2 & 205 & 0.59 & \cmark & 12 & 210 & 0.78 & \cmark & 6 & 226 & 1.07 & \cmark & 5 & 206 & 1.34 & \cmark \\
 & L2 & 3 & 226 & 1.31 & \cmark & 1 & 300 & 3.09 & \cmark & 10 & 700 & 4.78 & \cmark & 8 & 278 & 7.08 & \cmark \\
 & L3 & 28 & 1065 & 11.74 & \xmark & \textcolor{rbBad}{\xmark} & -- & -- & -- & \textcolor{rbBad}{\xmark} & -- & -- & -- & \textcolor{rbBad}{\xmark} & -- & -- & -- \\
\addlinespace[1pt]
Bridge & L1 & 2 & 172 & 0.69 & \cmark & 9 & 166 & 1.05 & \cmark & 4 & 168 & 0.96 & \xmark$\rightarrow$\cmark & 3 & 176 & 1.21 & \cmark \\
 & L2 & 2 & 172 & 0.69 & \cmark & 6 & 185 & 0.81 & \xmark$\rightarrow$\cmark & 4 & 195 & 5.02 & \xmark$\rightarrow$\cmark & 7 & 220 & 1.83 & \cmark \\
 & L3 & 5 & 177 & 0.69 & \cmark & 1 & 173 & 1.21 & \xmark & 7 & 189 & 1.44 & \cmark & 10 & 269 & 2.73 & \xmark \\
\addlinespace[1pt]
Rope & L1 & 8 & 233 & 1.71 & \cmark & 19 & 460 & 5.64 & \xmark$\rightarrow$\cmark & 11 & 526 & 7.67 & \xmark$\rightarrow$\cmark & 9 & 560 & 8.13 & \xmark$\rightarrow$\xmark \\
 & L2 & 8 & 431 & 1.51 & \xmark$\rightarrow$\cmark & \textcolor{rbBad}{\xmark} & -- & 4.00 & -- & 6 & 520 & 8.72 & \cmark & 59$^{\mathrm{P}}$ & 781 & 7.60 & \xmark$\rightarrow$\xmark \\
 & L3 & \textcolor{rbBad}{\xmark} & -- & -- & -- & \textcolor{rbBad}{\xmark} & -- & -- & -- & \textcolor{rbBad}{\xmark} & -- & -- & -- & 42$^{\mathrm{P}}$ & 778 & 11.98 & \cmark \\
\addlinespace[1pt]
Steps & L1 & 8 & 208 & 3.44 & \xmark$\rightarrow$\xmark & 28 & 498 & 3.20 & \cmark & 27 & 458 & 7.18 & \cmark & 23 & 208 & 5.54 & \xmark$\rightarrow$\cmark \\
 & L2 & 9 & 367 & 3.82 & \xmark$\rightarrow$\xmark & \textcolor{rbBad}{\xmark} & -- & -- & -- & 23 & 649 & 5.40 & \cmark & 16 & 653 & 10.48 & \xmark$\rightarrow$\cmark \\
 & L3 & 23 & 544 & 8.70 & \xmark & 46$^{\mathrm{P}}$ & 739 & 11.88 & \xmark & \textcolor{rbBad}{\xmark} & -- & -- & -- & 22 & 822 & 19.33 & \xmark \\
\addlinespace[1pt]
Bumps & L1 & 19 & 489 & 1.46 & \cmark & 20 & 249 & 5.54 & \cmark & 27 & 421 & 1.69 & \cmark & 9 & 419 & 1.11 & \cmark \\
 & L2 & 22 & 532 & 12.93 & \xmark$\rightarrow$\cmark & \textcolor{rbBad}{\xmark} & 270 & -- & -- & 24 & 352 & 13.23 & \cmark & \textcolor{rbBad}{\xmark} & -- & -- & -- \\
 & L3 & 54$^{\mathrm{P}}$ & 1037 & 1.19 & \cmark & \textcolor{rbBad}{\xmark} & 302 & -- & -- & 24 & 418 & 10.56 & \xmark & 18 & 467 & 3.54 & \cmark \\
\addlinespace[1pt]
Seesaw & L1 & 9 & 183 & 0.70 & \cmark & 30 & 197 & 1.17 & \xmark$\rightarrow$\cmark & 9 & 291 & 5.03 & \cmark & 10 & 371 & 4.43 & \cmark \\
 & L2 & 9 & 219 & 1.04 & \cmark & 9 & 209 & 4.64 & \cmark & 20 & 420 & 7.13 & \cmark & 24 & 440 & 5.17 & \cmark \\
 & L3 & 11 & 595 & 8.22 & \cmark & \textcolor{rbBad}{\xmark} & -- & -- & -- & \textcolor{rbBad}{\xmark} & -- & -- & -- & \textcolor{rbBad}{\xmark} & -- & -- & -- \\
\addlinespace[1pt]
Gap-drop & L1 & 2 & 207 & 0.59 & \cmark & 18 & 298 & 0.54 & \cmark & 11 & 240 & 3.94 & \cmark & 17 & 214 & 1.62 & \cmark \\
 & L2 & 2 & 201 & 0.54 & \cmark & 18 & 221 & 0.66 & \cmark & 10 & 213 & 4.37 & \cmark & 9 & 367 & 1.64 & \cmark \\
 & L3 & \textcolor{rbBad}{\xmark} & -- & 0.42 & -- & 4 & 470 & 1.98 & \cmark & 16 & 549 & 0.91 & \cmark & 15 & 366 & 1.06 & \xmark \\
\addlinespace[1pt]
Wall & L1 & 18 & 462 & 0.89 & \xmark$\rightarrow$\cmark & 24 & 330 & 0.77 & \cmark & 20 & 459 & 2.22 & \cmark & 10 & 633 & 2.27 & \xmark$\rightarrow$\cmark \\
 & L2 & 10 & 454 & 4.83 & \xmark$\rightarrow$\xmark & \textcolor{rbBad}{\xmark} & -- & -- & -- & 31$^{\mathrm{P}}$ & 389 & 2.93 & \xmark$\rightarrow$\cmark & \textcolor{rbBad}{\xmark} & -- & -- & -- \\
 & L3 & \textcolor{rbBad}{\xmark} & -- & -- & -- & \textcolor{rbBad}{\xmark} & -- & -- & -- & \textcolor{rbBad}{\xmark} & -- & -- & -- & 47$^{\mathrm{P}}$ & 989 & 2.59 & \xmark \\
\addlinespace[1pt]
U-turn & L1 & 22 & 214 & 1.19 & \cmark & \textcolor{rbBad}{\xmark}$^\ddagger$ & 237 & -- & \xmark$^\ddagger\!\rightarrow$\cmark & 51$^{\mathrm{P}}$ & 271 & 1.87 & \cmark & 30 & 227 & 5.15 & \cmark \\
 & L2 & \textcolor{rbBad}{\xmark} & 242 & 1.77 & -- & \textcolor{rbBad}{\xmark} & 365 & 4.34 & -- & 1 & 333 & 4.31 & \cmark & 4 & 255 & 5.23 & \cmark \\
 & L3 & \textcolor{rbBad}{\xmark} & -- & 2.85 & -- & \textcolor{rbBad}{\xmark} & -- & -- & -- & \textcolor{rbBad}{\xmark} & 346 & 7.76 & -- & \textcolor{rbBad}{\xmark} & 298 & -- & -- \\
\bottomrule
\end{tabular}}
\end{table}

%% file: sec/tab_robust.tex
\begin{table}[H]\centering\footnotesize
\caption{\textbf{The ranking does not depend on how the score is aggregated.} Mean: headline score; geo.: geometric mean; rank: mean rank over the four components (1 = best); 95\% CI: bootstrap over obstacles (\cref{sec:supp_detailed}).}
\label{tab:robust}
\setlength{\tabcolsep}{2pt}
\begin{tabular}{@{}lcccccccc@{}}
\toprule
\rowhead Method & \textsc{Cross/Print} & \textsc{All} & \textsc{Light} & \textsc{Fast} & mean & geo. & rank & 95\% CI \\
\midrule
Random search & 0.03 & 0.00 & 0.00 & 0.00 & 0.8 & 1.4 & 7.1 & -- \\
CMA-ES & 0.00 & 0.00 & 0.00 & 0.00 & 0.0 & 1.0 & 7.6 & -- \\
Template search & 0.00 & 0.00 & 0.00 & 0.00 & 0.0 & 1.0 & 7.6 & -- \\
Genetic algorithm & 0.03 & 0.00 & 0.02 & 0.01 & 1.7 & 1.7 & 5.9 & -- \\
Single LLM call & 0.07 & 0.00 & 0.00 & 0.00 & 1.7 & 1.6 & 6.8 & -- \\
\midrule
\opus & 0.80 & 0.50 & 0.75 & 0.90 & 73.9 & 72.2 & 1.4 & 60.9--87.1 \\
\sonnet & 0.53 & 0.30 & 0.57 & 0.37 & 44.3 & 42.8 & 3.5 & 31.9--58.1 \\
\astra & 0.77 & 0.60 & 0.60 & 0.34 & 57.6 & 55.2 & 2.4 & 48.0--67.5 \\
\sol & 0.77 & 0.50 & 0.62 & 0.32 & 55.1 & 52.5 & 2.8 & 43.8--66.7 \\
\bottomrule
\end{tabular}
\end{table}

%% file: main.bib
@article{nehme2026stepcad,
  title={StepCAD: Mesh-to-CAD Code Generation via LLM Policy and Geometry-Guided Search},
  author={Nehme, Ghadi and Ahmed, Faez},
  journal={arXiv preprint arXiv:2610.03799},
  year={2026}
}

@article{nehme2026cadfit,
  title={CADFit: Precise Mesh-to-CAD Program Generation with Hybrid Optimization},
  author={Nehme, Ghadi and Whalen, Eamon and Ahmed, Faez},
  journal={arXiv preprint arXiv:2605.01171},
  year={2026}
}

@article{man2025videocad,
  title={VideoCAD: A Dataset and Model for Learning Long-Horizon 3D CAD UI Interactions from Video},
  author={Man, King Yiu Brandon and Nehme, Ghadi and Alam, Md Ferdous and Ahmed, Faez},
  journal={Advances in Neural Information Processing Systems},
  volume={38},
  year={2026}
}

@article{makatura2023can,
  title={How can large language models help humans in design and manufacturing? Part 2: Synthesizing an end-to-end LLM-enabled design and manufacturing workflow},
  author={Makatura, Liane and Foshey, Michael and Wang, Bohan and H{\"a}hnLein, Felix and Ma, Pingchuan and Deng, Bolei and Tjandrasuwita, Megan and Spielberg, Andrew and Owens, Crystal and Chen, Peter Yichen and others},
  journal={Harvard Data Science Review},
  volume={5},
  year={2024},
  publisher={The MIT Press}
}

@article{picard2025concept,
  title={From concept to manufacturing: Evaluating vision-language models for engineering design},
  author={Picard, Cyril and Edwards, Kristen M and Doris, Anna C and Man, Brandon and Giannone, Giorgio and Alam, Md Ferdous and Ahmed, Faez},
  journal={Artificial Intelligence Review},
  volume={58},
  number={9},
  pages={288},
  year={2025},
  publisher={Springer}
}

@article{khan2024text2cad,
  title={Text2cad: Generating sequential cad designs from beginner-to-expert level text prompts},
  author={Khan, Mohammad S and Sinha, Sankalp and Sheikh, Talha U and Stricker, Didier and Ali, Sk A and Afzal, Muhammad Z},
  journal={Advances in Neural Information Processing Systems},
  volume={37},
  pages={7552--7579},
  year={2024}
}

@inproceedings{elrefaie2025agents,
  title={AI agents in engineering design: a multi-agent framework for aesthetic and aerodynamic car design},
  author={Elrefaie, Mohamed and Qian, Janet and Wu, Raina and Chen, Qian and Dai, Angela and Ahmed, Faez},
  booktitle={International Design Engineering Technical Conferences and Computers and Information in Engineering Conference},
  volume={89237},
  pages={V03BT03A048},
  year={2025},
  organization={American Society of Mechanical Engineers}
}

@article{wang2023voyager,
  title={Voyager: An open-ended embodied agent with large language models},
  author={Wang, Guanzhi and Xie, Yuqi and Jiang, Yunfan and Mandlekar, Ajay and Xiao, Chaowei and Zhu, Yuke and Fan, Linxi and Anandkumar, Anima},
  journal={arXiv preprint arXiv:2305.16291},
  year={2023}
}

@inproceedings{liang2023codeaspolicies,
  title={Code as policies: Language model programs for embodied control},
  author={Liang, Jacky and Huang, Wenlong and Xia, Fei and Xu, Peng and Hausman, Karol and Ichter, Brian and Florence, Pete and Zeng, Andy},
  booktitle={2023 IEEE International conference on robotics and automation (ICRA)},
  pages={9493--9500},
  year={2023},
  organization={IEEE}
}

@inproceedings{sims1994evolving,
  title={Evolving virtual creatures},
  author={Sims, Karl},
  booktitle={Seminal Graphics Papers: Pushing the Boundaries, Volume 2},
  pages={699--706},
  year={2023}
}

@article{bhatia2021evolution,
  title={Evolution gym: A large-scale benchmark for evolving soft robots},
  author={Bhatia, Jagdeep and Jackson, Holly and Tian, Yunsheng and Xu, Jie and Matusik, Wojciech},
  journal={Advances in Neural Information Processing Systems},
  volume={34},
  pages={2201--2214},
  year={2021}
}

@article{zhao2020robogrammar,
  title={Robogrammar: graph grammar for terrain-optimized robot design},
  author={Zhao, Allan and Xu, Jie and Konakovi{\'c}-Lukovi{\'c}, Mina and Hughes, Josephine and Spielberg, Andrew and Rus, Daniela and Matusik, Wojciech},
  journal={ACM Transactions on Graphics (TOG)},
  volume={39},
  number={6},
  pages={1--16},
  year={2020},
  publisher={ACM, New York, NY, USA}
}

@article{kriegman2020scalable,
  title={A scalable pipeline for designing reconfigurable organisms},
  author={Kriegman, Sam and Blackiston, Douglas and Levin, Michael and Bongard, Josh},
  journal={Proceedings of the National Academy of Sciences},
  volume={117},
  number={4},
  pages={1853--1859},
  year={2020},
  publisher={National Academy of Sciences}
}

@article{wang2019neural,
  title={Neural graph evolution: Towards efficient automatic robot design},
  author={Wang, Tingwu and Zhou, Yuhao and Fidler, Sanja and Ba, Jimmy},
  journal={arXiv preprint arXiv:1906.05370},
  year={2019}
}

@article{yuan2022transform2act,
  title={Transform2act: Learning a transform-and-control policy for efficient agent design},
  author={Yuan, Ye and Song, Yuda and Luo, Zhengyi and Sun, Wen and Kitani, Kris},
  journal={arXiv preprint arXiv:2110.03659},
  year={2021}
}

@article{gupta2021embodied,
  title={Embodied intelligence via learning and evolution},
  author={Gupta, Agrim and Savarese, Silvio and Ganguli, Surya and Fei-Fei, Li},
  journal={Nature communications},
  volume={12},
  number={1},
  pages={5721},
  year={2021},
  publisher={Nature Publishing Group UK, London}
}

@inproceedings{schaff2019jointly,
  title={Jointly learning to construct and control agents using deep reinforcement learning},
  author={Schaff, Charles and Yunis, David and Chakrabarti, Ayan and Walter, Matthew R},
  booktitle={2019 international conference on robotics and automation (ICRA)},
  pages={9798--9805},
  year={2019},
  organization={IEEE}
}

@inproceedings{tobin2017domain,
  title={Domain randomization for transferring deep neural networks from simulation to the real world},
  author={Tobin, Josh and Fong, Rachel and Ray, Alex and Schneider, Jonas and Zaremba, Wojciech and Abbeel, Pieter},
  booktitle={2017 IEEE/RSJ international conference on intelligent robots and systems (IROS)},
  pages={23--30},
  year={2017},
  organization={IEEE}
}

@inproceedings{peng2018sim,
  title={Sim-to-real transfer of robotic control with dynamics randomization},
  author={Peng, Xue Bin and Andrychowicz, Marcin and Zaremba, Wojciech and Abbeel, Pieter},
  booktitle={2018 IEEE international conference on robotics and automation (ICRA)},
  pages={3803--3810},
  year={2018},
  organization={IEEE}
}

@article{tan2018sim,
  title={Sim-to-real: Learning agile locomotion for quadruped robots},
  author={Tan, Jie and Zhang, Tingnan and Coumans, Erwin and Iscen, Atil and Bai, Yunfei and Hafner, Danijar and Bohez, Steven and Vanhoucke, Vincent},
  journal={arXiv preprint arXiv:1804.10332},
  year={2018}
}

@article{ramos2019bayessim,
  title={Bayessim: adaptive domain randomization via probabilistic inference for robotics simulators},
  author={Ramos, Fabio and Possas, Rafael Carvalhaes and Fox, Dieter},
  journal={arXiv preprint arXiv:1906.01728},
  year={2019}
}

@inproceedings{du2021auto,
  title={Auto-tuned sim-to-real transfer},
  author={Du, Yuqing and Watkins, Olivia and Darrell, Trevor and Abbeel, Pieter and Pathak, Deepak},
  booktitle={2021 IEEE International Conference on Robotics and Automation (ICRA)},
  pages={1290--1296},
  year={2021},
  organization={IEEE}
}

@inproceedings{todorov2012mujoco,
  title={Mujoco: A physics engine for model-based control},
  author={Todorov, Emanuel and Erez, Tom and Tassa, Yuval},
  booktitle={2012 IEEE/RSJ international conference on intelligent robots and systems},
  pages={5026--5033},
  year={2012},
  organization={IEEE}
}

@article{shinn2023reflexion,
  title={Reflexion: Language agents with verbal reinforcement learning},
  author={Shinn, Noah and Cassano, Federico and Gopinath, Ashwin and Narasimhan, Karthik and Yao, Shunyu},
  journal={Advances in neural information processing systems},
  volume={36},
  pages={8634--8652},
  year={2023}
}

@article{madaan2023selfrefine,
  title={Self-refine: Iterative refinement with self-feedback},
  author={Madaan, Aman and Tandon, Niket and Gupta, Prakhar and Hallinan, Skyler and Gao, Luyu and Wiegreffe, Sarah and Alon, Uri and Dziri, Nouha and Prabhumoye, Shrimai and Yang, Yiming and others},
  journal={Advances in neural information processing systems},
  volume={36},
  pages={46534--46594},
  year={2023}
}

@inproceedings{ma2024eureka,
  title={Eureka: Human-level reward design via coding large language models},
  author={Ma, Yecheng Jason and Liang, William and Wang, Guanzhi and Huang, De-An and Bastani, Osbert and Jayaraman, Dinesh and Zhu, Yuke and Fan, Jim and others},
  booktitle={International conference on learning Representations},
  volume={2024},
  pages={26516--26560},
  year={2024}
}

@inproceedings{jimenez2024swebench,
  title={Swe-bench: Can language models resolve real-world github issues?},
  author={Jimenez, Carlos E and Yang, John and Wettig, Alexander and Yao, Shunyu and Pei, Kexin and Press, Ofir and Narasimhan, Karthik},
  booktitle={International Conference on Learning Representations},
  volume={2024},
  pages={54107--54157},
  year={2024}
}

@article{du2024debate,
  title={Improving factuality and reasoning in language models through multiagent debate},
  author={Du, Yilun and Li, Shuang and Torralba, Antonio and Tenenbaum, Joshua B and Mordatch, Igor},
  journal={arXiv preprint arXiv:2305.14325},
  year={2023}
}

@inproceedings{ma2024dreureka,
  title={Dreureka: Language model guided sim-to-real transfer},
  author={Ma, Jason and Liang, William and Wang, Hung-Ju and Zhu, Yuke and Fan, Linxi and Bastani, Osbert and Jayaraman, Dinesh},
  year={2024},
  booktitle={Robotics: Science and Systems (RSS)}
}

@inproceedings{pun2025brickgpt,
  title={Generating physically stable and buildable brick structures from text},
  author={Pun, Ava and Deng, Kangle and Liu, Ruixuan and Ramanan, Deva and Liu, Changliu and Zhu, Jun-Yan},
  booktitle={2025 IEEE/CVF International Conference on Computer Vision (ICCV)},
  pages={14798--14809},
  year={2025},
  organization={IEEE}
}

@inproceedings{goldberg2024bloxnet,
  title={Blox-net: Generative design-for-robot-assembly using vlm supervision, physics simulation, and a robot with reset},
  author={Goldberg, Andrew and Kondap, Kavish and Qiu, Tianshuang and Ma, Zehan and Fu, Letian and Kerr, Justin and Huang, Huang and Chen, Kaiyuan and Fang, Kuan and Goldberg, Ken},
  booktitle={2025 IEEE International Conference on Robotics and Automation (ICRA)},
  pages={15493--15500},
  year={2025},
  organization={IEEE}
}

@inproceedings{khendry2026idfra,
  title={IDfRA: Self-verification for iterative design in robotic assembly},
  author={Khendry, Nishka and Margadji, Christos and Pattinson, Sebastian W},
  booktitle={2026 IEEE International Conference on Robotics and Automation (ICRA)},
  pages={1831--1838},
  year={2026},
  organization={IEEE}
}

@article{liu2026lmbuild,
  title={LMBuild: Evaluating LLM Agents for Generating Buildable and Functional Structures},
  author={Liu, Jiateng and Wang, Rushi and Qian, Cheng and Zhang, Xuejun and Li, Sun and Liu, Jiayu and Shen, Yifan and Cao, Xu and Yao, Jiarui and Li, Bingxuan and others},
  journal={arXiv preprint arXiv:2610.04292},
  year={2026}
}

@inproceedings{wang2025mechatronics,
  title={An LLM-enabled multi-agent autonomous mechatronics design framework},
  author={Wang, Zeyu and Lo, Frank Po Wen and Chen, Qian and Zhang, Yongqi and Lin, Chen and Chen, Xu and Yu, Zhenhua and Thompson, Alexander J and Yeatman, Eric M and Lo, Benny PL},
  booktitle={Proceedings of the IEEE/CVF Conference on Computer Vision and Pattern Recognition},
  pages={4244--4254},
  year={2025}
}

@inproceedings{ringel2025text2robot,
  title={Text2robot: Evolutionary robot design from text descriptions},
  author={Ringel, Ryan P and Charlick, Zachary S and Liu, Jiaxun and Xia, Boxi and Chen, Boyuan},
  booktitle={2025 IEEE International Conference on Robotics and Automation (ICRA)},
  pages={5789--5797},
  year={2025},
  organization={IEEE}
}

@inproceedings{qiu2024robomorph,
  title={Robomorph: Evolving robot morphology using large language models},
  author={Qiu, Kevin and Pa{\l}ucki, W{\l}adys{\l}aw and Ciebiera, Krzysztof and Fija{\l}kowski, Pawe{\l} and Cygan, Marek and Kuci{\'n}ski, {\L}ukasz},
  booktitle={2026 IEEE International Conference on Robotics and Automation (ICRA)},
  pages={17611--17618},
  year={2026},
  organization={IEEE}
}

@article{koos2013transferability,
  title={The transferability approach: Crossing the reality gap in evolutionary robotics},
  author={Koos, Sylvain and Mouret, Jean-Baptiste and Doncieux, St{\'e}phane},
  journal={IEEE Transactions on Evolutionary Computation},
  volume={17},
  number={1},
  pages={122--145},
  year={2012},
  publisher={IEEE}
}

@article{bongard2006resilient,
  title={Resilient machines through continuous self-modeling},
  author={Bongard, Josh and Zykov, Victor and Lipson, Hod},
  journal={Science},
  volume={314},
  number={5802},
  pages={1118--1121},
  year={2006},
  publisher={American Association for the Advancement of Science}
}

@article{chen2026agenticreal2sim,
  title={Agentic real2sim: Physics-based world modeling with vision-language agents},
  author={Chen, Guanxiong and Xia, Qianjun and Peng, Jiawei and Zhang, Heng and Jing, Pengyu and Ma, Bole and Qian, Justin and Cheng, Yixian and Jiao, Ziyi and Zhou, Bingyang and others},
  journal={arXiv preprint arXiv:2607.19190},
  year={2026}
}

@book{boothroyd2010dfma,
  title={Product design for manufacture and assembly},
  author={Boothroyd, Geoffrey and Dewhurst, Peter and Knight, Winston A},
  year={2010},
  publisher={CRC press}
}

@article{samy2010complexity,
author = {Samy, S. N. and ElMaraghy, H.},
title = {A model for measuring products assembly complexity},
journal = {International Journal of Computer Integrated Manufacturing},
volume = {23},
number = {11},
pages = {1015--1027},
year = {2010},
publisher = {Taylor \& Francis},
doi = {10.1080/0951192X.2010.511652},
URL = {https://doi.org/10.1080/0951192X.2010.511652},
eprint = { https://doi.org/10.1080/0951192X.2010.511652}
}

@article{lehman2011novelty,
  author={Lehman, Joel and Stanley, Kenneth O.},
  journal={Evolutionary Computation}, 
  title={Abandoning Objectives: Evolution Through the Search for Novelty Alone}, 
  year={2011},
  volume={19},
  number={2},
  pages={189-223},
  doi={10.1162/EVCO_a_00025}
}

@article{friedman2023vendi,
  title={The Vendi Score: A Diversity Evaluation Metric for Machine Learning},
  author={Dan Friedman and Adji Bousso Dieng},
  journal={Transactions on Machine Learning Research},
  issn={2835-8856},
  year={2023},
  url={https://openreview.net/forum?id=g97OHbQyk1},
}

@article{shah2003ideation,
  title={Metrics for measuring ideation effectiveness},
  author={Shah, Jami J and Smith, Steve M and Vargas-Hernandez, Noe},
  journal={Design studies},
  volume={24},
  number={2},
  pages={111--134},
  year={2003},
  publisher={Elsevier}
}

@article{oquab2024dinov2,
  title={{DINO}v2: Learning Robust Visual Features without Supervision},
  author={Maxime Oquab and Timoth{\'e}e Darcet and Th{\'e}o Moutakanni and Huy V. Vo and Marc Szafraniec and Vasil Khalidov and Pierre Fernandez and Daniel HAZIZA and Francisco Massa and Alaaeldin El-Nouby and Mido Assran and Nicolas Ballas and Wojciech Galuba and Russell Howes and Po-Yao Huang and Shang-Wen Li and Ishan Misra and Michael Rabbat and Vasu Sharma and Gabriel Synnaeve and Hu Xu and Herve Jegou and Julien Mairal and Patrick Labatut and Armand Joulin and Piotr Bojanowski},
  journal={Transactions on Machine Learning Research},
  issn={2835-8856},
  year={2024},
  url={https://openreview.net/forum?id=a68SUt6zFt}
}

@article{liang2023helm,
  title={Holistic evaluation of language models},
  author={Bommasani, Rishi and Liang, Percy and Lee, Tony},
  journal={Annals of the New York Academy of Sciences},
  volume={1525},
  number={1},
  pages={140--146},
  year={2023},
  publisher={Wiley Online Library}
}

@article{agarwal2021precipice,
  title={Deep reinforcement learning at the edge of the statistical precipice},
  author={Agarwal, Rishabh and Schwarzer, Max and Castro, Pablo Samuel and Courville, Aaron C and Bellemare, Marc},
  journal={Advances in neural information processing systems},
  volume={34},
  pages={29304--29320},
  year={2021}
}

@incollection{lehman2022elm,
  title={Evolution through large models},
  author={Lehman, Joel and Gordon, Jonathan and Jain, Shawn and Ndousse, Kamal and Yeh, Cathy and Stanley, Kenneth O},
  booktitle={Handbook of evolutionary machine learning},
  pages={331--366},
  year={2023},
  publisher={Springer}
}

@article{lipson2000automatic,
  title={Automatic design and manufacture of robotic lifeforms},
  author={Lipson, Hod and Pollack, Jordan B},
  journal={Nature},
  volume={406},
  number={6799},
  pages={974--978},
  year={2000},
  publisher={Nature Publishing Group UK, London}
}

@article{cheney2013unshackling,
  title={Unshackling evolution: evolving soft robots with multiple materials and a powerful generative encoding},
  author={Cheney, Nick and MacCurdy, Robert and Clune, Jeff and Lipson, Hod},
  journal={ACM SIGEVOlution},
  volume={7},
  number={1},
  pages={11--23},
  year={2014},
  publisher={ACM, New York, NY, USA}
}

@inproceedings{luck2019coadaptation,
  title={Data-efficient co-adaptation of morphology and behaviour with deep reinforcement learning},
  author={Luck, Kevin Sebastian and Amor, Heni Ben and Calandra, Roberto},
  booktitle={Conference on Robot Learning},
  pages={854--869},
  year={2020},
  organization={PMLR}
}

@article{coros2013mechanical,
  title={Computational design of mechanical characters},
  author={Coros, Stelian and Thomaszewski, Bernhard and Noris, Gioacchino and Sueda, Shinjiro and Forberg, Moira and Sumner, Robert W and Matusik, Wojciech and Bickel, Bernd},
  journal={ACM Transactions on Graphics (TOG)},
  volume={32},
  number={4},
  pages={1--12},
  year={2013},
  publisher={ACM, New York, NY, USA}
}

@article{megaro2015interactive,
  title={Interactive design of 3d-printable robotic creatures},
  author={Megaro, Vittorio and Thomaszewski, Bernhard and Nitti, Maurizio and Hilliges, Otmar and Gross, Markus and Coros, Stelian},
  journal={ACM Transactions on Graphics (TOG)},
  volume={34},
  number={6},
  pages={1--9},
  year={2015},
  publisher={ACM, New York, NY, USA}
}

@inproceedings{jakobi1995noise,
  title={Noise and the reality gap: The use of simulation in evolutionary robotics},
  author={Jakobi, Nick and Husbands, Phil and Harvey, Inman},
  booktitle={European conference on artificial life},
  pages={704--720},
  year={1995},
  organization={Springer}
}

@inproceedings{chebotar2019closing,
  title={Closing the sim-to-real loop: Adapting simulation randomization with real world experience},
  author={Chebotar, Yevgen and Handa, Ankur and Makoviychuk, Viktor and Macklin, Miles and Issac, Jan and Ratliff, Nathan and Fox, Dieter},
  booktitle={2019 international conference on robotics and automation (ICRA)},
  pages={8973--8979},
  year={2019},
  organization={IEEE}
}

@article{hwangbo2019learning,
  title={Learning agile and dynamic motor skills for legged robots},
  author={Hwangbo, Jemin and Lee, Joonho and Dosovitskiy, Alexey and Bellicoso, Dario and Tsounis, Vassilios and Koltun, Vladlen and Hutter, Marco},
  journal={Science robotics},
  volume={4},
  number={26},
  pages={eaau5872},
  year={2019},
  publisher={American Association for the Advancement of Science}
}

@inproceedings{hanna2017grounded,
  title={Grounded action transformation for robot learning in simulation},
  author={Hanna, Josiah and Stone, Peter},
  booktitle={Proceedings of the AAAI Conference on Artificial Intelligence},
  volume={31},
  number={1},
  year={2017}
}

@article{kwiatkowski2019selfmodeling,
  title={Task-agnostic self-modeling machines},
  author={Kwiatkowski, Robert and Lipson, Hod},
  journal={Science Robotics},
  volume={4},
  number={26},
  pages={eaau9354},
  year={2019},
  publisher={American Association for the Advancement of Science}
}

@inproceedings{yao2023react,
  title={ReAct: Synergizing Reasoning and Acting in Language Models},
  author={Shunyu Yao and Jeffrey Zhao and Dian Yu and Nan Du and Izhak Shafran and Karthik Narasimhan and Yuan Cao},
  booktitle={11th International Conference on Learning Representations, ICLR 2023},
  year={2023},
  url={https://openreview.net/forum?id=WE_vluYUL-X}
}

@article{yang2024sweagent,
  title={Swe-agent: Agent-computer interfaces enable automated software engineering},
  author={Yang, John and Jimenez, Carlos and Wettig, Alexander and Lieret, Kilian and Yao, Shunyu and Narasimhan, Karthik and Press, Ofir},
  journal={Advances in Neural Information Processing Systems},
  volume={37},
  pages={50528--50652},
  year={2024}
}

@inproceedings{wang2025openhands,
  title={Openhands: An open platform for ai software developers as generalist agents},
  author={Wang, Xingyao and Li, Boxuan and Song, Yufan and Xu, Frank F and Tang, Xiangru and Zhuge, Mingchen and Pan, Jiayi and Song, Yueqi and Li, Bowen and Singh, Jaskirat and others},
  booktitle={International Conference on Learning Representations},
  volume={2025},
  pages={65882--65919},
  year={2025}
}

@inproceedings{liu2024agentbench,
  title={Agentbench: Evaluating llms as agents},
  author={Liu, Xiao and Yu, Hao and Zhang, Hanchen and Xu, Yifan and Lei, Xuanyu and Lai, Hanyu and Gu, Yu and Ding, Hangliang and Men, Kaiwen and Yang, Kejuan and others},
  booktitle={International Conference on Learning Representations},
  volume={2024},
  pages={52989--53046},
  year={2024}
}

@inproceedings{chan2025mlebench,
  title={Mle-bench: Evaluating machine learning agents on machine learning engineering},
  author={Chan, Jun Shern and Chowdhury, Neil and Jaffe, Oliver and Aung, James and Sherburn, Dane and Mays, Evan and Starace, Giulio and Liu, Kevin and Maksin, Leon and Patwardhan, Tejal and others},
  booktitle={International Conference on Learning Representations},
  volume={2025},
  pages={50466--50494},
  year={2025}
}

@inproceedings{wu2024autogen,
  title={AutoGen: Enabling Next-Gen {LLM} Applications via Multi-Agent Conversations},
  author={Qingyun Wu and Gagan Bansal and Jieyu Zhang and Yiran Wu and Beibin Li and Erkang Zhu and Li Jiang and Xiaoyun Zhang and Shaokun Zhang and Jiale Liu and Ahmed Hassan Awadallah and Ryen W White and Doug Burger and Chi Wang},
  booktitle={First Conference on Language Modeling},
  year={2024},
  url={https://openreview.net/forum?id=BAakY1hNKS}
}

@inproceedings{hong2024metagpt,
  title={MetaGPT: Meta programming for a multi-agent collaborative framework},
  author={Hong, Sirui and Zhuge, Mingchen and Chen, Jonathan and Zheng, Xiawu and Cheng, Yuheng and Wang, Jinlin and Zhang, Ceyao and Yau, Steven and Lin, Zijuan and Zhou, Liyang and others},
  booktitle={International Conference on Learning Representations},
  volume={2024},
  pages={23247--23275},
  year={2024}
}

@inproceedings{qian2024chatdev,
  title={Chatdev: Communicative agents for software development},
  author={Qian, Chen and Liu, Wei and Liu, Hongzhang and Chen, Nuo and Dang, Yufan and Li, Jiahao and Yang, Cheng and Chen, Weize and Su, Yusheng and Cong, Xin and others},
  booktitle={Proceedings of the 62nd annual meeting of the association for computational linguistics (volume 1: Long papers)},
  pages={15174--15186},
  year={2024}
}

@article{romeraparedes2024funsearch,
  title={Mathematical discoveries from program search with large language models},
  author={Romera-Paredes, Bernardino and Barekatain, Mohammadamin and Novikov, Alexander and Balog, Matej and Kumar, M Pawan and Dupont, Emilien and Ruiz, Francisco JR and Ellenberg, Jordan S and Wang, Pengming and Fawzi, Omar and others},
  journal={Nature},
  volume={625},
  number={7995},
  pages={468--475},
  year={2024},
  publisher={Nature Publishing Group UK, London}
}

@article{novikov2025alphaevolve,
  title={Alphaevolve: A coding agent for scientific and algorithmic discovery},
  author={Novikov, Alexander and V{\~u}, Ng{\^a}n and Eisenberger, Marvin and Dupont, Emilien and Huang, Po-Sen and Wagner, Adam Zsolt and Shirobokov, Sergey and Kozlovskii, Borislav and Ruiz, Francisco JR and Mehrabian, Abbas and others},
  journal={arXiv preprint arXiv:2506.13131},
  year={2025}
}

@article{lu2024aiscientist,
  title={The ai scientist: Towards fully automated open-ended scientific discovery},
  author={Lu, Chris and Lu, Cong and Lange, Robert Tjarko and Foerster, Jakob and Clune, Jeff and Ha, David},
  journal={arXiv preprint arXiv:2408.06292},
  year={2024}
}

@inproceedings{yang2024opro,
  title={Large language models as optimizers},
  author={Yang, Chengrun and Wang, Xuezhi and Lu, Yifeng and Liu, Hanxiao and Le, Quoc V and Zhou, Denny and Chen, Xinyun},
  booktitle={International Conference on Learning Representations},
  volume={2024},
  pages={12028--12068},
  year={2024}
}

@inproceedings{zhang2026compositional,
  title={Compositional machine design as program synthesis with LLMs},
  author={Zhang, Wenqian and Huang, Yangyi and Liu, Weiyang and Liu, Zhen},
  booktitle={EMNLP},
  volume={3},
  pages={6},
  year={2026}
}

@article{xia2026buildarena,
  title={Buildarena: A physics-aligned interactive benchmark of llms for engineering construction},
  author={Xia, Tian and Gao, Tianrun and Deng, Wenhao and Wei, Long and Qian, Xiaowei and Yu, Chenglei and Wu, Tailin},
  journal={arXiv preprint arXiv:2510.16559},
  year={2025}
}

@article{qiu2025debate2create,
  title={Debate2Create: Robot Co-design via Multi-Agent LLM Debate},
  author={Qiu, Kevin and Cygan, Marek},
  journal={arXiv preprint arXiv:2510.25850},
  year={2025}
}

@article{fang2025robomore,
  title={Robomore: Llm-based robot co-design via joint optimization of morphology and reward},
  author={Fang, Jiawei and Sun, Yuxuan and Ma, Chengtian and Lu, Qiuyu and Yao, Lining},
  journal={arXiv preprint arXiv:2506.00276},
  year={2025}
}

@inproceedings{song2025laser,
  title={Laser: Towards diversified and generalizable robot design with large language models},
  author={Song, Junru and Yang, Yang and Xiao, Huan and Peng, Wei and Yao, Wen and Wang, Feifei},
  booktitle={International Conference on Learning Representations},
  volume={2025},
  pages={58460--58489},
  year={2025}
}

@inproceedings{gao2026vlmgineer,
  title={VLMgineer: Vision-language models as robotic toolsmiths},
  author={Gao, George and Li, Tianyu and Shi, Junyao and Li, Yihan and Zhang, Zizhe and Figueroa, Nadia and Jayaraman, Dinesh},
  booktitle={International Conference on Learning Representations},
  volume={2026},
  pages={91732--91763},
  year={2026}
}

@article{chen2026aidsr,
  title={Bridging Language and Physics: Automated Design of Continuum Robots with Large Language Models},
  author={Chen, Jingyi and Zhang, Mohan and Yao, Laura and Ni, Yingtai and Ji, Jianmin and Peng, Jie and Wang, Song and Chen, Tianlong},
  journal={arXiv preprint arXiv:2609.08220},
  year={2026}
}

@inproceedings{alrashedy2025cadcodeverify,
  title={Generating cad code with vision-language models for 3d designs},
  author={Alrashedy, Kamel and Tambwekar, Pradyumna and Zaidi, Zulfiqar Haider and Langwasser, Megan and Xu, Wei and Gombolay, Matthew},
  booktitle={International Conference on Learning Representations},
  volume={2025},
  pages={52236--52262},
  year={2025}
}

@inproceedings{mallis2025cadassistant,
  title={Cad-assistant: Tool-augmented vllms as generic cad task solvers},
  author={Mallis, Dimitrios and Karadeniz, Ahmet Serda and Cavada, Sebastian and Rukhovich, Danila and Foteinopoulou, Niki and Cherenkova, Kseniya and Kacem, Anis and Aouada, Djamila},
  booktitle={2025 IEEE/CVF International Conference on Computer Vision (ICCV)},
  pages={7284--7294},
  year={2025},
  organization={IEEE}
}

@article{guan2025cadcoder,
  title={Cad-coder: Text-to-cad generation with chain-of-thought and geometric reward},
  author={Guan, Yandong and Wang, Xilin and Xing, Ximing and Zhang, Jing and Xu, Dong and Yu, Qian},
  journal={Advances in Neural Information Processing Systems},
  volume={38},
  pages={59765--59789},
  year={2026}
}

@inproceedings{rukhovich2025cadrecode,
  title={Cad-recode: Reverse engineering cad code from point clouds},
  author={Rukhovich, Danila and Dupont, Elona and Mallis, Dimitrios and Cherenkova, Kseniya and Kacem, Anis and Aouada, Djamila},
  booktitle={2025 IEEE/CVF International Conference on Computer Vision (ICCV)},
  pages={9801--9811},
  year={2025},
  organization={IEEE}
}

@article{dong2026muse,
  title={MUSE: Benchmarking Assembly Structure and Interface Geometry in Multi-Component Text-to-CAD Generation},
  author={Dong, Xiaoyu and Chenbo, Zeng and Li, Zhi and Wu, Xiao-Ming},
  journal={eprint arXiv: 2605.28579},
  pages={arXiv--2605},
  year={2026}
}

@article{hu2026simex,
  title={SimEX: Simulation-Integrated Robotics AutoResearch},
  author={Hu, Jiaheng and Martin-Martin, Roberto and Stone, Peter and Duan, Rocky and Jiang, Zhenyu and Shi, Guanya},
  journal={arXiv preprint arXiv:2609.38982},
  year={2026}
}

@article{qiu2026vid2sid,
  title={Vid2sid: Videos can help close the sim2real gap},
  author={Qiu, Kevin and Zhang, Yu and Cygan, Marek and Hughes, Josie},
  journal={arXiv preprint arXiv:2602.19359},
  year={2026}
}

@article{qiu2026swim2real,
  title={Swim2Real: VLM-Guided System Identification for Sim-to-Real Transfer},
  author={Qiu, Kevin and Walker, Kyle and Michelis, Mike Y and Cygan, Marek and Hughes, Josie},
  journal={arXiv preprint arXiv:2603.20827},
  year={2026}
}

@inproceedings{jiang2025phystwin,
  title={Phystwin: Physics-informed reconstruction and simulation of deformable objects from videos},
  author={Jiang, Hanxiao and Hsu, Hao-Yu and Zhang, Kaifeng and Yu, Hsin-Ni and Wang, Shenlong and Li, Yunzhu},
  booktitle={2025 IEEE/CVF International Conference on Computer Vision (ICCV)},
  pages={7219--7230},
  year={2025},
  organization={IEEE}
}

@article{guo2025engdesign,
  title={Toward engineering AGI: Benchmarking the engineering design capabilities of LLMs},
  author={Guo, Xingang and Li, Yaxin and Kong, Xiangyi and Jiang, Yilan and Zhao, Xiayu and Gong, Zhihua and Zhang, Yufan and Li, Daixuan and Sang, Tianle and Zhu, Beixiao and others},
  journal={Advances in Neural Information Processing Systems},
  volume={38},
  year={2026}
}

@article{ghugare2025builderbench,
  title={Builderbench: The building blocks of intelligent agents},
  author={Ghugare, Raj and Castanyer, Roger Creus and Ji, Catherine and Wantlin, Kathryn and Schofield, Jin and Narasimhan, Karthik and Eysenbach, Benjamin},
  journal={arXiv preprint arXiv:2510.06288},
  year={2025}
}

@article{liu2025prompt2product,
  title={Prompt-to-product: Generative assembly via bimanual manipulation},
  author={Liu, Ruixuan and Huang, Philip and Pun, Ava and Deng, Kangle and Aggarwal, Shobhit and Tang, Kevin and Liu, Michelle and Ramanan, Deva and Zhu, Jun-Yan and Li, Jiaoyang and others},
  journal={arXiv preprint arXiv:2508.21063},
  year={2025}
}

@article{wen2026bricksim,
  title={BrickSim: A Physics-Based Simulator for Manipulating Interlocking Brick Assemblies},
  author={Wen, Haowei and Liu, Ruixuan and Piao, Weiyi and Li, Siyu and Liu, Changliu},
  journal={arXiv preprint arXiv:2603.16853},
  year={2026}
}

@article{cai2026forkflush,
  title={Fork-and-Flush: Escaping Idea Basins in Autoresearch Agents},
  author={Cai, Ziyang and Ziakas, Christos and Kontonis, Vasilis and Pearce, Tim and Sen, Siddhartha and Krishnamurthy, Akshay and Garg, Shivam and Papailiopoulos, Dimitris},
  journal={arXiv preprint arXiv:2610.07447},
  year={2026}
}

@article{tan2026forgetting,
  title={Scaling Discovery Requires Forgetting},
  author={Tan, Shangyin and Zhou, Ben and Agrawal, Lakshya A and Zhang, Jialin and Chen, Yilin and Mang, Qiuyang and Lee, Donghyun and Jacob, Dennis and Sen, Koushik and Dimakis, Alex and others},
  journal={arXiv preprint arXiv:2610.20563},
  year={2026}
}

@article{sarkar2011assessing,
  title={Assessing design creativity},
  author={Sarkar, Prabir and Chakrabarti, Amaresh},
  journal={Design studies},
  volume={32},
  number={4},
  pages={348--383},
  year={2011},
  publisher={Elsevier}
}

@article{regenwetter2023beyond,
  title={Beyond statistical similarity: Rethinking metrics for deep generative models in engineering design},
  author={Regenwetter, Lyle and Srivastava, Akash and Gutfreund, Dan and Ahmed, Faez},
  journal={Computer-Aided Design},
  volume={165},
  pages={103609},
  year={2023},
  publisher={Elsevier}
}
